\documentclass[10pt,twocolumn,letterpaper]{article}
\usepackage[pagenumbers]{cvpr} 

\usepackage[dvipsnames]{xcolor}

\usepackage[accsupp]{axessibility}
\usepackage{pifont}
\usepackage{diagbox}
\usepackage{graphicx}
\usepackage{amsmath}
\usepackage{amssymb}
\usepackage{booktabs}
\usepackage{enumitem}
\usepackage{multirow}
\usepackage{booktabs}
\usepackage{array}
\usepackage{bm}
\usepackage{xcolor,colortbl}
\usepackage{color, soul}
\usepackage{bbding}
\definecolor{color4}{rgb}{0.94,0.94,1}
\definecolor{green1}{rgb}{0.13,0.8,0.13}
\definecolor{red1}{rgb}{0.9,0.2,0.2}
\setitemize[1]{itemsep=0pt,partopsep=0pt,parsep=\parskip,topsep=5pt}

\usepackage[pagebackref,breaklinks,colorlinks,citecolor=cvprblue]{hyperref}
\usepackage[capitalize]{cleveref}
\crefname{section}{Sec.}{Secs.}
\Crefname{section}{Section}{Sections}
\Crefname{table}{Table}{Tables}
\crefname{table}{Tab.}{Tabs.}
\definecolor{cvprblue}{rgb}{0.21,0.49,0.74}

\def\paperID{3001} 
\def\confName{CVPR}
\def\confYear{2024}

\begin{document}
\title{AdaRevD: Adaptive Patch Exiting Reversible Decoder \\
Pushes the Limit of Image Deblurring}

\author{Xintian Mao
~~~~~
Qingli Li
~~~~~
Yan Wang\footnotemark[1] \\
Shanghai Key Laboratory of Multidimensional Information Processing\\ East China Normal University\\
\tt\small 52265904010@stu.ecnu.edu.cn, qlli@cs.ecnu.edu.cn, ywang@cee.ecnu.edu.cn\\
\small Code: \url{https://github.com/DeepMed-Lab-ECNU/Single-Image-Deblur}
\vspace{-1.5em}
}

\twocolumn[{%
\renewcommand\twocolumn[1][]{#1}%
\maketitle
\begin{center}
    \centering
    \captionsetup{type=figure}
    \vspace{-1.em}    \includegraphics[width=0.96\linewidth]{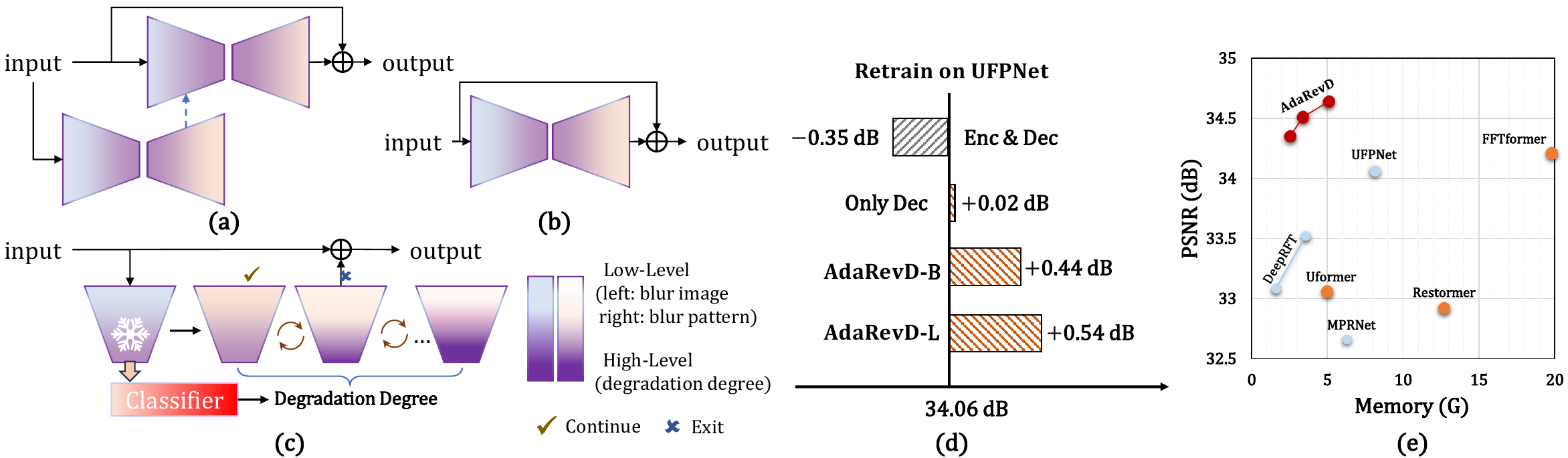}
    \vspace{-1.em}
    \caption{
    \label{fig:models-structure}
    Architectures and performance comparisons of AdaRevD and other methods. (a) multi-stage architecture~\cite{Chen2021hinet, Zamir2021multi}; (b) one-stage architecture~\cite{Chen2022simple, Wang2022uformer, Zamir2021restormer}; (c) architecture of our method; (d) comparison of different continue-training strategies; (e) PSNR \emph{vs.} training memory cost on GoPro dataset. Our method pushes the limit of image deblurring by exploring the insufficient decoding capability.} 
\end{center}%
}]
 \setcounter{footnote}{0}
 \renewcommand{\thefootnote}{\fnsymbol{footnote}}
 \footnotetext[1]{Corresponding author: Y. Wang (\url{ywang@cee.ecnu.edu.cn})}
 \renewcommand{\thefootnote}{\arabic{footnote}}

\begin{abstract}
Despite the recent progress in enhancing the efficacy of image deblurring, the limited decoding capability constrains the upper limit of State-Of-The-Art (SOTA) methods. This paper proposes a pioneering work, Adaptive Patch Exiting Reversible Decoder (AdaRevD), to explore their insufficient decoding capability. By inheriting the weights of the well-trained encoder, we refactor a reversible decoder which scales up the single-decoder training to multi-decoder training while remaining GPU memory-friendly. Meanwhile, we show that our reversible structure gradually disentangles high-level degradation degree and low-level blur pattern (residual of the blur image and its sharp counterpart) from compact degradation representation. Besides, due to the spatially-variant motion blur kernels, different blur patches have various deblurring difficulties. We further introduce a classifier to learn the degradation degree of image patches, enabling them to exit at different sub-decoders for speedup. Experiments show that our AdaRevD pushes the limit of image deblurring, e.g., achieving 34.60~dB in PSNR on GoPro dataset. 
\end{abstract} 
 
\section{Introduction}
\label{sec:intro}

As a sub-task of image restoration, image deblurring aims at removing degraded blur artifacts to recover clean images. 
Generally, two types of backbone networks, \emph{i.e.}, multi-stage and one-stage architectures, consisting of multiple encoders and decoders, have been proposed for the image deblurring task. Encoders are used to learn a compact degradation representation from a blur image and decoders decode the degradation representation to blur patterns\footnote{The residual of the blur image and its sharp counterpart is defined as blur pattern in this paper.}. The \emph{compact degradation representation} can be treated as a mid-level feature, equipped with \emph{high-level degradation degree}\footnote{The degradation degree, \emph{i.e.}, the difficulty of restoring an image or a patch, is decided by the PSNR of blur image and its sharp counterpart.} information and \emph{low-level blur pattern}. See Fig.~\ref{fig:cka} and Sec.~\ref{sec:disentangle} for detailed analysis. 

Multi-stage architectures \cite{Nah2017deep,Zamir2021multi,Chen2021hinet} (see Fig.~\ref{fig:models-structure} (a)) decompose the feature extraction process into multiple sub-networks. 
Recent research explorations for image restoration \cite{Zamir2021restormer, Wang2022uformer,Chen2022simple} have shown the ability of one-stage architecture (see Fig.~\ref{fig:models-structure} (b)). Instead of focusing on the overall design of the model, more attention is paid to the core components design of the one-stage architecture, such as Res-Block~\cite{He2016ResNet} and Transformer-Block~\cite{vaswani2017transformer}. In fact, several one-stage architectures have delicately designed heavyweight encoders. 
MSDI-Net~\cite{li2022MSDINet} introduces an extra degradation representation encoder to enhance the image deblurring performance.
In UFPNet~\cite{fang2023UFPNet}, the kernel prior module is pre-trained to estimate the spatially variant blur kernel information, which is then integrated into the encoder. Although the encoder is delicately designed to learn robust degradation representation,
the size of the whole network is limited to what the GPU can accommodate. Thus, these networks have to design lightweight decoders to decode blur patterns from the degradation representation. Insufficient decoding capability constrains the model's upper limit. 

The performance of existing image deblurring networks reaches saturation when training is completed, signifying the performance pinnacle of the current network. Even with more training iterations, the performance remains unchanged, or even leads to a decrease in performance, as shown in Fig.~\ref{fig:models-structure} (d). 
But, constrained by the lightweight decoders, the deblurring results may not be optimal. 
This makes us wonder: Is it possible to \emph{inherit} the weights of the well-trained encoder and \emph{refactor} the decoder with high capacity, enabling us to map the learned degradation representation to the blur pattern more effectively and push the limits of image deblurring? 
\begin{figure}[t]
\begin{center}
    \includegraphics[width=0.9 \linewidth]{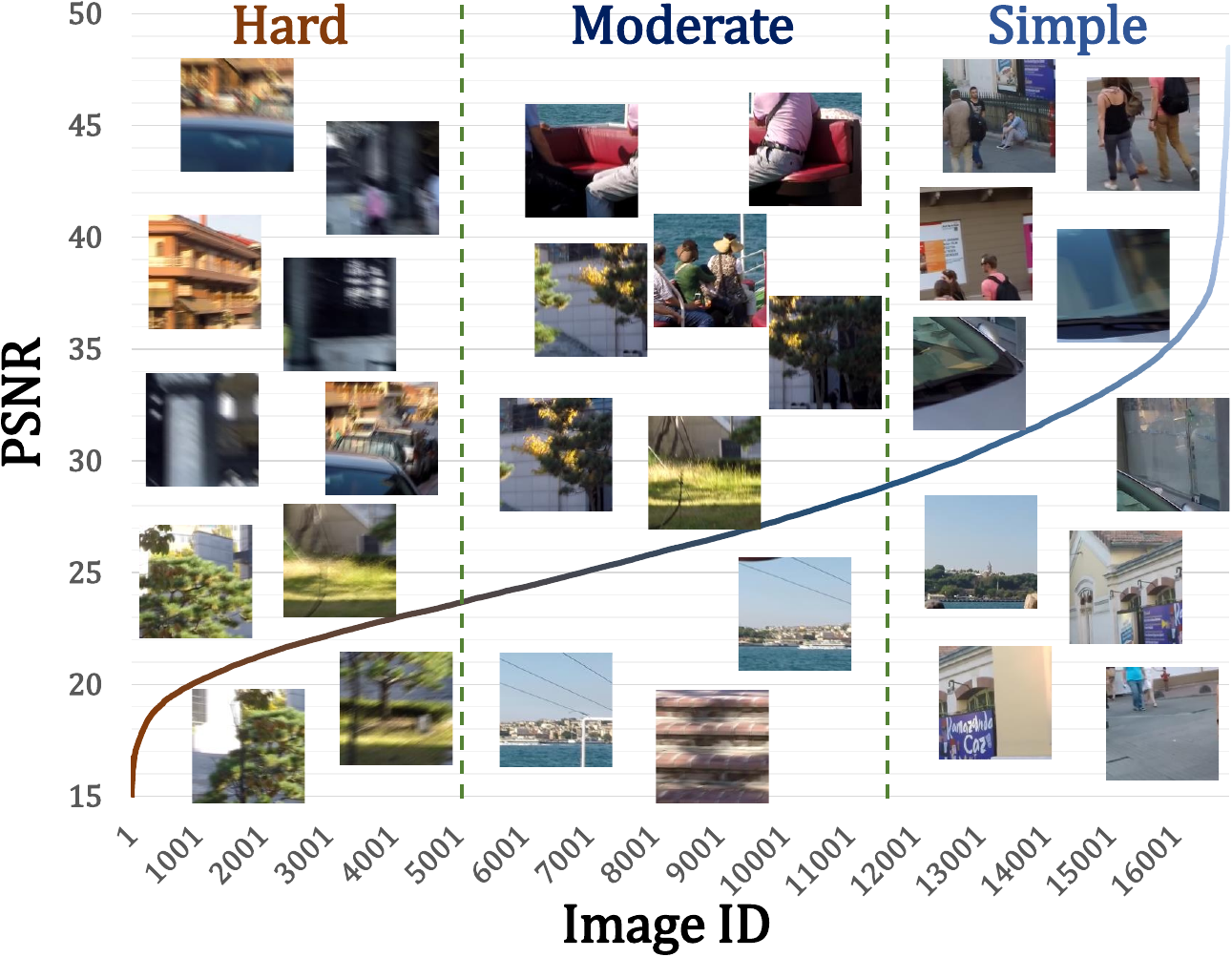}
\end{center}
\vspace{-1.5em}
\caption{The ranked PSNR curve of the image patches from GoPro~\cite{Nah2017deep} train set and the visualization of the patches with various degradation degrees (\emph{e.g.} hard, moderate and simple).} 
\label{fig:diff_psnr}
\vspace{-1.5em}
\end{figure}

A straightforward solution is to refactor the original decoder with a heavyweight decoder, and retrain the network by only updating the decoder. Since the capability of the decoder is constrained by the GPU memory, a memory-saving learning paradigm is important, especially for large models. Besides, for a typical decoder, layers close to the encoder contain more high-level degradation information, while features close to the output are decoded blur patterns, indicating the pixel-level residual between blur and sharp images. Learning a direct mapping from a compact degradation representation to the pixel-level blur pattern like previous decoders may suffer from inferior performances during testing, since information unrelated to the target details may be gradually decoded and accumulated during the layer-by-layer propagation due to the UNet architecture. 

\begin{figure}[t]
\begin{center}
    \includegraphics[width=0.95 \linewidth]{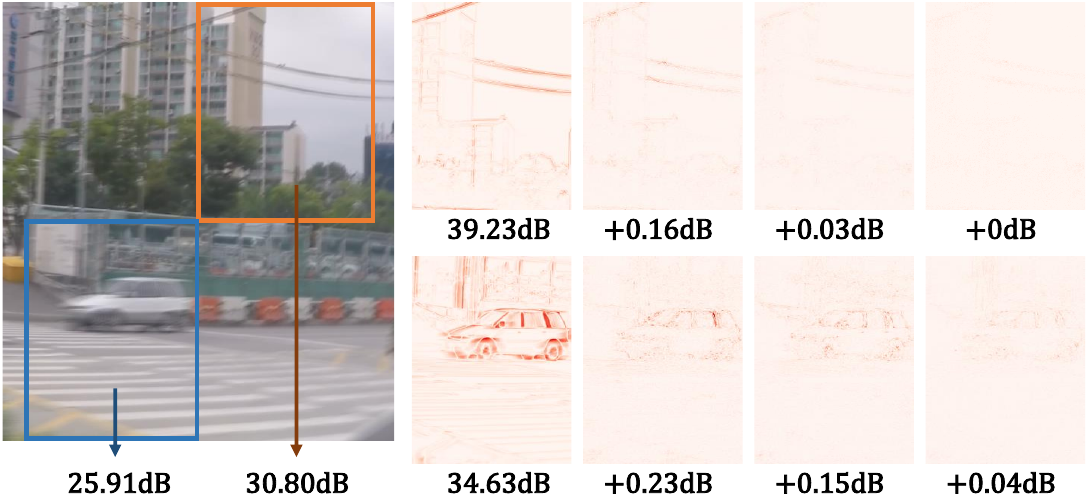}
\end{center}
\vspace{-1.5em}
\caption{Visual comparison of the outputs from different sub-decoders. The first column is the difference between blur image and the first sub-decoder's output. The rest of the columns are the residual between the current sub-decoder and the former one.} 
\label{fig:diff_dec}
\vspace{-1.5em}
\end{figure}

To solve the problems and push the limit of image deblurring model, we propose an Adaptive Patch Exiting Reversible Decoder (AdaRevD), as shown in Fig.~\ref{fig:models-structure}(c). Concretely, our decoder is composed of $K$ sub-decoders, trained in a reversible manner \cite{cai2022revcol}, each of which takes the degradation representation as input and generates a sharp prediction. A sub-decoder consists of multi-level features, from high-level semantics to low-level blur patterns. This design enables a large model capacity while consuming only limited GPU memory. It progressively separates low-level and high-level information by disentangling features. It is the first attempt to train a memory-friendly decoder in a backward feature propagation manner without loss of information in the field of high-resolution image deblurring. 

Additionally, as shown in Fig.~\ref{fig:diff_psnr} and Fig.~\ref{fig:diff_dec}, given the varying degrees of blur in \emph{image patches}\footnote{To include sufficient content information for deblurring, we use a reasonably large patch size (384$\times$384) to our model. Image patch also can be understood as sub-image.}, the difficulty of restoring each patch varies. With gradually stacking more sub-decoders, the restoration performances of different patches may reach bottlenecks. 
Due to the obstacles brought by the deblurring network's multi-scale feature property, no prior work attempts to design an exiting strategy for image deblurring. Thanks to our sub-decoder structures, we design an adaptive patch exiting strategy effortlessly. We employ a classifier to predict the degradation degree of each blur patch, allowing a patch to exit at a specific sub-decoder to achieve speedup. 

The main contributions can be summarized as follows:

\begin{itemize}
    \item We propose a first work to push the limit of State-Of-The-Art (SOTA) image deblurring networks by exploring their insufficient decoding capability.
    \item We propose a reversible deblurring decoder with a substantial capacity while remaining GPU memory-friendly. Meanwhile, it gradually disentangles high-level degradation degree and low-level task-related features, learning blur patterns while maintaining the overall semantics. 
    \item We propose a simple classifier to get the degradation degree of the input image, enabling the patches to exit at various decoders for speedup.
    \item Extensive experiments show that our proposed AdaRevD can push the limit of image deblurring. It achieves SOTA results on image deblurring task, \emph{e.g.}, 34.60~dB in PSNR for GoPro dataset. The PSNR (dB) \emph{vs.} GPU-memory (G) compared with others is shown in Fig.~\ref{fig:models-structure} (e).
\end{itemize}

 \section{Related Works}
\label{sec:related}

\subsection{Deblurring Methods}
Recently, many methods~\cite{Nah2017deep, Tao2018scale, Kupyn2018deblurgan, Lin2019TellMW, Zhang2019DMPHN, Zamir2021multi, Cho2021rethinking, XintianMao2023DeepRFT, Tu2022maxim, Chen2022simple, Whang_2022_CVPR, Zamir2021restormer, fang2023UFPNet, kong2023fftformer} apply end-to-end trained deep neural network for image deblurring. In order to achieve better performance, most improvements are made around the model structure or the specific component. For the structure, many studies such as DeepDeblur~\cite{Nah2017deep}, DMPHN~\cite{Zhang2019DMPHN} and MPRNet~\cite{Zamir2021multi} prefer to use multi-stage architecture, which learns degradation pattern progressively. The diffusion-based work~\cite{Whang_2022_CVPR} trains a stochastic sampler that refines the output of a deterministic predictor. Per contra, the design of specific block with UNet shows its capacity of deblurring. NAFNet~\cite{Chen2022simple} applies LayerNorm (LN)~\cite{JimmyBa2016LayerN} to stabilize the training process with a high initial learning rate. Uformer~\cite{Wang2022uformer} and Stripformer~\cite{Tsai2022Stripformer} apply Local Self-Attention (SA) to capture long-range dependencies with low complexity. Restormer~\cite{Zamir2021restormer} models global context by Global Channel SA. DeepRFT~\cite{XintianMao2023DeepRFT} proposes Res-FFT-ReLU-Block for frequency selection. MRLPFNet~\cite{dong2023MRLPFNet} constructs Residual Low-Pass Filter Module based on DeepRFT~\cite{XintianMao2023DeepRFT} and Restormer~\cite{Zamir2021restormer}. However, few methods consider the varying levels of image degradation. 

\subsection{Reversible Architectures}
In the process of gradient backpropagation, 
a lot of resources are used to store intermediate features. As the networks become deeper and wider for SOTA performance, GPU memory has been a bottleneck limiting the further development of the model. To solve this bottleneck, Reversible Residual Block (RevBlock)~\cite{gomez2017reversible} 
lets each block's activations be reestablished from the following ones. \textit{i}-RevNet~\cite{jacobsen2018irevnet} builds a fully inverted network by providing an explicit inverse.  Rev-ViT~\cite{mangalam2022revViT} extends reversible CNN block to reversible Transformer block, which promotes saving GPU memory and allows training ViT model with higher batch size. RevBiFPN~\cite{chiley2023revBiFPN} builds a fully reversible bidirectional feature pyramid network by BiFPN~\cite{tan2020BiFPN}. RevCol~\cite{cai2022revcol} proposes a reversible column-based foundation model design paradigm. In accordance with this paradigm, our AdaRevD is the inaugural endeavor to incorporate reversible columns as sub-decoders for image deblurring. This approach maintains a consistent consumption of GPU memory during training, akin to a single sub-decoder. Furthermore, the utilization of multiple sub-decoders facilitates the exit of the network's forward propagation through the pertinent sub-decoder.

\begin{figure*}[t]
\begin{center}
\includegraphics[width=0.87\linewidth]{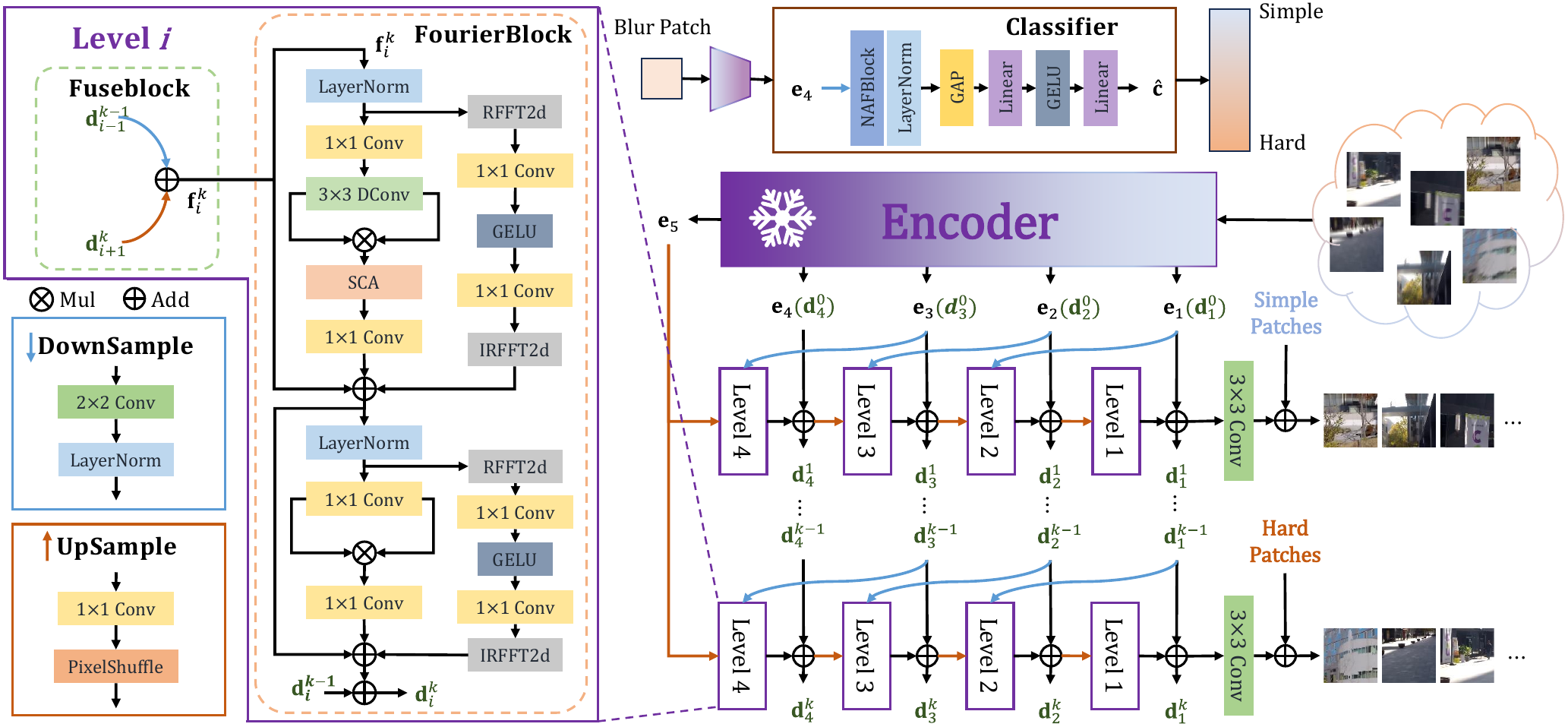} 
\end{center}
\vspace{-1.em}
\caption{Architecture of AdaRevD. AdaRevD consists of three parts: a pre-trained encoder, several sub-decoders and a classifier. To push the limit of image deblurring networks, and map the learned degradation representation to the blur pattern more effectively, the pre-trained encoder is fixed during training. Each sub-decoder is composed of four Level modules, including a Fuseblock and a FourierBlock. SCA means Simple Channel Attention proposed in NAFNet~\cite{Chen2022simple}. The classifier predicts the degradation degree of each image patch, which allows the network to exit in the appropriate sub-decoder.} 
\label{fig:framework}
\vspace{-1.em}
\end{figure*}

\subsection{Adaptive Inference}
Since the kernels are always spatially-variant for the blur image, different patches receive various degrees of degradation. Facing the deployment of neural networks under different conditions, many methods~\cite{yu2018slimmable, yu2019universally, yu2019autoslim, jin2020adabits, yang2020mutualnet, cai2020once, liu2020deepada, kong2021classsr, wang2022adaptiveSR} are proposed to permit instant and adaptive accuracy-efficiency trade-offs at runtime. Slimmable networks~\cite{yu2018slimmable} enable a single network executable at different widths which can instantly adjust the width in runtime. For image super-resolution, different image regions usually have various restoration difficulties~\cite{wang2022adaptiveSR}. AdaDSR~\cite{liu2020deepada} introduces a lightweight adapter module to predict the network depth map, which facilitates efficient adaptive inference with sparse convolution. ClassSR~\cite{kong2021classsr} classifies the repair difficulty of patches into three types: easy, medium, and difficult. Different types are restored by various networks. APE-SR~\cite{wang2022adaptiveSR} exits early in the intermediate ResBlocks by using a simple regressor to estimate the incremental prediction. Unlike the image super-resolution models which apply a sequence of ResBlocks~\cite{He2016ResNet} without down-sample and up-sample as the main body, deblurring models always tend to use UNet~\cite{UNet} model. However, The structure of UNet model is not conducive to exit in different stages. Thus, it is difficult to apply these adaptive patterns to deblurring model directly. In AdaRevD, we construct a multi-decoder architecture with a degradation degree classifier, enabling the possibility to exit in different sub-decoders.

\section{Method}
\label{sec:method}

\subsection{Main Backbone}
Recently, deep learning based deblurring methods always employ a network composed of four parts: head, encoder, decoder and tail, which directly map a blur image to its paired sharp image. This architectural design has gained mainstream prominence in recent years~\cite{Chen2022simple}. The head part usually applies a $3\times 3$ convolutional layer, whose parameters are denoted as $\mathbf{\Theta}_{head}$ to extract shallow features $\mathbf{h}$ from blur image $\mathbf{B}$\vspace{-0.5em}:
\begin{equation}
\label{eq:head}
    \mathbf{h}=\mathcal{H}(\mathbf{B}, \mathbf{\Theta}_{head}).
    \vspace{-0.5em}
\end{equation}
Then, the encoder, parameterized by $\mathbf{\Theta}_{enc}$, takes $\mathbf{h}$ as the input and generates $N$ hidden features $\mathbf{e}_1$, $\mathbf{e}_2$, ..., $\mathbf{e}_N$\vspace{-0.5em}:
\begin{equation}
\label{eq:encoder}
    \mathbf{e}_1, \mathbf{e}_2, ..., \mathbf{e}_N = \mathcal{E}(\mathbf{h}, \mathbf{\Theta}_{enc}).
    \vspace{-0.5em}
\end{equation}
After that, the decoder, parameterized by $\mathbf{\Theta}_{dec}$, decodes the hidden features as the blur pattern feature $\mathbf{d}_1$\vspace{-0.5em}:
\begin{equation}
\label{eq:decoder}
    \mathbf{d}_1=\mathcal{D}(\mathbf{e}_1, \mathbf{e}_2, ..., \mathbf{e}_N, \mathbf{\Theta}_{dec}).
    \vspace{-0.5em}
\end{equation}
At last, the tail part, parameterized by $\mathbf{\Theta}_{tail}$, takes the degradation feature to obtain the blur pattern $\mathbf{t}$ with a $3\times 3$ convolutional layer and obtain the restored sharp image $\hat{\mathbf{S}}$\vspace{-0.5em}:
\begin{align}
    &\mathbf{t}=\mathcal{T}(\mathbf{d}_1, \mathbf{\Theta}_{tail})\\
    &\hat{\mathbf{S}}={\mathbf{B}}+\mathbf{t}.
    \vspace{-0.5em}
\end{align}
In our AdaRevD, we construct a multi-decoder structure. The intermediate feature from Level $i$ of $j$th sub-decoder is denoted as $\mathbf{d}_i^j$. Thus, Eq.~\ref{eq:decoder} is rewritten as\vspace{-0.5em}:
\begin{align}
    \mathbf{d}_1^1,...,\mathbf{d}_{N-1}^1&=\mathcal{D}^1(\mathbf{e}_1,...,\mathbf{e}_N, \mathbf{\Theta}_{dec}^1),\\
    \mathbf{d}_1^k,...,\mathbf{d}_{N-1}^k&=\mathcal{D}^1(\mathbf{d}_1^{k-1},...,\mathbf{d}_{N-1}^{k-1},\mathbf{e}_N, \mathbf{\Theta}_{dec}^k).
    \vspace{-0.5em}
\end{align}
Finally, the restored sharp image from $j$th decoder is\vspace{-0.5em}:
\begin{equation}
    \label{eq:restore}\hat{\mathbf{S}}^j=\mathbf{B}+\mathcal{T}^j(\mathbf{d}_{{1}}^j,\mathbf{\Theta}_{tail}^j).\vspace{-0.5em}
\end{equation}

The overview of AdaRevD is shown in Fig.~\ref{fig:framework}. Different from RevCol~\cite{cai2022revcol} , which separates the encoder into multiple reversible sub-encoders, AdaRevD builds a reversible model which contains multiple reversible sub-decoders. For practical speedup, we design an adaptive patch-exiting method on top of the reversible structures to determine whether to early exit in various sub-decoders by a classifier.  

There are two training phases in AdaRevD: decoder training and classifier training. In decoder training, AdaRevD optimizes the decoder based on the well trained encoder from UFPNet~\cite{fang2023UFPNet} and keeps the encoder frozen during training for lower memory consumption. In the classifier training phase, only the classifier will be optimized. Thanks to the reversible architecture, AdaRevD can enhance the size of a single-stage model, thereby equipping it with a larger capacity, all while maintaining low memory requirements. Thanks to our adaptive classifier, AdaRevD enables the patch to exit at the optimal layer. For sub-decoders, we propose a FourierBlock for large-scale sensing capability purposes. From Level-1 to Level-4, the number of blocks are $[1,1,1,1]$ (each sub-decoder).  

\subsection{Reversible Decoder}
\paragraph{Forward and Inverse Structure}
RevCol \cite{cai2022revcol} builds a reversible column architecture composed of multiple sub-networks (columns) in the encoder for classification-related tasks. But, image deblurring models \cite{fang2023UFPNet,Chen2022simple} are usually designed with very heavy encoders. Thus, fewer benefits can be gained from reversible architectures. Instead, we propose reversible decoders which take reversible columns as sub-decoders to save the GPU memory consumption during training. It is worth noting that the reversible decoder facilitates the utilization of early exit when representations are presented in a hierarchical multi-scale manner. Formally, the forward and inverse computations are\vspace{-0.5em}:
\begin{align}
\small
\text{Forward: }\mathbf{d}_i^j=\begin{cases}
     \mathcal{L}_i^j(\mathbf{d}_{i+1}^j, \mathbf{d}_{i-1}^{j-1}) + \alpha \mathbf{d}_i^{j-1}~~~~~&i > 1  
    \\
    \mathcal{L}_i^j(\mathbf{d}_{i+1}^j) + \alpha \mathbf{d}_i^{j-1}~~~~~&i = 1  
    \end{cases}
\label{eq:reversible-forward}\vspace{-0.5em}
\end{align}
\begin{align}
\small
\text{Inverse: }\mathbf{d}_i^{j-1}=\begin{cases}
     \alpha^{-1}(\mathbf{d}_i^j-\mathcal{L}_i^j(\mathbf{d}_{i+1}^j, \mathbf{d}_{i-1}^{j-1}))~&i > 1  
    \\
    \alpha^{-1}(\mathbf{d}_i^j-\mathcal{L}_i^j(\mathbf{d}_{i+1}^j))~&i = 1  
    \end{cases}
\label{eq:reversible-backward}\vspace{-0.5em}
\end{align}
where $\mathcal{L}_i^j$ is the $i$th Level module of $j$th decoder , $\mathbf{d}_i^j$ is the output feature of the $i$th Level in the $j$th decoder, $\alpha$ is the learnable  scaling parameter.

\paragraph{Level Module}
The Level modules in reversible decoders are composed of two blocks: FuseBlock and FourierBlock. Because reversible architecture can help us save the training memory, heavier modules can be used to replace the NAFBlock~\cite{Chen2022simple}  for further improvement of deblurring. In AdaRevD, we apply a Fuseblock to fuse the feature maps from the current and previous sub-decoder and a FourierBlock~\cite{XintianMao2023DeepRFT} to decode better blur pattern.

\begin{figure}[t]
\begin{center}
    \includegraphics[width=0.95 \linewidth]{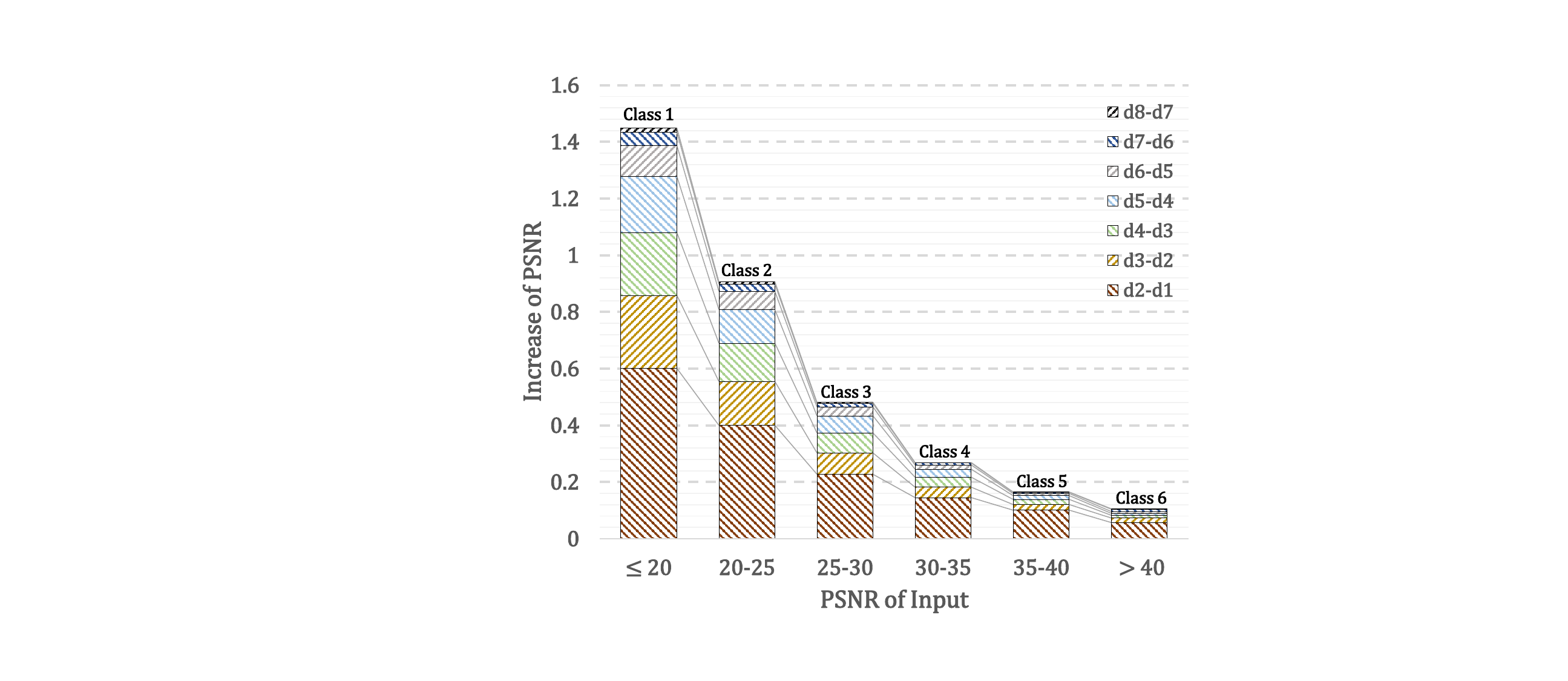}
\end{center}
\vspace{-1.em}
\caption{The increment for degraded patches belonging to each degradation degree class in different sub-decoders. The patches are generated from GoPro~\cite{Nah2017deep} train set. d$(i)$-d$(i-1)$ means the average increment PSNR for the patches in the $i$th sub-decoder.} 
\label{fig:grow_psnr_gopro_8decoder}
\vspace{-1.em}
\end{figure}

\subsection{Adaptive Classifier}
In order to establish a multi-exit deblur model, we design a multi-decoder structure which predicts sharp images in each sub-decoder. Generally, the performance of the model usually increases when the model goes deeper. However, the blur kernels are always spatially-variant, which means the recoverability of image patches varies. Fig.~\ref{fig:diff_dec} and Fig.~\ref{fig:grow_psnr_gopro_8decoder} indicate that the incremental capacity of each sub-decoder varies with different degraded patches. There is no need to cost overmany sub-decoders for the patches whose PSNRs are already high, \emph{e.g.}, higher than 40~dB. For GoPro~\cite{Nah2017deep} dataset, we group the patches to 6 degradation degrees ($\tilde{c}$) via the PSNR between the blur patch and the sharp patch: ($\le$ 20~dB, $\tilde{c}=1$), ($\le$ 25~dB, $\tilde{c}=2$), ($\le$ 30~dB, $\tilde{c}=3$), ($\le$ 35~dB, $\tilde{c}=4$), ($\le$ 40~dB, $\tilde{c}=5$) and ($>$ 40~dB, $\tilde{c}=6$). Then, an additional classifier is introduced to predict the degradation degree of the blur patch\vspace{-0.5em}: 
\begin{align}
\hat{{c}}&=\mathtt{Classifier}(\mathbf{e}_4).\label{eq:reg1} 
\end{align}
The classifier takes $\mathbf{e}_4$ as the input and predicts the degradation degree classification $\hat{{c}}$ of the input patch. As indicated in Fig.~\ref{fig:framework}, the classifier contains a DownSample layer, an NAFBlock, a LayerNorm layer and an MLP (GAP-Linear-GELU-Linear) block. $\mathtt{GAP}$ means global average pooling.

To avoid consuming computing resources on simple patches, we define the early-exit signal ${E}_c$, indicating which sub-decoder to exit when processing a patch belonging to the $c$th class. ${E}_c$ is defined as\vspace{-0.5em}:
\begin{equation}
\label{eq:Ec}
    E_c = \left\{ \begin{array}{cl}
j-1,~~~& \text{if} ~~~{\exists}~ \mathbf{O}_c^j < \tau, \\
J,~~~& \text{otherwise},
\end{array} \right. 
\vspace{-0.5em}
\end{equation}
where $J$ means the total sub-decoder number, $\tau$ is a pre-defined threshold, and $\mathbf{O}_c^j$ is collected during the training process, calculated as\vspace{-0.5em}:
\begin{equation}
    \mathbf{O}_c^j= \frac{1}{|\Omega_c|}\sum_{\mathbf{P}\in\Omega_c}\mathtt{PSNR}(\hat{\mathbf{P}}^j, \mathbf{P})-\mathtt{PSNR}(\hat{\mathbf{P}}^{j-1}, \mathbf{P}),
    \vspace{-0.5em}
\end{equation}
where $\mathbf{P}$ is a sharp patch and 
$\Omega_c$ means the set of sharp patches whose corresponding blur patches belong to the $c$th class. $|\Omega_c|$ is the cardinality of $\Omega_c$. $\hat{\mathbf{P}}^j$ means the prediction from the $j$th sub-decoder. 
With the incremental prediction $\mathbf{O}_{c}^j$ and the degradation degree classification $\hat{\mathbf{C}}$ of the input patch, AdaRevD is able to let the patch exit in the $(j-1)$th sub-decoder when $\mathbf{O}_{c}^j$ is smaller than the threshold $\tau$.



\begin{figure*}[t]
\captionsetup[subfigure]{justification=centering, labelformat=empty}
 \centering
	\begin{subfigure}{0.12\linewidth}
	\includegraphics[width=0.99\linewidth]{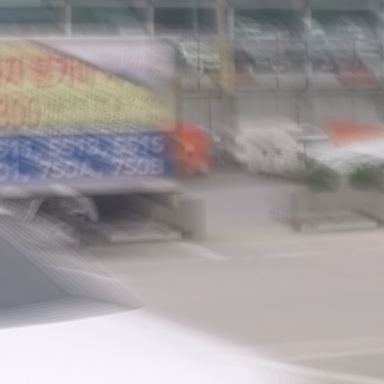}
	\end{subfigure}
 \centering
	\begin{subfigure}{0.12\linewidth}
	\includegraphics[width=0.99\linewidth]{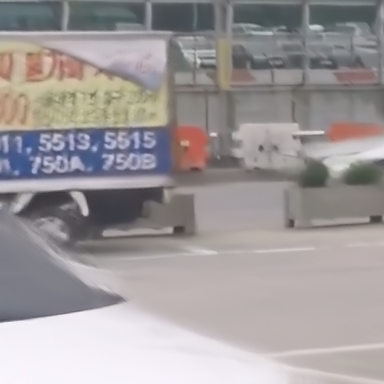}
	\end{subfigure}
 \centering
	\begin{subfigure}{0.12\linewidth}
	\includegraphics[width=0.99\linewidth]{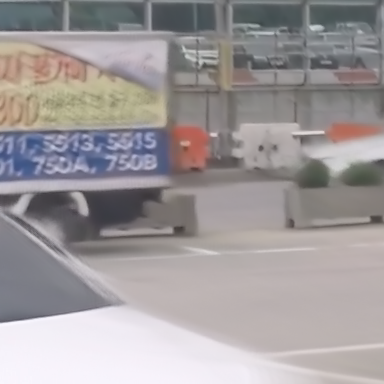}
	\end{subfigure}
 \centering
  	\begin{subfigure}{0.12\linewidth}
	\includegraphics[width=0.99\linewidth]{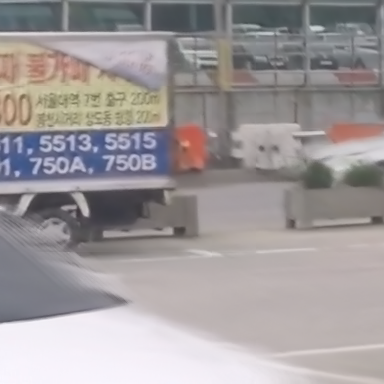}
	\end{subfigure}
 \centering
 	\begin{subfigure}{0.12\linewidth}
	\includegraphics[width=0.99\linewidth]{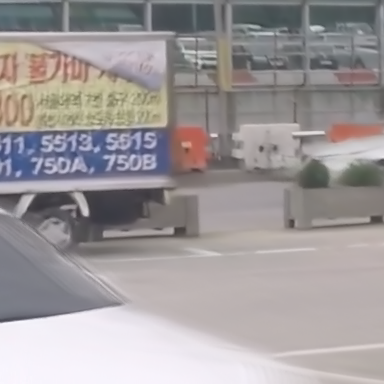}
	\end{subfigure}
 \centering
	\begin{subfigure}{0.12\linewidth}
	\includegraphics[width=0.99\linewidth]{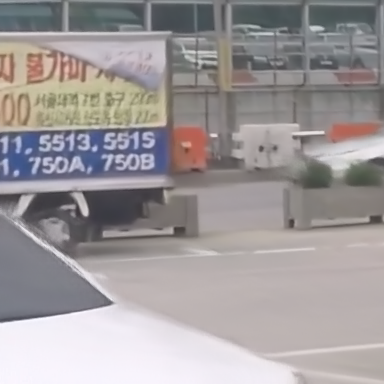}
	\end{subfigure}
 \centering
	\begin{subfigure}{0.12\linewidth}
	\includegraphics[width=0.99\linewidth]{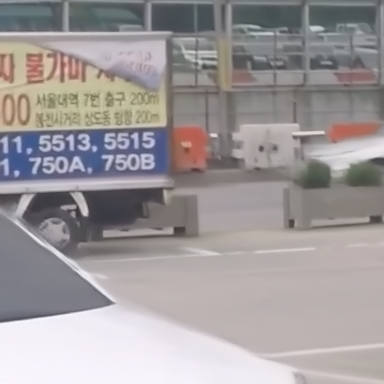}
	\end{subfigure}
 \centering
     \begin{subfigure}{0.12\linewidth}
	\includegraphics[width=0.99\linewidth]{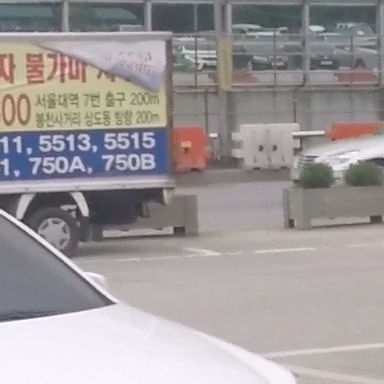}
	\end{subfigure}
	\quad
 \centering
	\begin{subfigure}{0.12\linewidth}
	\includegraphics[width=0.99\linewidth]{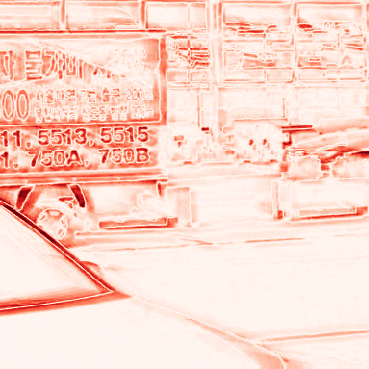}
        \caption{Blur}
	\end{subfigure}
 \centering
	\begin{subfigure}{0.12\linewidth}
	\includegraphics[width=0.99\linewidth]{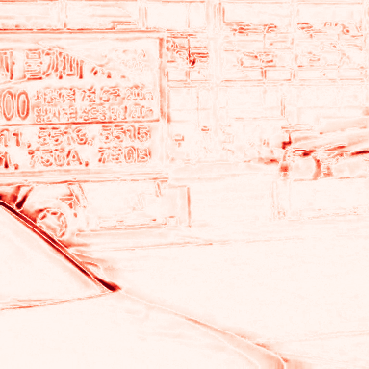}
    \caption{MPRNet~\cite{Zamir2021multi}} 
	\end{subfigure}
 \centering
	\begin{subfigure}{0.12\linewidth}
	\includegraphics[width=0.99\linewidth]{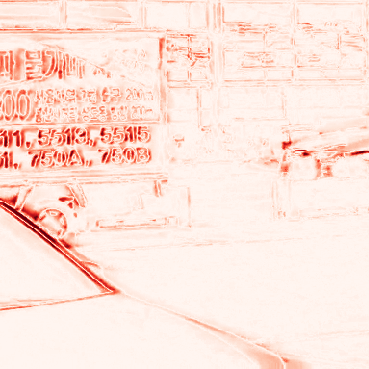}
    \caption{Restormer~\cite{Zamir2021restormer}} 
	\end{subfigure}
 \centering
  	\begin{subfigure}{0.12\linewidth}
	\includegraphics[width=0.99\linewidth]{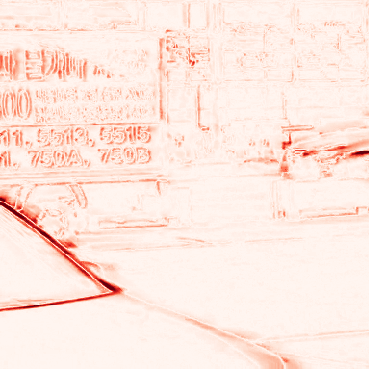}
        \caption{DeepRFT+\cite{XintianMao2023DeepRFT}} 
	\end{subfigure}
 \centering
 	\begin{subfigure}{0.12\linewidth}
	\includegraphics[width=0.99\linewidth]{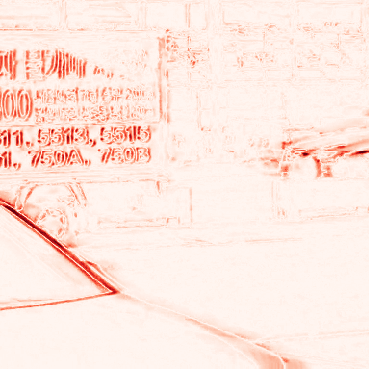}
  \caption{NAFNet64~\cite{Chen2022simple}} 
	\end{subfigure}
 \centering
	\begin{subfigure}{0.12\linewidth}
	\includegraphics[width=0.99\linewidth]{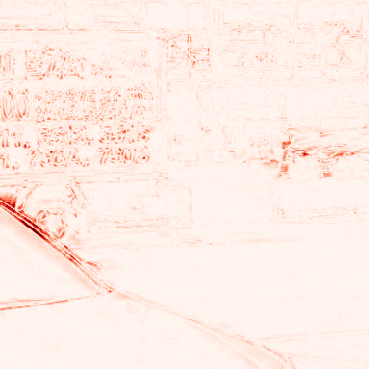}
 	\caption{UFPNet\cite{fang2023UFPNet}} 
	\end{subfigure}
 \centering
	\begin{subfigure}{0.12\linewidth}
	\includegraphics[width=0.99\linewidth]{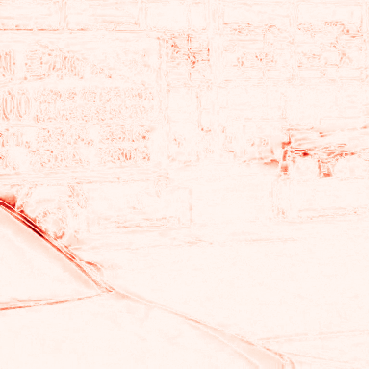}
     	\caption{Ours} 
	\end{subfigure}
 \centering
     \begin{subfigure}{0.12\linewidth}
	\includegraphics[width=0.99\linewidth]{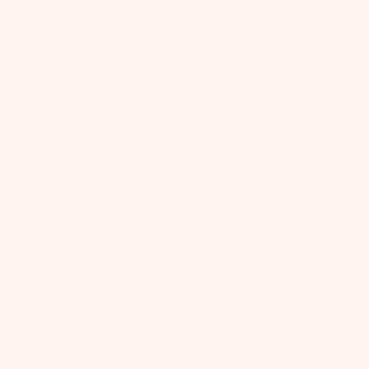}
     	\caption{Sharp}
	\end{subfigure}
\vspace{-0.7em}
\caption{Examples on the GoPro test dataset. The first row shows blur image, predicted images of different methods, and ground-truth sharp image. The second row shows the residual of the blur image / predicted sharp images and the ground-truth sharp image.}
\label{fig:GoPro}
\vspace{-1.5em}
\end{figure*}

\subsection{Loss Function}
For decoder training phase, the loss is defined as\vspace{-0.5em}: 
\begin{align}
\ell_{m}=\ell_{1}+0.01\ell_{fr}\label{eq:loss1}\vspace{-0.5em}
\end{align}
where $\ell_{1}=\frac{1}{N} \sum_{j=1}^{K} ||{\hat{\mathbf{S}}^j}-{\mathbf{S}}||_1$  and $\ell_{fr}=\frac{1}{K} \sum_{j=1}^{K} ||\mathcal{F}({\hat{\mathbf{S}}^j})-\mathcal{F}({\mathbf{S}})||_1$. $\mathcal{F}(\cdot)$ represents 2D Fast Fourier Transform. The loss is uniformly applied to each sub-decoder with the same weight.

For classifier-training phase, Cross Entropy Loss is used to measure the difference between $\hat{{c}}$ and ground-truth degradation degree classification $\tilde{c}$\vspace{-0.5em}:
\begin{align}
\ell_{c}= \mathtt{CrossEntropy}(\tilde{c}, \hat{{c}})\label{eq:loss3}\vspace{-0.5em}
\end{align}

\section{Experiment}
\label{sec:exxperiment}

\subsection{Experimental Setup}
\label{sec:dataset-detail}
\paragraph{Dataset} We evaluate our method on the four datasets: GoPro~\cite{Nah2017deep} / HIDE~\cite{Shen2019human} / RealBlur-R / RealBlur-J~\cite{Rim2020real}. GoPro dataset is also used for the experiments in Sec.~\ref{sec:analysis}. More details will be shown in the supplementary material\vspace{-0.7em}. 


\paragraph{Model Configuration} We build 3 reversible decoder models based on the trained encoder from NAFNet~\cite{Chen2022simple} and UFPNet (\textbf{default})~\cite{fang2023UFPNet}: RevD-S (2 sub-decoders), RevD-B (4 sub-decoders) and RevD-L (8 sub-decoders). Because of the various distributions of different datasets, the class number varies. The patches from GoPro~\cite{Nah2017deep} and RealBlur-J~\cite{Rim2020real} are divided into 6 classes. Besides, there are 8 classes for RealBlur-R~\cite{Rim2020real}. With the degradation degree classifier, AdaRevD-B and AdaRevD-L apply adaptive patch exiting based on various incremental predictions of different sub-decoders from the train set. The threshold $\tau$ is configured to 0.05 for the early-exit signal ${E}_c$. More details will be shown in the supplementary material\vspace{-0.7em}.

\begin{table}[t]
\begin{center}
\caption{Comparison on GoPro~\cite{Nah2017deep}, HIDE~\cite{Shen2019human} and RealBlur~\cite{Rim2020real} datasets for setting $\mathcal{A}$. }
\label{tab:trained on GoPro}
\vspace{-0.7em}
\renewcommand\arraystretch{0.7}
\resizebox{1\linewidth}{!}{
\begin{tabular}{l c | c | c | c }
\toprule[0.15em]
 & \textbf{GoPro} & \textbf{HIDE} & \textbf{RealBlur-R} & \textbf{\textbf{RealBlur-J}} \\
 \textbf{Method} & PSNR~\colorbox{color4}{SSIM} & PSNR~\colorbox{color4}{SSIM} & PSNR~\colorbox{color4}{SSIM} & PSNR~\colorbox{color4}{SSIM}\\
\midrule[0.15em]
DeepDeblur~\cite{Nah2017deep}  & 29.08 \colorbox{color4}{0.914} & 25.73 \colorbox{color4}{0.874}  &  32.51 \colorbox{color4}{0.841}  &  27.87 \colorbox{color4}{0.827} \\
SRN~\cite{Tao2018scale}  & 30.26 \colorbox{color4}{0.934} & 28.36 \colorbox{color4}{0.915} & 35.66 \colorbox{color4}{0.947} &  28.56  \colorbox{color4}{0.867} \\
DMPHN~\cite{Zhang2019DMPHN} & 31.20 \colorbox{color4}{0.940}  & 29.09 \colorbox{color4}{0.924} &  35.70 \colorbox{color4}{0.948} & 28.42 \colorbox{color4}{0.860} \\
DBGAN~\cite{Zhang2020deblurring} & 31.10 \colorbox{color4}{0.942} & 28.94 \colorbox{color4}{0.915}  & 33.78 \colorbox{color4}{0.909}     & 24.93 \colorbox{color4}{0.745} \\
MT-RNN~\cite{Park2020multi} & 31.15 \colorbox{color4}{0.945} & 29.15 \colorbox{color4}{0.918}   & 35.79 \colorbox{color4}{0.951}     & 28.44 \colorbox{color4}{0.862}\\
MPRNet~\cite{Zamir2021multi} & 32.66 \colorbox{color4}{0.959} & {30.96} \colorbox{color4}{0.939} & {35.99} \colorbox{color4}{0.952} & {28.70} \colorbox{color4}{0.873}\\
HINet~\cite{Chen2021hinet} & 32.71 \colorbox{color4}{0.959} & 30.32  \colorbox{color4}{0.932} & - & -\\
MIMO-UNet+~\cite{Cho2021rethinking} & 32.45 \colorbox{color4}{0.957} & 29.99 \colorbox{color4}{0.930} & 35.54 \colorbox{color4}{0.947} & 27.63 \colorbox{color4}{0.837}\\
Whang~\cite{Whang_2022_CVPR} & 33.23 \colorbox{color4}{0.963} & - & - & -\\
Uformer~\cite{Wang2022uformer} & 33.06 \colorbox{color4}{0.967} & 30.90 \colorbox{color4}{0.953} & 36.19 \colorbox{color4}{0.956} & 29.09 \colorbox{color4}{0.886}\\
NAFNet64~\cite{Chen2022simple} & 33.69 \colorbox{color4}{0.967} &31.32  \colorbox{color4}{0.943} & 35.84 \colorbox{color4}{0.952} & 27.94 \colorbox{color4}{0.854} \\
Stripformer~\cite{Tsai2022Stripformer} & 33.08 \colorbox{color4}{0.962} &31.03  \colorbox{color4}{0.940} & - & -  \\
Restormer~\cite{Zamir2021restormer} & {32.92} \colorbox{color4}{{0.961}} & {31.22} \colorbox{color4}{{0.942}} & {36.19} \colorbox{color4}{{0.957}} & {28.96} \colorbox{color4}{{0.879}}\\
DeepRFT+~\cite{XintianMao2023DeepRFT} & 33.52 \colorbox{color4}{0.965} &31.66  \colorbox{color4}{0.946} & 36.11 \colorbox{color4}{0.955} & 28.90 \colorbox{color4}{0.881} \\
FFTformer~\cite{kong2023fftformer} & \underline{34.21} \colorbox{color4}{0.968} &31.62  \colorbox{color4}{0.946} & - & -  \\
UFPNet~\cite{fang2023UFPNet} & 34.06 \colorbox{color4}{0.968} & \underline{31.74} \colorbox{color4}{0.947} & \underline{36.25} \colorbox{color4}{0.953} & \underline{29.87} \colorbox{color4}{0.884}  \\
MRLPFNet~\cite{dong2023MRLPFNet} & 34.01 \colorbox{color4}{0.968} & 31.63 \colorbox{color4}{0.947} & - & -  \\
\arrayrulecolor{black!30}\midrule
RevD-S(UFPNet) & \textbf{34.35} \colorbox{color4}{0.970} & \textbf{32.08} \colorbox{color4}{0.950} & \textbf{36.56} \colorbox{color4}{0.957} & \textbf{30.09} \colorbox{color4}{0.892}\\
\arrayrulecolor{black!30}\midrule
RevD-B(NAFNet) & 34.10 \colorbox{color4}{0.969} & 31.86 \colorbox{color4}{0.948} & 36.04 \colorbox{color4}{0.952} & 28.46 \colorbox{color4}{0.863}\\
RevD-B(UFPNet) & \textbf{34.51} \colorbox{color4}{0.971} & \textbf{32.27} \colorbox{color4}{0.952} & \textbf{36.58} \colorbox{color4}{0.957} & \textbf{30.12} \colorbox{color4}{0.893}\\
AdaRevD-B & \textbf{34.50} \colorbox{color4}{0.971} & \textbf{32.26} \colorbox{color4}{0.952} & \textbf{36.56} \colorbox{color4}{0.957} & \textbf{30.12} \colorbox{color4}{0.894}\\
\arrayrulecolor{black!30}\midrule
RevD-L(NAFNet) & 34.18 \colorbox{color4}{0.970} & 31.93 \colorbox{color4}{0.948} & 36.03 \colorbox{color4}{0.952} & 28.46 \colorbox{color4}{0.863}\\
RevD-L(UFPNet) & \textbf{34.64} \colorbox{color4}{0.972} & \textbf{32.37} \colorbox{color4}{0.953} & \textbf{36.60} \colorbox{color4}{0.958} & \textbf{30.14} \colorbox{color4}{0.895}\\
AdaRevD-L & \textbf{34.60} \colorbox{color4}{0.972} & \textbf{32.35} \colorbox{color4}{0.953} & \textbf{36.53} \colorbox{color4}{0.957} & \textbf{30.12} \colorbox{color4}{0.894}\\
\arrayrulecolor{black}\bottomrule[0.1em]
\vspace{-3em}
\end{tabular}}
\end{center}
\end{table}

\begin{table}[t]
\footnotesize
\begin{center}
\caption{Comparison on RealBlur~\cite{Rim2020real} for setting $\mathcal{B}$.}
\vspace{-0.8em}
\label{tab:trained on RealBlur}
\renewcommand\arraystretch{0.7}
\setlength{\tabcolsep}{1.9pt}
\resizebox{1\linewidth}{!}{
\begin{tabular}{l c | c | c | c  }
\toprule[0.15em]
 & \textbf{RealBlur-R} & \textbf{RealBlur-J} & \textbf{Average} & Memory \\
 \textbf{Method} & PSNR~\colorbox{color4}{SSIM} & PSNR~\colorbox{color4}{SSIM} & PSNR~\colorbox{color4}{SSIM} &  (MB) \\
 \midrule
DeblurGAN-v2~\cite{Kupyn2019deblurgan} & 36.44 \colorbox{color4}{0.935} & 29.69 \colorbox{color4}{0.870} & 33.07 \colorbox{color4}{0.903} & -\\
SRN~\cite{Tao2018scale} &  38.65 \colorbox{color4}{0.965} & 31.38 \colorbox{color4}{0.909}& 35.02 \colorbox{color4}{0.937} & -\\
MPRNet~\cite{Zamir2021multi} & {39.31} \colorbox{color4}{0.972} & 31.76 \colorbox{color4}{0.922} & 35.54 \colorbox{color4}{0.947} & 6,294 \\
MAXIM~\cite{Tu2022maxim} & {39.45} \colorbox{color4}{0.962} & 32.84 \colorbox{color4}{0.935} & 36.15 \colorbox{color4}{0.949} & - \\
Stripformer~\cite{Tsai2022Stripformer} & 39.84 \colorbox{color4}{0.974} & 32.48 \colorbox{color4}{0.929} & 36.16 \colorbox{color4}{0.952}  & 3,449  \\
DeepRFT+~\cite{XintianMao2023DeepRFT} & 40.01 \colorbox{color4}{0.973} & 32.63 \colorbox{color4}{0.933} & 36.32 \colorbox{color4}{0.953}  & 3,569 \\
FFTformer~\cite{kong2023fftformer} & 40.11 \colorbox{color4}{0.975} & 32.62 \colorbox{color4}{0.933} & 36.37 \colorbox{color4}{0.954}  & 19,800  \\
UFPNet~\cite{fang2023UFPNet} & 40.61 \colorbox{color4}{0.974} & \underline{33.35} \colorbox{color4}{0.934} & 36.98 \colorbox{color4}{0.954}  & 8,164 \\
MRLPFNet~\cite{dong2023MRLPFNet} & \underline{40.92} \colorbox{color4}{0.975} & 33.19 \colorbox{color4}{0.936}  & \underline{37.06} \colorbox{color4}{0.956}  & -\\
\arrayrulecolor{black!30}\midrule
RevD-B  & \textbf{41.10} \colorbox{color4}{\textbf{0.978}} & \textbf{33.84} \colorbox{color4}{\textbf{0.943}} & \textbf{37.47} \colorbox{color4}{\textbf{0.961}} & 3,386\\
AdaRevD-B  & \textbf{41.09} \colorbox{color4}{\textbf{0.978}} & \textbf{33.84} \colorbox{color4}{\textbf{0.943}}& \textbf{37.47} \colorbox{color4}{\textbf{0.961}} & -\\
\arrayrulecolor{black!30}\midrule
RevD-L  & \textbf{41.22} \colorbox{color4}{\textbf{0.979}} & \textbf{33.99} \colorbox{color4}{\textbf{0.944}}& \textbf{37.61} \colorbox{color4}{\textbf{0.962}} & 5,087\\
AdaRevD-L  & \textbf{41.19} \colorbox{color4}{\textbf{0.979}} & \textbf{33.96} \colorbox{color4}{\textbf{0.944}}& \textbf{37.58} \colorbox{color4}{\textbf{0.962}} & -\\

\arrayrulecolor{black}\bottomrule[0.15em]
\end{tabular}}
\end{center}\vspace{-2.3em}
\end{table}

\paragraph{Implementation Details and Evaluation Metric} 
We adopt the training strategy from NAFNet~\cite{Chen2022simple} unless otherwise specified. \textit{I.e.}, the network training hyperparameters (and the default values) are data augmentation (horizontal and vertical flips), optimizer Adam ($\beta_1=0.9$, $\beta_2=0.9$, weight decay 1$\times$10$^{-3}$), initial learning rate (1$\times$10$^{-3}$). The learning rate is steadily decreased to 1$\times$10$^{-7}$. RevD is trained with patch size 256$\times$256 and batch size 16 for 200K iterations with ema decay 0.999. For AdaRevD, an additional classifier is trained based on the pre-trained encoder with patch size 384$\times$384 and batch size 16 for 10K iterations. Due to the spatially-variant kernel and statistics distribution shifts between training and testing~\cite{chu2021tlc}, we utilize a step of 352  sliding window strategy with the window size of 384$\times$384, and an overlap size of 32 for testing.

Performances in terms of PSNR and SSIM over all testing sets are calculated using official algorithms. Besides, GPU memory is calculated by training a single patch size 256$\times$256 image with NVIDIA GeForce RTX 3090 GPU.

\subsection{Main Results}
\paragraph{Setting $\mathcal{A.}$}
Our models are trained on 2,103 image pairs from GoPro~\cite{Nah2017deep}, and compared with other SOTA methods through the test set of GoPro~\cite{Nah2017deep} , HIDE~\cite{Shen2019human} and RealBlur~\cite{Rim2020real}. As shown in Table~\ref{tab:trained on GoPro}, AdaRevD outperforms the other models in both PSNR and SSIM on the GoPro and HIDE test set. Besides, AdaRevD-B obtains robust results on other datasets. For example, AdaRevD-B achieves 36.56 / 30.12~dB on RealBlur-R / J test set, 0.31 / 0.25~dB higher than UFPNet~\cite{fang2023UFPNet}. Figure~\ref{fig:GoPro} shows the visual comparison on GoPro test set. Our approach extends the capabilities of the well-trained encoder by enhancing its capacity. In comparison to other SOTA methods, ours effectively minimizes a greater amount of the blur pattern\vspace{-1em}.

\begin{figure}[t]
\captionsetup[subfigure]{justification=centering, labelformat=empty}
 \centering
	\begin{subfigure}{0.241\linewidth}
		\includegraphics[width=0.99\linewidth]{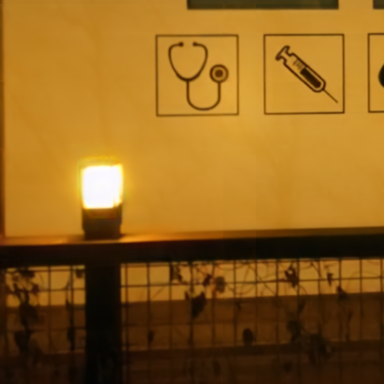}
	\end{subfigure}
	\begin{subfigure}{0.241\linewidth}
		\includegraphics[width=0.99\linewidth]{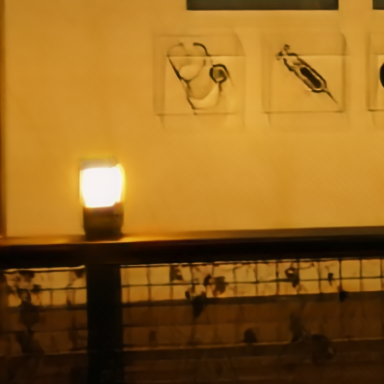}
	\end{subfigure}
	\begin{subfigure}{0.241\linewidth}
		\includegraphics[width=0.99\linewidth]{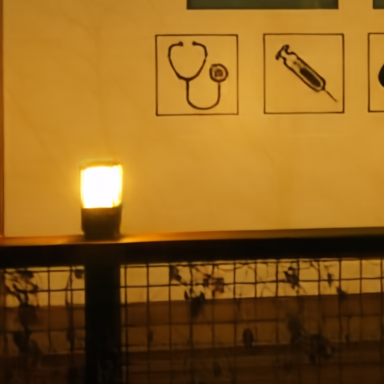}
	\end{subfigure}
	\begin{subfigure}{0.241\linewidth}
		\includegraphics[width=0.99\linewidth]{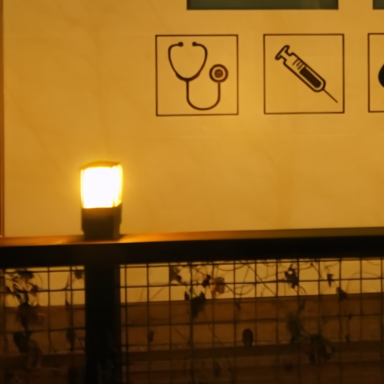}
	\end{subfigure}
	\quad
	\begin{subfigure}{0.241\linewidth}
		\includegraphics[width=0.99\linewidth]{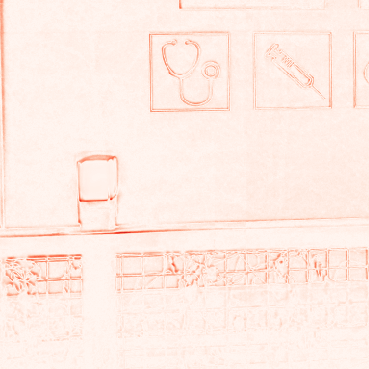}
        \caption{DeepRFT+~\cite{XintianMao2023DeepRFT}}
	\end{subfigure}
	\begin{subfigure}{0.241\linewidth}
		\includegraphics[width=0.99\linewidth]{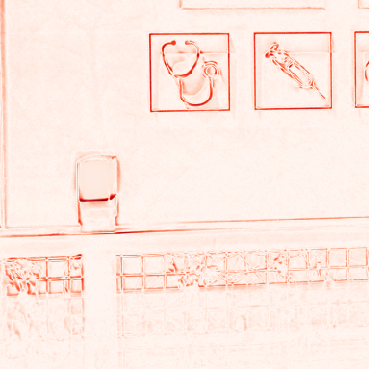}
        \caption{Stripformer~\cite{Tsai2022Stripformer}}
	\end{subfigure}
	\begin{subfigure}{0.241\linewidth}
		\includegraphics[width=0.99\linewidth]{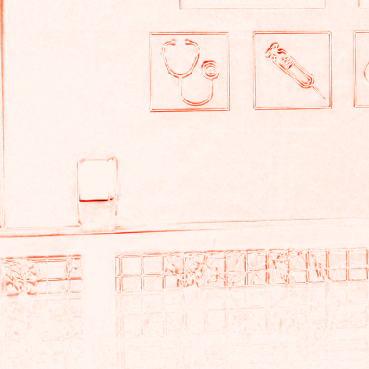}
          \caption{UFPNet~\cite{fang2023UFPNet}}
	\end{subfigure}
	\begin{subfigure}{0.241\linewidth}
		\includegraphics[width=0.99\linewidth]{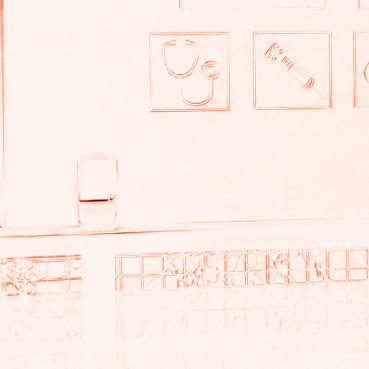}
        \caption{Ours}
	\end{subfigure}

 \vspace{-0.5em}
\caption{Examples on the RealBlur-J test dataset.}
\label{fig:RealBlur-J}
\vspace{-1.5em}
\end{figure}

\paragraph{Setting $\mathcal{B.}$}
Our models are trained and tested on RealBlur-J / RealBlur-R ~\cite{Rim2020real} respectively. As seen in Table~\ref{tab:trained on RealBlur}, AdaRevD-B / L achieves 33.84 / 33.96~dB on RealBlur-J test set, 0.49 / 0.61~dB higher than UFPNet~\cite{fang2023UFPNet}. Figure~\ref{fig:RealBlur-J} shows the visual comparison on RealBlur-J test set. In comparison with alternative approaches, our method excels in producing cleaner image restorations. Specifically, when applied to RealBlur-R, AdaRevD-L achieves a superior outcome (41.19~dB) compared to UFPNet (40.61~dB).

 \subsection{Analysis and Discussions}
\label{sec:analysis}

\begin{figure*}[t]
\begin{center}
    \includegraphics[width=0.85\linewidth]{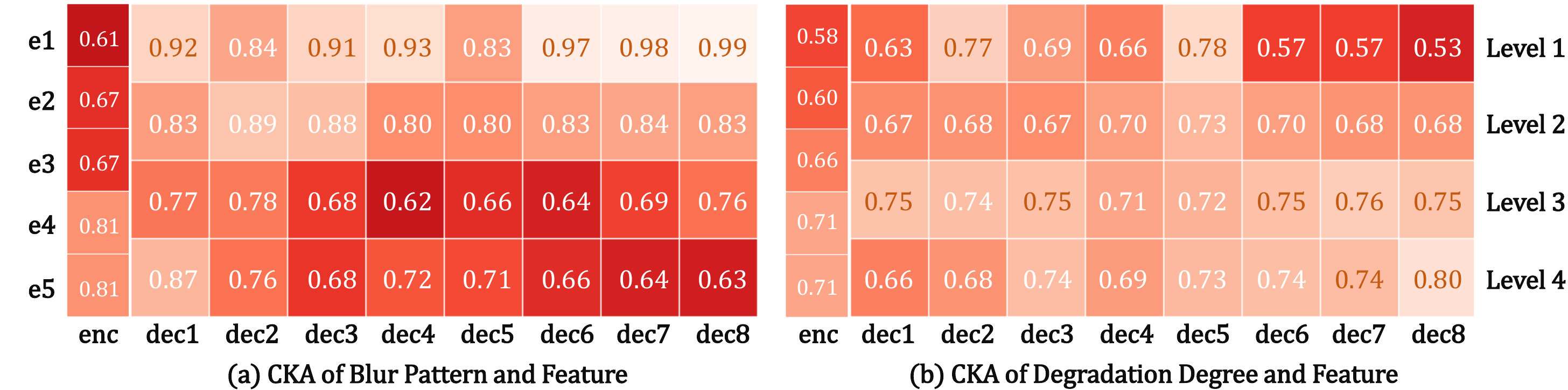}
\end{center}
\vspace{-1.8em}
\caption{CKA similarities~\cite{kornblith2019CKA} of features and blur pattern / degradation degree for different levels and decoders. The blur pattern is obtained by subtracting the blur image from the ground-truth sharp image. The degradation degree comes from the feature after the NAFBlock in the classifier. } 
\label{fig:cka}
\vspace{-1.5em}
\end{figure*}

\subsubsection{Effect of Reversible Image Deblurring}
\paragraph{Memory Consumption}
In AdaRevD, we propose to enhance the size of a single-stage model with the reversible sub-decoders. As indicated in Table~\ref{tab:sub-decoders-performance}, the cost of training memory is more friendly for reversible architecture when the size of the model is increasing. The memory consumption of a non-reversible architecture increases at a much faster rate than that of a reversible architecture as the number of sub-decoders rises\vspace{-1em}.
\paragraph{FourierBlock}
As shown in Table~\ref{tab:Ablation-level-module}, compared with the original NAFBlock~\cite{Chen2022simple}, FourierConv (RFFT2d-1$\times$1Conv-GELU-1$\times$1Conv-IRFFT2d) can significantly boost
the model's performance from 34.15~dB to 34.51~dB\vspace{-1em}. 
\paragraph{Retrain on SOTA Methods}
Since our AdaRevD retrains on the encoder of SOTA methods, we also train original SOTA methods with more iterations to see what the results are. As shown in Table~\ref{tab:Ablation-strategy}, if we allow the well-trained model to undergo additional iterations using the same training strategy, the improvement is marginal for NAFNet~\cite{Chen2022simple}. We also perform experiments to investigate the impact of freezing the encoder, and the results are reported in the Table~\ref{tab:Ablation-strategy}. For UFPNet~\cite{fang2023UFPNet}, keeping both encoder and decoder training for more iterations would even cause a decrease from 34.09~dB (freezing encoder) to 33.71~dB. To conserve additional GPU memory, AdaRevD investigates the construction of a multi-decoder architecture by leveraging a frozen encoder from from other SOTA methods\vspace{-1em}.

\begin{table} [t]
    \centering
\caption{Performance of different sub-decoders. Dec-Idx is the sub-decoder number. Non-Rev means reversible mode is not used.}
\vspace{-0.5em}
\label{tab:sub-decoders-performance}
\renewcommand\arraystretch{0.9}
\setlength{\tabcolsep}{1.9pt}
\resizebox{1.0\linewidth}{!}{
    \begin{tabular}{l|c|cccccccc}
    \toprule[0.15em]
         & Dec-Idx& 1 & 2 & 3 & 4 & 5 & 6 & 7 & 8\\ 
          \midrule
          RevD-S& \multirow{2}{*}{PSNR} & 34.142 & 34.348 & -  & - & - & - & - & -\\ 
          RevD-B& \multirow{2}{*}{(dB)} & 34.074 & 34.313 & 34.459  & 34.515 & - & - & - & -\\ 
         RevD-L&   & 34.020 & 34.287 & 34.376  & 34.481 & 34.577 & 34.620 & 34.635 & 34.640\\ 
         \arrayrulecolor{black!30}\midrule
        Reversible&  Memory & 1,980 & 2,537 & 2,960 & 3,386 & 3,810 & 4,233 & 4,660 & 5,087\\ 
         Non-Rev& (MB) & 1,980  & 3,260 & 4,548 & 5,834 & 7,111 & 8,398 & 9,676 & 10,960\\ 
    \arrayrulecolor{black}\bottomrule[0.15em]
    \end{tabular}}\vspace{-0.3em}
\end{table}

\begin{table}[t]
    \centering
\caption{Ablation studies. Num-Dec is sub-decoder's number. $N$* means using a decoder with $N$ blocks for each level.}
\vspace{-0.5em}
\label{tab:Ablation-level-module}
\renewcommand\arraystretch{0.7}
\setlength{\tabcolsep}{1.9pt}
\resizebox{0.9\linewidth}{!}{
    \begin{tabular}{cccc|cc}
    \toprule[0.15em]
        FuseBlock & \multicolumn{2}{c}{FourierBlock}   & Num-Dec & Memory & PSNR\\
        \arrayrulecolor{black!30}\cmidrule(lr){2-3}
       ~  & NAFBlock & Fourier &~ &(MB) & (dB)\\
       \midrule
       $\times$ & \checkmark   & $\times$ & 4  &1,864 & 34.03 \\
      \checkmark & \checkmark  &  $\times$ & 4 &1,974 & 34.15 \\
      $\times$ & \checkmark  &\checkmark & 4  &3,258 & 34.24 \\
       \checkmark  & \checkmark  &\checkmark & 2*  &3,124 & 34.15 \\
       \checkmark  & \checkmark  &\checkmark & 4  &3,386 & 34.51 \\
       \arrayrulecolor{black!30}\midrule
       \checkmark  & \checkmark  &\checkmark & 4*  &5,387 & 34.49 \\
       \checkmark  & \checkmark  &\checkmark & 8  &5,087 & 34.64 \\
    \arrayrulecolor{black}\bottomrule[0.15em]
    \end{tabular}}\vspace{-0.8em}
\end{table}

\paragraph{Disentanglement of Degradation Representation and Blur Pattern}
\label{sec:disentangle}
We use Centered Kernel Alignment (CKA) similarity metric~\cite{kornblith2019CKA} to measure the similarity between learned features in RevD-L and blur pattern (residual between the blur image and its sharp counterpart), as well as the similarity between learned features and degradation degree (output feature of NAFBlock in Classifier), shown as the $2$nd $\sim$ $9$th columns in Fig.~\ref{fig:cka}(a) and Fig.~\ref{fig:cka}(b). Besides, we also compute the similarity between encoder features and the blur pattern / degradation degree, shown in the first columns. The encoder features (enc, $\mathbf{e}_1\sim\mathbf{e}_5$) gradually learn mid-level degradation representation which is mixed with the information of low-level blur pattern and high-level degradation degree. 

For features of Level 1$\sim$Level 4 in 8 sub-decoders, we observe two intriguing phenomenons: \textbf{(1)} One sub-decoder (dec1) cannot decode the accurate blur pattern from degradation representations (see the $2$nd column in Fig.~\ref{fig:cka}(a)). This may be due to the reason that encoder features in an UNet architecture inevitably bring information unrelated to the target details to decoder features via operations such as skip connections. Details related to the input blur image itself are likely to be superimposed on the decoder features. Thus, the target blur pattern (residual between the blur and sharp images) cannot be decoded precisely. \textbf{(2)} reversible sub-decoders enable the ability to gradually disentangle low-level blur pattern and high-level degradation degree (see the $2$nd to the last column in Fig.~\ref{fig:cka}). If we observe the $2$nd to the last column in Fig.~\ref{fig:cka} row-by-row, it is obvious that more accurate low-level blur patterns are learned from Level modules connected to the output (\emph{e.g.}, Level 1 in Fig.~\ref{fig:cka}(a)), and more high-level degradation degree knowledge is gathered from Level modules connected to the encoder (\emph{e.g.}, Level 4 in Fig.~\ref{fig:cka}(b))\vspace{-1.em}.

\begin{table}[t] 
\footnotesize
    \centering
\caption{Comparison for continued training on SOTA methods. For UFPNet, since we freeze the kernel prior model, the training memory reduces from 8,164~MB (in Table~\ref{tab:trained on RealBlur}) to 4,311~MB\vspace{-0.5em}. }

\label{tab:Ablation-strategy}
\renewcommand\arraystretch{1.0}
\setlength{\tabcolsep}{1.9pt}
    \begin{tabular}{c|cc|cc}
    \toprule[0.15em]

       Model & \multicolumn{2}{c}{NAFNet} & \multicolumn{2}{c}{UFPNet} \\
       \midrule
       Encoder trainable & \checkmark & $\times$ & \checkmark & $\times$ \\
       Memory (MB)  & ~3,394~  & ~1,160~~ & ~4,311~ & ~1,617~ \\
       PSNR (dB)  & 33.75  & 33.70 & 33.71 & 34.09 \\
    \arrayrulecolor{black}\bottomrule[0.15em]
    \end{tabular}\vspace{-1.em}
\end{table}
\begin{table}[h] 
    \centering
\caption{Ablation for classifier. Reg and Cls means regressor and classifier respectively. Reg($ \mathbf{d}_4^{k-1} $) indicates that the regressor takes $ \mathbf{d}_4^{k-1}$ as input and output the incremental prediction for the $k$th sub-decoder. $\tau$ is the threshold for the regressor and the early-exit signal ($E_c$ in Eq.~\ref{eq:Ec}). The numbers for ${E}_1\sim{E}_6$ mean the specific sub-decoder to exit for a patch. Acc is the accuracy of the classifier. L1 is the L1 distance between the regressor's output and the real increment of each sub-decoder. D-Rate means the utilization rate of the sub-decoders.}
\vspace{-0.5em}
\label{tab:Ablation-classifier}
\renewcommand\arraystretch{0.9}
\setlength{\tabcolsep}{1.9pt}
\resizebox{1\linewidth}{!}{
    \begin{tabular}{c|c|cccccc|ccc}
    \toprule[0.15em]
        
       ~ & $\tau$ & ${E}_{1}$ & ${E}_{2}$ & ${E}_{3}$ & ${E}_{4}$ & ${E}_{5}$ & ${E}_{6}$ & Acc / L1 & PSNR & D-Rate\\
       \midrule
       \multirow{2}{*}{RevD-B} & - & 4 &4 & 4 & 4 & 4 & 4 & - & 34.515 & 100.0\% \\
       & - & 3 & 3 & 3 & 3 & 3 & 3 & - & 34.459 & 75.0\% \\
       \arrayrulecolor{black!30}\midrule
      \multirow{2}{*}{Reg($ \mathbf{d}_4^{k-1} $)} & 0.05 & - & - & - & - & - & -  & \multirow{2}{*}{0.071} & 34.497 & 82.5\% \\
       & 0.10 & - & - & - & - & - & -  &  & 34.457 & 65.9\% \\
       \arrayrulecolor{black!30}\midrule
       \multirow{2}{*}{Cls($ \mathbf{e}_5 $)} & 0.05 & 4 & 4 & 3 & 2 & 1 & 1  & \multirow{2}{*}{84.43\%} & 34.502 & 85.8\% \\
        & 0.10 & 4 & 3 & 2 & 2 & 1 & 1  &  & 34.437 & 62.1\% \\
        \arrayrulecolor{black!30}\midrule
       \multirow{2}{*}{Cls($\mathbf{e}_4$)} & 0.05 & 4 & 4 & 3 & 2 & 1 & 1  & \multirow{2}{*}{89.84\%} & 34.501 & 84.3\% \\
        & 0.10 & 4 & 3 & 2 & 2 & 1 & 1  &  & 34.436 & 60.6\% \\

    \arrayrulecolor{black}\bottomrule[0.15em]
    \end{tabular}}\vspace{-1.em}
\end{table}
\subsubsection{Effect of Adaptive Patch Exiting}
Patch-exiting has been proposed in super-resolution tasks, whose modules usually learn features with the same scale~\cite{wang2022adaptiveSR}. Thus, in image deblurring, no prior work has been attempted to design a patch-exiting method.  Different from ClassSR~\cite{kong2021classsr} and APE-SR~\cite{wang2022adaptiveSR}, AdaRevD builds an adaptive patch exiting model based on multiple sub-decoders and a simple classifier. ClassSR~\cite{kong2021classsr} has to train three different models to solve various types of low-resolution patches, which raises the training burden. AdaRevD employs multiple reversible sub-decoders to address diverse blur patches, resulting in significant GPU memory savings and providing flexibility for patch exiting. APE-SR~\cite{wang2022adaptiveSR} introduces a regressor to estimate the incremental prediction, while restoration models always tend to perform better on the train set than the test set. Different from APE-SR~\cite{wang2022adaptiveSR}, we conduct experiments which use a regressor to predict the increment directly, while the the regressor seems not perform well on the test set. As can be seen in Table~\ref{tab:Ablation-classifier}, the L1 distance of the regressor is 0.071~dB, while the improvement in the latter sub-decoders shown in Table~\ref{tab:sub-decoders-performance} would be smaller than 0.071~dB. In addition, the regressor method has to estimate the incremental prediction in each sub-decoder while the classifier only needs to predict the degradation degree once. Table~\ref{tab:Ablation-classifier} indicates that the classifier using $\mathbf{e}_4$ as input achieves higher accuracy (89.84\%) than $\mathbf{e}_5$ (84.43\%). As shown in Table~\ref{tab:Ablation-classifier}, AdaRevD maintains performance with a minimal drop (0.014 dB) while utilizing fewer computing resources (84.3\%)\vspace{-0.5em}.

 \section{Conclusion}
\label{sec:conclusion}
In this work, we propose an adaptive patch exiting architecture with multiple reversible sub-decoders to push the limit of SOTA deblurring networks (\emph{e.g.} NAFNet and UFPNet). Reversible sub-decoders help exploring the well trained model's insufficient decoding capability while maintaining low training memory. Experiments indicate that the reversible structure is able to decode the better blur pattern from degradation representations. Since blur image patches always have different degradation degrees due to the spatially-variant motion blur kernel, we introduce a classifier to predict the degradation degree of the blur patch. Therefore, AdaRevD is able to let the patches exit in the appropriate sub-decoder by the degradation degree with almost no loss of performance. Extensive experiments and visualizations on various image deblurring datasets demonstrate the superiority of our method. 
  \section{Acknowledgement}
This work was supported by the National Natural Science Foundation of China (Grant No. 62101191, 61975056), Shanghai Natural Science Foundation (Grant No. 21ZR1420800), and the Science and Technology Commission of Shanghai Municipality (Grant No. 22DZ2229004).
\clearpage

\newpage
{
    \small
    \bibliographystyle{ieeenat_fullname}
    \bibliography{main}
}
\clearpage
\setcounter{page}{1}
\maketitlesupplementary

\begin{table}[t]
\begin{center}
\caption{Summary of four public datasets.}
\label{tab:Datasets}
\renewcommand\arraystretch{1}
\setlength{\tabcolsep}{1.9pt}
\resizebox{0.7\linewidth}{!}{
\begin{tabular}{l | c | cc}
\toprule[0.15em]
\textbf{Dataset} & \textbf{Types} & \textbf{Train} & \textbf{Test} \\
\midrule[0.15em] 
GoPro~\cite{Nah2017deep}& synthetic & 2,103 & 1,111\\
\arrayrulecolor{black!30}\midrule
HIDE~\cite{Shen2019human} & synthetic & -  & 2,025\\
\arrayrulecolor{black!30}\midrule
RealBlur-R~\cite{Rim2020real} & real-world & 3,758   & 980\\
\arrayrulecolor{black!30}\midrule
RealBlur-J~\cite{Rim2020real} & real-world & 3,758  & 980\\
\arrayrulecolor{black}\bottomrule[0.15em]
\end{tabular}}
\end{center} \vspace{-1.5em}
\end{table}

\begin{table}[t] 
    \centering
\caption{Confusion Matrix of the Classifier on GoPro testset.  }
\label{tab:confusion-matrix}
\renewcommand\arraystretch{0.5}
\setlength{\tabcolsep}{1.9pt}
\resizebox{0.9\linewidth}{!}{
    \begin{tabular}{c|cccccc|c}
    \toprule[0.15em]
     \diagbox{gt}{pred} & $\le$ 20~dB & $\le$ 25~dB &  $\le$ 30~dB & $\le$ 35~dB& $\le$ 40~dB & $>$ 40~dB & total\\
    \midrule
    $\le$ 20~dB & \colorbox{color4}{394} & 61  & 0 & 0 & 0 & 0 & 455\\
    $\le$ 25~dB & 38 & \colorbox{color4}{2998} & 164 & 0 & 0 & 0 & 3200\\
    $\le$ 30~dB & 0  & 141 & \colorbox{color4}{3148} & 206 & 0 & 0 & 3495\\
    $\le$ 35~dB & 0 & 0 & 101 & \colorbox{color4}{1248} & 154 & 0 & 1503\\
    $\le$ 40~dB & 0 & 0 & 0 & 22 & \colorbox{color4}{192} & 3 & 217\\
    $>$ 40~dB & 0 & 0 & 0 & 0 & 13 & \colorbox{color4}{5} & 18\\
    \midrule
    total & 432 & 3200 & 3413 & 1476 & 359 & 8 & 8888\\
    \bottomrule[0.15em]
    \end{tabular}}
    \vspace{-1em}
\end{table}

\begin{table*}[t]
\begin{center}
\caption{The comparison involves the computational complexity of MACs (G) and the number of parameters (M), when the input size is 256 $\times$ 256. PSNR (dB) is calculated on GoPro test set.}
\label{tab:macs}
\renewcommand\arraystretch{1}
\setlength{\tabcolsep}{1.9pt}
\resizebox{0.995\linewidth}{!}{
\begin{tabular}{l | ccccc|ccc}
\toprule[0.15em]
\textbf{Method} & \textbf{MIMO-UNet++~\cite{Cho2021rethinking}} & \textbf{DeepRFT+~\cite{XintianMao2023DeepRFT}} & \textbf{Restormer~\cite{Zamir2021restormer}} & \textbf{NAFNet64~\cite{Chen2022simple}} & \textbf{UFPNet~\cite{fang2023UFPNet}} & \textbf{RevD-B(NAFNet)} & \textbf{RevD-B(UFPNet)} & \textbf{RevD-L(UFPNet)} \\
\midrule[0.15em] 
MACs (G) & 617 & 187 & 141 & 64 & 243 & 168 & 347 & 460 \\
Params (M) & 16.1 & 19.5 & 26.1 & 65.0 & 80.3 & 131.0 & 142.5 & 210.8 \\
\arrayrulecolor{black!30}\midrule
PSNR (dB) & 32.68 & 33.52 & 32.92 & 33.69 & 34.06 & 34.10 & 34.51 & 34.64 \\
\arrayrulecolor{black}\bottomrule[0.15em]
\end{tabular}}
\end{center} 
\end{table*}

\section{Details of Experiment}
\label{sec:supp_dataset}
\paragraph{Dataset} As shown in Sec.~4.1 in the main paper, we evaluate our method on the four datasets shown in Table~\ref{tab:Datasets} and report two groups of results:
\begin{itemize}
\item $\mathcal{A.}$ train on GoPro, test on GoPro~\cite{Nah2017deep} / HIDE~\cite{Shen2019human} / RealBlur-R / RealBlur-J~\cite{Rim2020real};
\item $\mathcal{B.}$ train and test on RealBlur-J / RealBlur-R respectively;\vspace{-0.6em}
\end{itemize}

\begin{table}[t]
\begin{center}
\caption{Improvement of diffent sub-decoders in RevD-B on GoPro dataset. The value in the $c$th row and the $j$th column is $\mathbf{O}_c^j$. The first \colorbox{color4}{$\mathbf{O}_c^{j-1}$} that $\mathbf{O}_c^{j}$ smaller than $\tau=0.05$ is highlighted in the table.}
\label{tab:RevD-B-GoPro}
\renewcommand\arraystretch{1}
\setlength{\tabcolsep}{1.9pt}
\resizebox{0.99\linewidth}{!}{
\begin{tabular}{c|c|c|c|c|c|c|c|c}
\toprule[0.15em]
& \multicolumn{4}{c}{\textbf{TrainSet}} &\multicolumn{4}{c}{\textbf{TestSet}}\\
\arrayrulecolor{black!30}\cmidrule(lr){2-5}\cmidrule(lr){6-9}
\textbf{Degree} & \textbf{dec1} & \textbf{dec2} & \textbf{dec3} & \textbf{dec4} & \textbf{dec1} & \textbf{dec2} & \textbf{dec3} & \textbf{dec4} \\
\midrule[0.15em] 
{$\le$20} & 11.134 &0.642 &0.351& \cellcolor{color4}0.178&  10.275 &  0.653 & 0.383 & \cellcolor{color4}0.170\\
\arrayrulecolor{black!30}\midrule
{20-25} & 10.959 &0.406 &0.211 & \cellcolor{color4}0.100
&  9.622 & 0.355 & 0.208 & \cellcolor{color4}0.093\\
\arrayrulecolor{black!30}\midrule
{25-30} & 9.184 &0.214 &\cellcolor{color4}0.105 &0.047
&  8.097 & 0.191 & \cellcolor{color4}0.100 & 0.045\\
\arrayrulecolor{black!30}\midrule
{30-35} & 6.215 &0.121 & \cellcolor{color4}0.050 &0.021
&  5.397 & 0.103 &\cellcolor{color4} 0.050 & 0.021\\
\arrayrulecolor{black!30}\midrule
{35-40} & 3.468 &\cellcolor{color4}0.079 &0.024 &0.011
 &  2.859 & \cellcolor{color4}0.073 & 0.014 & 0.010\\
\arrayrulecolor{black!30}\midrule
{$>$40} & \cellcolor{color4}2.380 &0.047 &0.016 &0.009
& \cellcolor{color4} 1.510 & 0.022 & 0.006 & 0.004\\
\arrayrulecolor{black}\bottomrule[0.15em]
\end{tabular} }
\end{center} 
\end{table}

\begin{table}[t]
\begin{center}
\caption{Improvement of diffent sub-decoders in RevD-B on RealBlur-J dataset.}
\label{tab:RevD-B-RealBlur-J}
\renewcommand\arraystretch{1}
\setlength{\tabcolsep}{1.9pt}
\resizebox{0.99\linewidth}{!}{
\begin{tabular}{c|c|c|c|c|c|c|c|c}
\toprule[0.15em]
& \multicolumn{4}{c}{\textbf{TrainSet}} &\multicolumn{4}{c}{\textbf{TestSet}}\\
\arrayrulecolor{black!30}\cmidrule(lr){2-5}\cmidrule(lr){6-9}
\textbf{Degree} & \textbf{dec1} & \textbf{dec2} & \textbf{dec3} & \textbf{dec4} & \textbf{dec1} & \textbf{dec2} & \textbf{dec3} & \textbf{dec4} \\
\midrule[0.15em] 
{$\le$20} & 11.855 &0.758 &0.441& \cellcolor{color4}0.201& 5.223  &  0.093 & \cellcolor{color4}0.089 & 0.041\\
\arrayrulecolor{black!30}\midrule
{20-25} & 11.236 &0.563 &0.336 & \cellcolor{color4}0.139
&  4.842 & 0.138 & \cellcolor{color4}0.092 & 0.041\\
\arrayrulecolor{black!30}\midrule
{25-30} &  10.041 & 0.390 & 0.214 & \cellcolor{color4}0.080
&  4.138 & 0.107 & \cellcolor{color4}0.066 & 0.024\\
\arrayrulecolor{black!30}\midrule
{30-35} & 8.406 &0.249 & \cellcolor{color4}0.115 &0.042
&  3.461 & \cellcolor{color4}0.067 & 0.036 & 0.010\\
\arrayrulecolor{black!30}\midrule
{35-40} & 7.101 &0.168 &\cellcolor{color4}0.064 &0.030
 &  2.537 & \cellcolor{color4}0.073 & 0.034 & 0.007\\
\arrayrulecolor{black!30}\midrule
{$>$40} & 5.525 &\cellcolor{color4}0.126 &0.044 &0.028
& \cellcolor{color4} 1.380 & -0.030 & 0.000 & 0.000\\
\arrayrulecolor{black}\bottomrule[0.15em]
\end{tabular} }
\end{center} 
\end{table}

\begin{table}[t]
\begin{center}
\caption{Improvement of diffent sub-decoders in RevD-B on RealBlur-R dataset.}
\label{tab:RevD-B-RealBlur-R}
\renewcommand\arraystretch{1}
\setlength{\tabcolsep}{1.9pt}
\resizebox{0.99\linewidth}{!}{
\begin{tabular}{c|c|c|c|c|c|c|c|c}
\toprule[0.15em]
& \multicolumn{4}{c}{\textbf{TrainSet}} &\multicolumn{4}{c}{\textbf{TestSet}}\\
\arrayrulecolor{black!30}\cmidrule(lr){2-5}\cmidrule(lr){6-9}
\textbf{Degree} & \textbf{dec1} & \textbf{dec2} & \textbf{dec3} & \textbf{dec4} & \textbf{dec1} & \textbf{dec2} & \textbf{dec3} & \textbf{dec4} \\
\midrule[0.15em] 
{$\le$20} &12.401&0.863&0.308 &\cellcolor{color4}0.098
&5.541 &0.152 & \cellcolor{color4}0.063 &0.029
\\
\arrayrulecolor{black!30}\midrule
{20-25} & 12.356 &0.781 &0.272 &\cellcolor{color4}0.081
&5.305 &0.182 &\cellcolor{color4}0.061 &0.022
\\
\arrayrulecolor{black!30}\midrule
{25-30} &  12.268 &0.702 &0.215 &\cellcolor{color4}0.060
&5.657 &0.155 &\cellcolor{color4}0.056 &0.017
\\
\arrayrulecolor{black!30}\midrule
{30-35} & 11.593 &0.588 &0.178 &\cellcolor{color4}0.0513
&  4.233 &\cellcolor{color4}0.139 &0.039 &0.012
\\
\arrayrulecolor{black!30}\midrule
{35-40} & 9.681 &0.390 &\cellcolor{color4}0.115 &0.042
 & 3.431 &\cellcolor{color4}0.133 &0.037 &0.016
\\
\arrayrulecolor{black!30}\midrule
{40-45} &7.423 &0.245 &\cellcolor{color4}0.0778 &0.032
 &  3.043 &\cellcolor{color4}0.111 &0.028 &0.012
\\
\arrayrulecolor{black!30}\midrule
{45-50} & 5.210 &\cellcolor{color4}0.136 &0.042 &0.021
 &  \cellcolor{color4}2.005 &0.042 &0.012 &0.008
\\
\arrayrulecolor{black!30}\midrule
{$>$50} & 3.227 &\cellcolor{color4}0.074 &0.018 &0.009

& \cellcolor{color4} 1.102 &0.014 &0.006 &0.006
\\
\arrayrulecolor{black}\bottomrule[0.15em]
\end{tabular} }
\end{center} 
\end{table}

\begin{figure*}[t]
\begin{center}
    \includegraphics[width=0.95\linewidth]{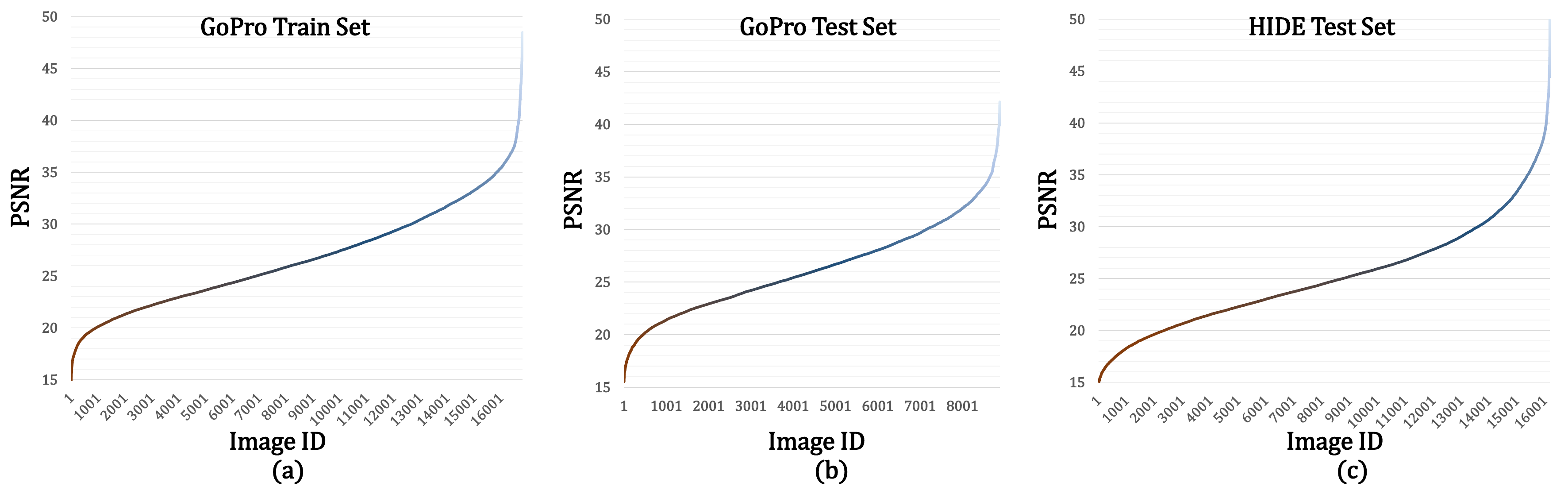}
\end{center}
\vspace{-1.em}
\caption{Distribution of GoPro~\cite{Nah2017deep} and HIDE~\cite{Shen2019human} Dataset. (a) The ranked PSNR curve of the image patches from GoPro train set; (b) The ranked PSNR curve of the image patches from GoPro test set; (c) The ranked PSNR curve of the image patches from HIDE test set.} 
\label{fig:distribution_gop_hide}
\vspace{-1.0em}
\end{figure*}

\begin{figure*}[t]
\begin{center}
    \includegraphics[width=0.7\linewidth]{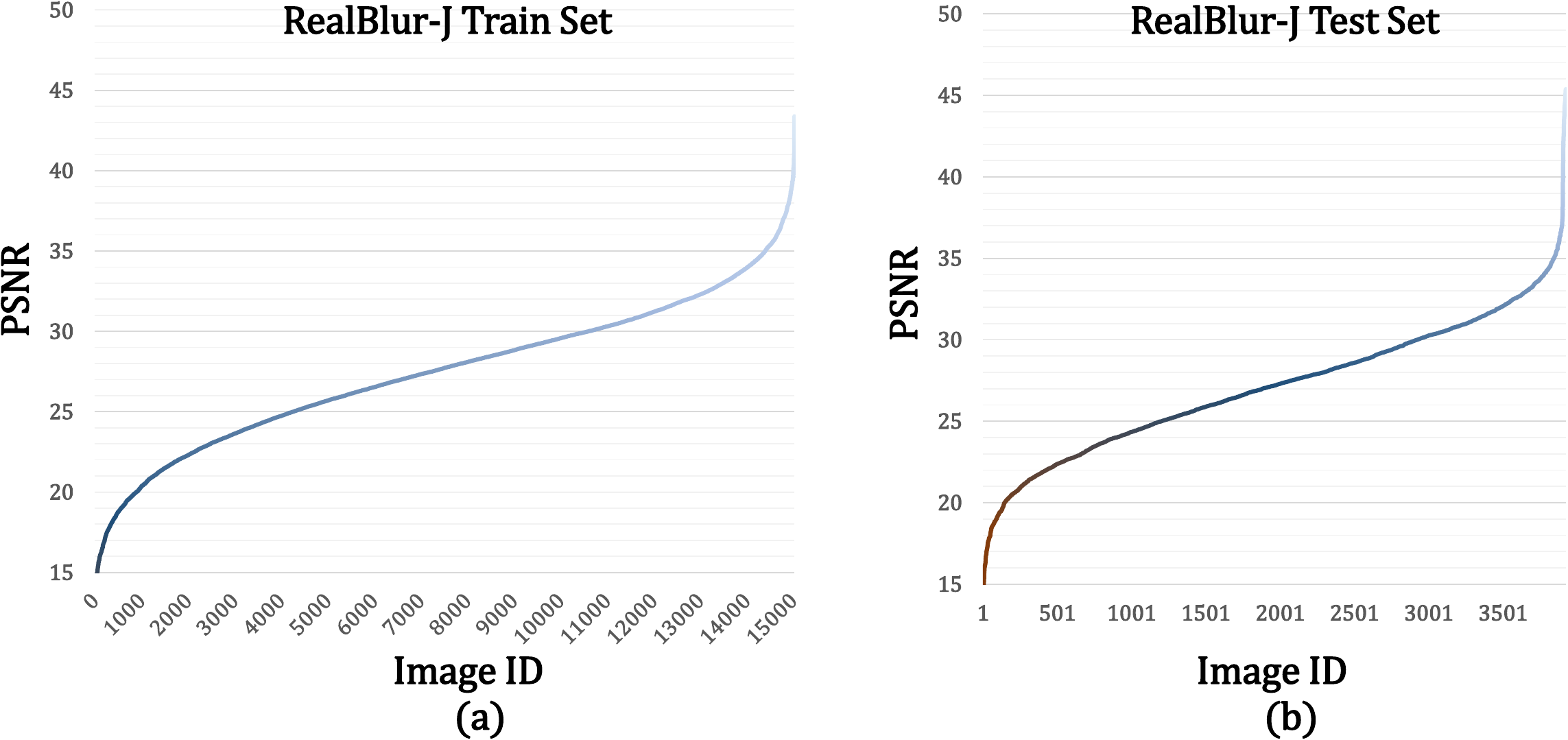}
\end{center}
\vspace{-1.em}
\caption{Distribution of RealBlur-J~\cite{Rim2020real} Dataset. (a) The ranked PSNR curve of the image patches from RealBlur-J train set; (b) The ranked PSNR curve of the image patches from RealBlur-J test set.} 
\label{fig:distribution_realblurj}
\vspace{-1.0em}
\end{figure*}

\begin{figure*}[t]
\begin{center}
    \includegraphics[width=0.7\linewidth]{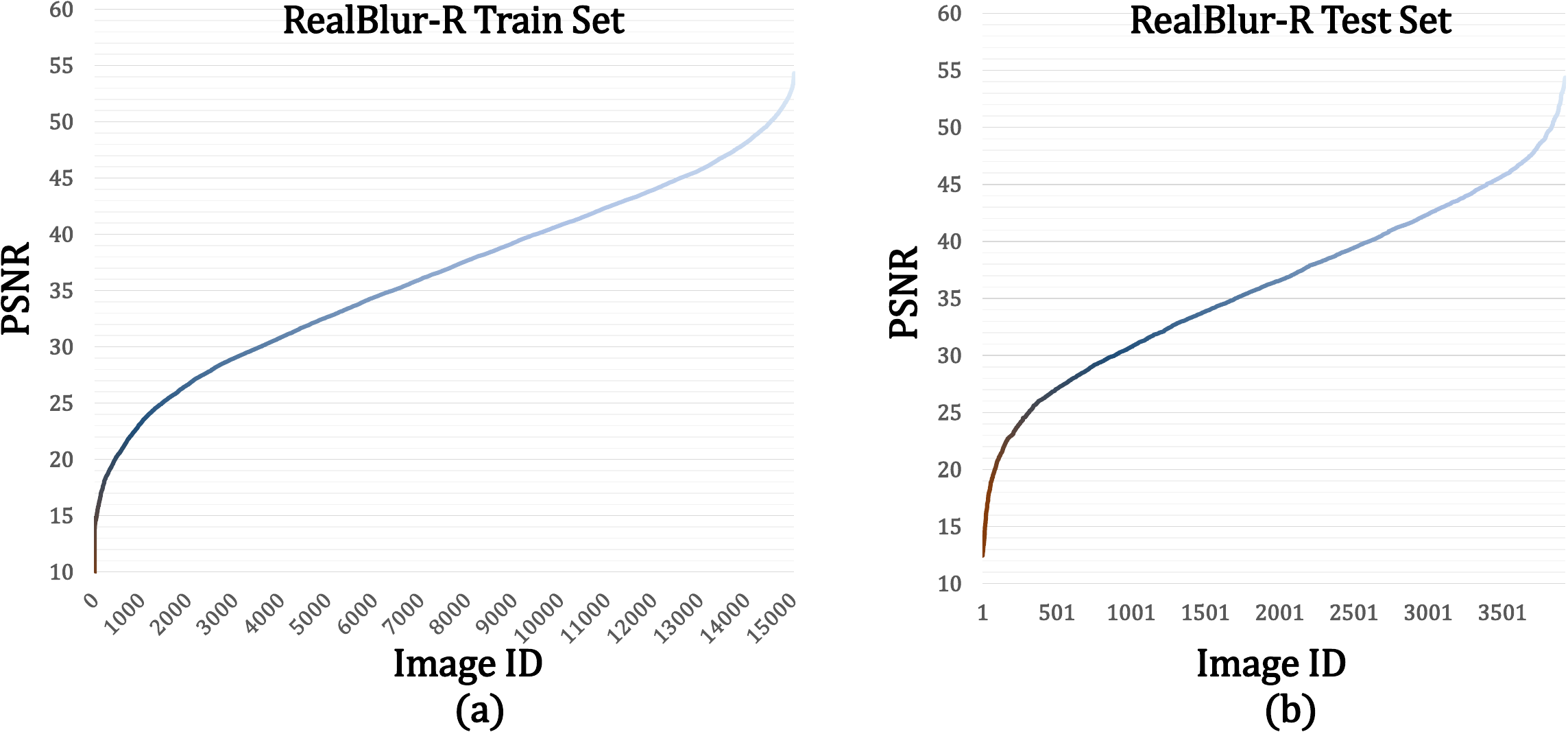}
\end{center}
\vspace{-1.em}
\caption{Distribution of RealBlur-R~\cite{Rim2020real} Dataset. (a) The ranked PSNR curve of the image patches from RealBlur-R train set; (b) The ranked PSNR curve of the image patches from RealBlur-R test set.} 
\label{fig:distribution_realblurr}
\vspace{-1.0em}
\end{figure*}

\paragraph{Dataset Distribution}
Figs.~\ref{fig:distribution_gop_hide}~-~\ref{fig:distribution_realblurr} shows the distribution of the blur patches from different dataset. As indicated in Fig.~\ref{fig:distribution_gop_hide} and Fig.~\ref{fig:distribution_realblurj}, the degraded patches from GoPro, HIDE and RealBlur-J dataset are almost all fall within the range of [15~dB, 45~dB]. Thus, we group the patches to 6 degradation degrees: ($\le$ 20~dB, $\tilde{c}=1$), ($\le$ 25~dB, $\tilde{c}=2$), ($\le$ 30~dB, $\tilde{c}=3$), ($\le$ 35~dB, $\tilde{c}=4$), ($\le$ 40~dB, $\tilde{c}=5$) and ($>$ 40~dB, $\tilde{c}=6$) for the classifier training. Different from other dataset, the degraded patches from RealBlur-R are almost all fall in the range of [15~dB, 55~dB] (shown in Fig.~\ref{fig:distribution_realblurr}). Thus, we group the patches from RealBlur-R to 8  degradation degrees: ($\le$ 20~dB, $\tilde{c}=1$), ($\le$ 25~dB, $\tilde{c}=2$), ($\le$ 30~dB, $\tilde{c}=3$), ($\le$ 35~dB, $\tilde{c}=4$), ($\le$ 40~dB, $\tilde{c}=5$), ($\le$ 45~dB, $\tilde{c}=6$), ($\le$ 50~dB, $\tilde{c}=7$) and ($>$ 50~dB, $\tilde{c}=8$) for the classifier training.

\paragraph{Clustering Criteria}
We split the image patches into different classes according to PSNR, which is a direct and efficient measure of degradation degree. We conduct experiments: \textcolor{red}{\large{\ding{172}}}  In paper, we apply a step ($\gamma$) of 5~dB to cluster the blur patches into 6 degradation degrees from 15~dB to 45~dB. Here we change $\gamma$ within the range of $[3,4,5,6,10]$, and obtain almost the same PSNRs (34.50 dB). Classification accuracies and utilization rates of the sub-decoders (D-rates) are (83.7\%, 87.2\%), (87.0\%, 86.0\%), (89.8\%, 84.3\%), (91.4\%, 87.7\%), and (94.6\%, 84.8\%) respectively. First, although the larger step acquires better classification accuracy, AdaRevD-B ($4$ sub-decoders) has a big tolerance for accuracy corresponding to a small $\gamma$ (\emph{e.g.}, $9$ classes when $\gamma=3$). Second, it is observed that almost all the misclassified patches are classified to the adjacent degradation degree, shown in Table~\ref{tab:confusion-matrix} ($\gamma=5$). Only a few patches would exit at earlier sub-decoder (slightly reduce PSNR of the whole image), while a few exit at later sub-decoder (slightly increase PSNR), which have certain complementary effects on the final PSNR. \textcolor{red}{\large{\ding{173}}} Following ClassSR~\cite{kong2021classsr}, we separate PSNRs into 6 classes with the \textbf{same numbers} of blur patches, the PSNR is also the same (34.50 dB), even with lower classification accuracy (85.1\%). Thus, AdaRevD does not demand a very high classification accuracy, and it is acceptable that a small number of patches are classified to adjacent degradation degree.\\

\paragraph{Evaluation Metric} The computational complexity of MACs (G) and the number of parameters (M) are reported in Table~\ref{tab:macs}. Table~\ref{tab:macs} illustrates that our method can further explore the well-trained NAFNet’s~\cite{Chen2022simple} insufficient decoding capability (33.69 dB) to a higher level (34.10 dB), which is similar to UFPNet~\cite{fang2023UFPNet} (34.06 dB 243 G), but with fewer MACs (168 G).  

\paragraph{Early-exit Signal} In AdaRevD, early-exit signal $E_c$ is determined by $\mathbf{O}_c^j$ and $\tau$. The $\mathbf{O}_c^j$ of RevD-B on GoPro, RealBlur-J and RealBlur-R datasets are summarized in Tables~\ref{tab:RevD-B-GoPro},~\ref{tab:RevD-B-RealBlur-J} and~\ref{tab:RevD-B-RealBlur-R}. Furthermore, the $\mathbf{O}_c^j$ of RevD-L on these three datasets are shown in Tables~\ref{tab:RevD-L-GoPro},~\ref{tab:RevD-L-RealBlur-J} and~\ref{tab:RevD-L-RealBlur-R}. The first \colorbox{color4}{$\mathbf{O}_c^{j-1}$} where its next $\mathbf{O}_c^{j}$ is smaller than $\tau=0.05$ (the patch exit in the ($(j-1)$th sub-decoder) is highlighted in the tables. 

As illustrated in these tables, blur patches with varying degradation degrees exhibit distinct improvements in PSNR within the same sub-decoder. The higher the PSNR, the less restoration the patch undergoes in the identical sub-decoder. As more sub-decoders are progressively stacked, the model's capacity to recover images reaches saturation. Tables~\ref{tab:RevD-B-GoPro} and~\ref{tab:RevD-L-GoPro} demonstrate that the $E_c$ remains consistent between the training set and test set when $\tau=0.05$. Moreover, the performance of the various sud-decoders on the train and test set in the tables indicate that selecting the early-exit signal $E_c$ based on the train set ensures effective recovery of patches from the test set. In essence, opting for $E_c$ from the train set is rational, as the sub-decoder saturation observed in the train set aligns with the saturation observed in the test set.

\section{Viasulizations}
\label{sec:supp_degradation_degree}
\begin{figure*}[t]
\captionsetup[subfigure]{justification=centering, labelformat=empty}
\centering
	\begin{subfigure}{0.135\linewidth}
		\includegraphics[width=0.99\linewidth]{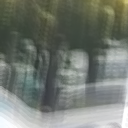}
	\end{subfigure}
  	\begin{subfigure}{0.135\linewidth}
		\includegraphics[width=0.99\linewidth]{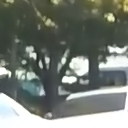}
	\end{subfigure}
 	\begin{subfigure}{0.135\linewidth}
		\includegraphics[width=0.99\linewidth]{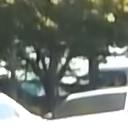}
	\end{subfigure}
	\begin{subfigure}{0.135\linewidth}
		\includegraphics[width=0.99\linewidth]{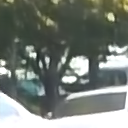}
	\end{subfigure}
 	\begin{subfigure}{0.135\linewidth}
		\includegraphics[width=0.99\linewidth]{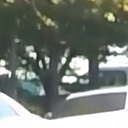}
	\end{subfigure}
	\begin{subfigure}{0.135\linewidth}
		\includegraphics[width=0.99\linewidth]{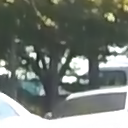}
	\end{subfigure}
     \begin{subfigure}{0.135\linewidth}
		\includegraphics[width=0.99\linewidth]{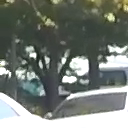}
	\end{subfigure}
	\quad
	\centering
	\begin{subfigure}{0.135\linewidth}
		\includegraphics[width=0.99\linewidth]{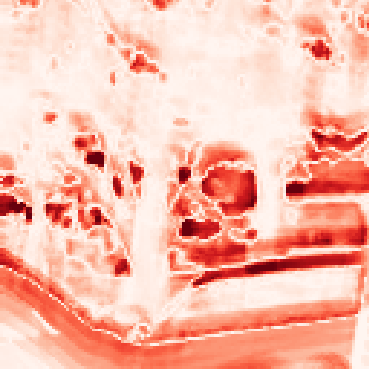}
	\end{subfigure}
  	\begin{subfigure}{0.135\linewidth}
		\includegraphics[width=0.99\linewidth]{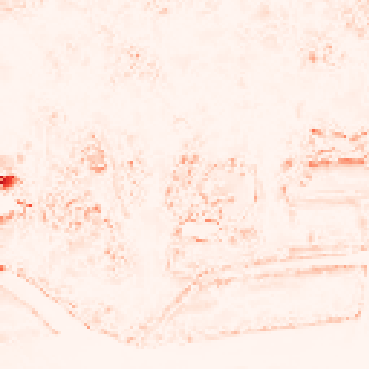}
	\end{subfigure}
 	\begin{subfigure}{0.135\linewidth}
		\includegraphics[width=0.99\linewidth]{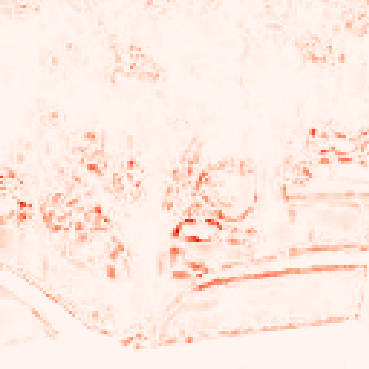}
	\end{subfigure}
	\begin{subfigure}{0.135\linewidth}
		\includegraphics[width=0.99\linewidth]{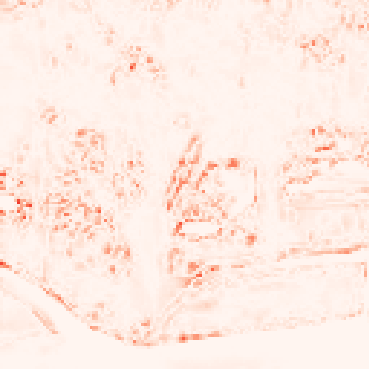}
	\end{subfigure}
    \begin{subfigure}{0.135\linewidth}
		\includegraphics[width=0.99\linewidth]{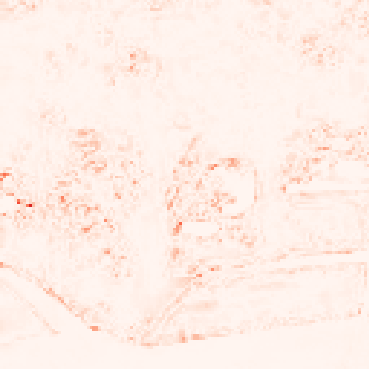}
	\end{subfigure}
	\begin{subfigure}{0.135\linewidth}
		\includegraphics[width=0.99\linewidth]{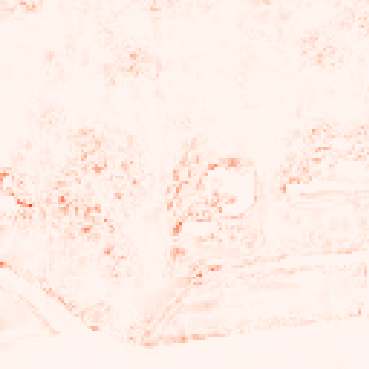}
	\end{subfigure}
     \begin{subfigure}{0.135\linewidth}
		\includegraphics[width=0.99\linewidth]{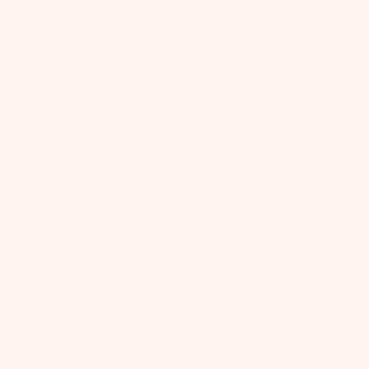}
	\end{subfigure}
    \quad
    \centering
	\begin{subfigure}{0.135\linewidth}
		\includegraphics[width=0.99\linewidth]{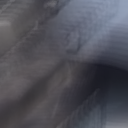}
	\end{subfigure}
  	\begin{subfigure}{0.135\linewidth}
		\includegraphics[width=0.99\linewidth]{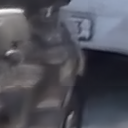}
	\end{subfigure}
 	\begin{subfigure}{0.135\linewidth}
		\includegraphics[width=0.99\linewidth]{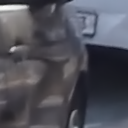}
	\end{subfigure}
	\begin{subfigure}{0.135\linewidth}
		\includegraphics[width=0.99\linewidth]{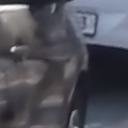}
	\end{subfigure}
 	\begin{subfigure}{0.135\linewidth}
		\includegraphics[width=0.99\linewidth]{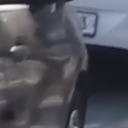}
	\end{subfigure}
	\begin{subfigure}{0.135\linewidth}
		\includegraphics[width=0.99\linewidth]{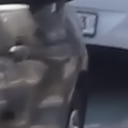}
	\end{subfigure}
     \begin{subfigure}{0.135\linewidth}
		\includegraphics[width=0.99\linewidth]{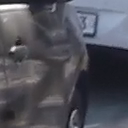}
	\end{subfigure}
	\quad
	\centering
	\begin{subfigure}{0.135\linewidth}
		\includegraphics[width=0.99\linewidth]{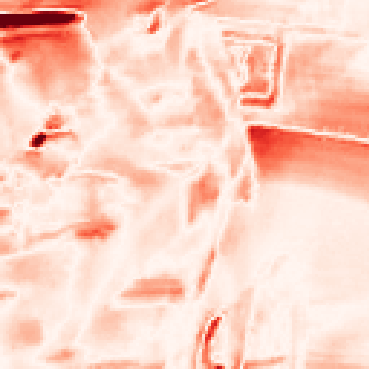}
	\end{subfigure}
  	\begin{subfigure}{0.135\linewidth}
		\includegraphics[width=0.99\linewidth]{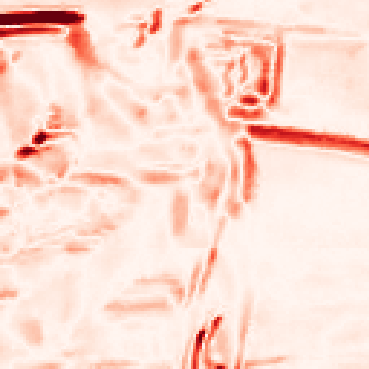}
	\end{subfigure}
 	\begin{subfigure}{0.135\linewidth}
		\includegraphics[width=0.99\linewidth]{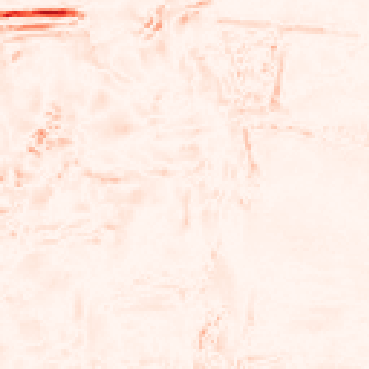}
	\end{subfigure}
	\begin{subfigure}{0.135\linewidth}
		\includegraphics[width=0.99\linewidth]{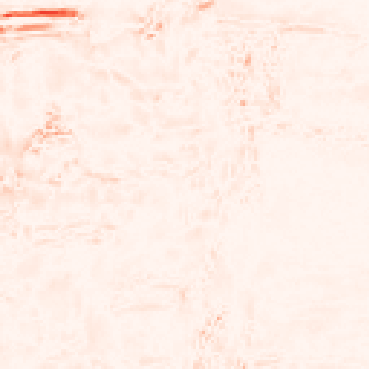}
	\end{subfigure}
    \begin{subfigure}{0.135\linewidth}
		\includegraphics[width=0.99\linewidth]{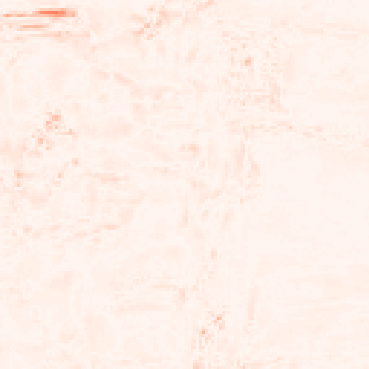}
	\end{subfigure}
	\begin{subfigure}{0.135\linewidth}
		\includegraphics[width=0.99\linewidth]{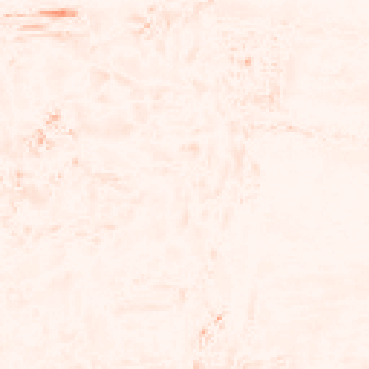}
	\end{subfigure}
     \begin{subfigure}{0.135\linewidth}
		\includegraphics[width=0.99\linewidth]{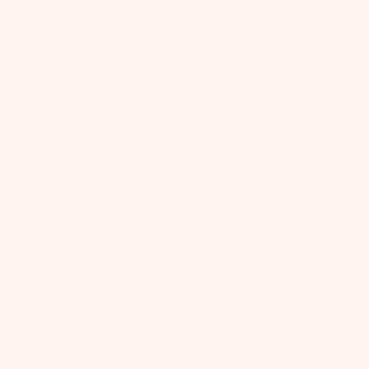}
	\end{subfigure}
    \quad
    \centering
	\begin{subfigure}{0.135\linewidth}
		\includegraphics[width=0.99\linewidth]{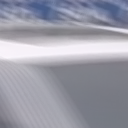}
	\end{subfigure}
  	\begin{subfigure}{0.135\linewidth}
		\includegraphics[width=0.99\linewidth]{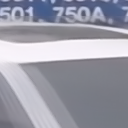}
	\end{subfigure}
 	\begin{subfigure}{0.135\linewidth}
		\includegraphics[width=0.99\linewidth]{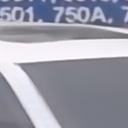}
	\end{subfigure}
	\begin{subfigure}{0.135\linewidth}
		\includegraphics[width=0.99\linewidth]{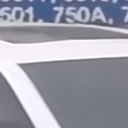}
	\end{subfigure}
 	\begin{subfigure}{0.135\linewidth}
		\includegraphics[width=0.99\linewidth]{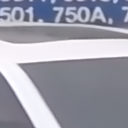}
	\end{subfigure}
	\begin{subfigure}{0.135\linewidth}
		\includegraphics[width=0.99\linewidth]{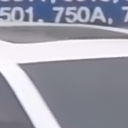}
	\end{subfigure}
     \begin{subfigure}{0.135\linewidth}
		\includegraphics[width=0.99\linewidth]{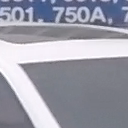}
	\end{subfigure}
	\quad
	\centering
	\begin{subfigure}{0.135\linewidth}
		\includegraphics[width=0.99\linewidth]{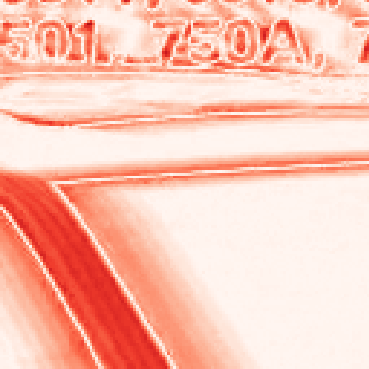}
	\end{subfigure}
  	\begin{subfigure}{0.135\linewidth}
		\includegraphics[width=0.99\linewidth]{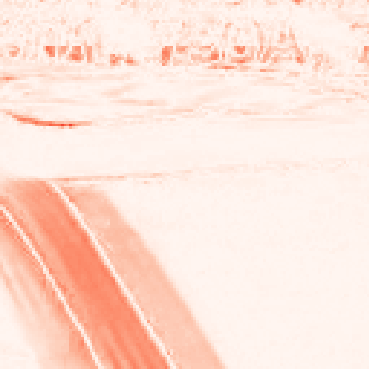}
	\end{subfigure}
 	\begin{subfigure}{0.135\linewidth}
		\includegraphics[width=0.99\linewidth]{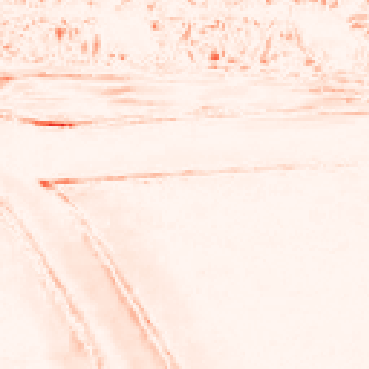}
	\end{subfigure}
	\begin{subfigure}{0.135\linewidth}
		\includegraphics[width=0.99\linewidth]{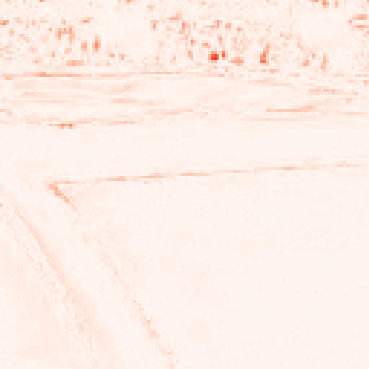}
	\end{subfigure}
    \begin{subfigure}{0.135\linewidth}
		\includegraphics[width=0.99\linewidth]{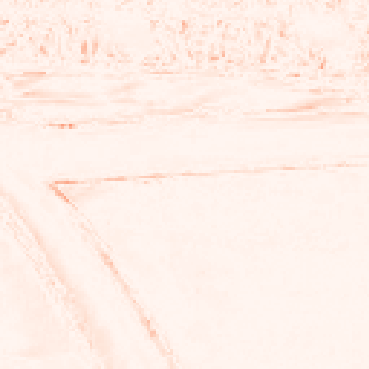}
	\end{subfigure}
	\begin{subfigure}{0.135\linewidth}
		\includegraphics[width=0.99\linewidth]{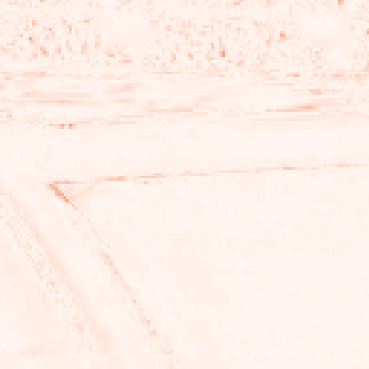}
	\end{subfigure}
     \begin{subfigure}{0.135\linewidth}
		\includegraphics[width=0.99\linewidth]{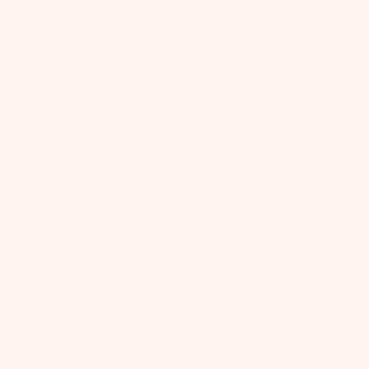}
	\end{subfigure}
     \quad
    \centering
	\begin{subfigure}{0.135\linewidth}
		\includegraphics[width=0.99\linewidth]{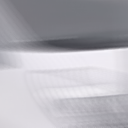}
	\end{subfigure}
  	\begin{subfigure}{0.135\linewidth}
		\includegraphics[width=0.99\linewidth]{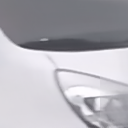}
	\end{subfigure}
 	\begin{subfigure}{0.135\linewidth}
		\includegraphics[width=0.99\linewidth]{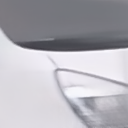}
	\end{subfigure}
	\begin{subfigure}{0.135\linewidth}
		\includegraphics[width=0.99\linewidth]{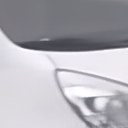}
	\end{subfigure}
 	\begin{subfigure}{0.135\linewidth}
		\includegraphics[width=0.99\linewidth]{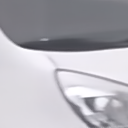}
	\end{subfigure}
	\begin{subfigure}{0.135\linewidth}
		\includegraphics[width=0.99\linewidth]{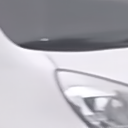}
	\end{subfigure}
     \begin{subfigure}{0.135\linewidth}
		\includegraphics[width=0.99\linewidth]{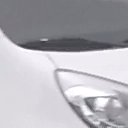}
	\end{subfigure}
	\quad
	\centering
	\begin{subfigure}{0.135\linewidth}
		\includegraphics[width=0.99\linewidth]{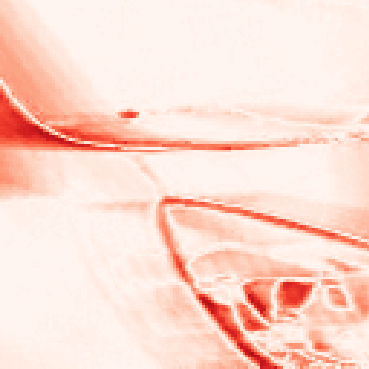}
        \caption{Blur}
	\end{subfigure}
  	\begin{subfigure}{0.135\linewidth}
		\includegraphics[width=0.99\linewidth]{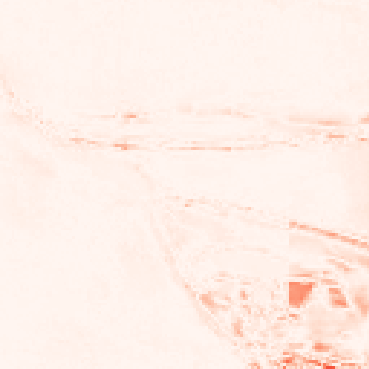}
        \caption{DeepRFT+\cite{XintianMao2023DeepRFT}} 
	\end{subfigure}
 	\begin{subfigure}{0.135\linewidth}
		\includegraphics[width=0.99\linewidth]{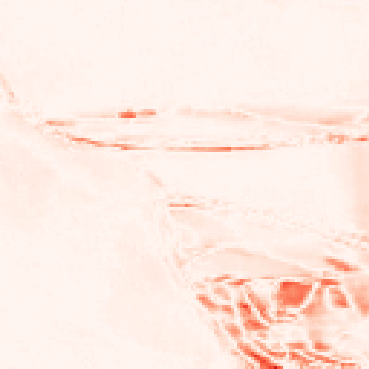}
  		\caption{NAFNet64~\cite{Chen2022simple}} 
	\end{subfigure}
	\begin{subfigure}{0.135\linewidth}
		\includegraphics[width=0.99\linewidth]{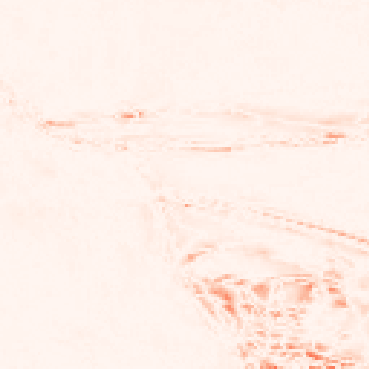}
 		\caption{UFPNet\cite{fang2023UFPNet}} 
	\end{subfigure}
    \begin{subfigure}{0.135\linewidth}
		\includegraphics[width=0.99\linewidth]{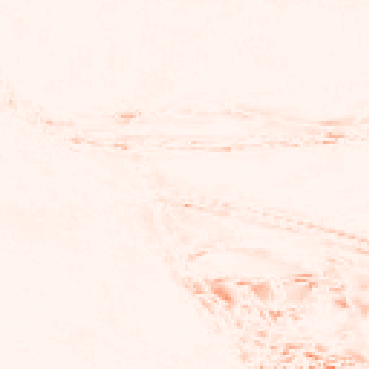}
     	\caption{AdaRevD-B} 
	\end{subfigure}
	\begin{subfigure}{0.135\linewidth}
		\includegraphics[width=0.99\linewidth]{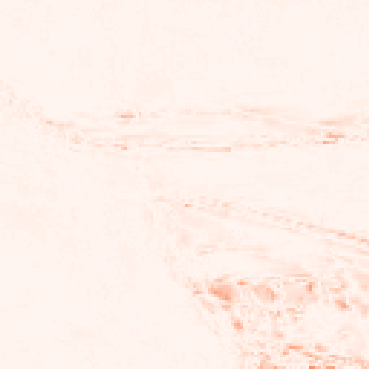}
     	\caption{AdaRevD-L} 
	\end{subfigure}
    \begin{subfigure}{0.135\linewidth}
		\includegraphics[width=0.99\linewidth]{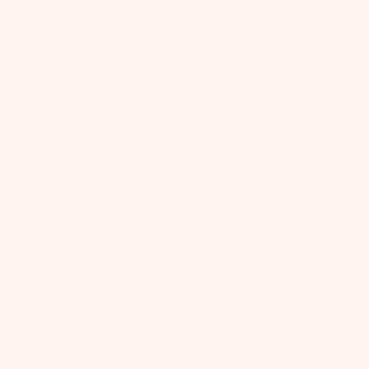}
     	\caption{Sharp}
	\end{subfigure}
\caption{Examples on the GoPro test dataset. The odd rows show blur image, predicted images of different methods, and ground-truth sharp image. The even rows show the residual of the blur image / predicted sharp images and the ground-truth sharp image.}
\label{fig:supp_GoPro}
\vspace{-0.5em}
\end{figure*}

\begin{figure*}[t]
\captionsetup[subfigure]{justification=centering, labelformat=empty}
\centering
	\begin{subfigure}{0.135\linewidth}
		\includegraphics[width=0.99\linewidth]{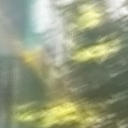}
	\end{subfigure}
  	\begin{subfigure}{0.135\linewidth}
		\includegraphics[width=0.99\linewidth]{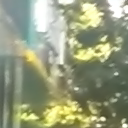}
	\end{subfigure}
 	\begin{subfigure}{0.135\linewidth}
		\includegraphics[width=0.99\linewidth]{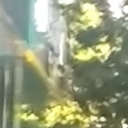}
	\end{subfigure}
	\begin{subfigure}{0.135\linewidth}
		\includegraphics[width=0.99\linewidth]{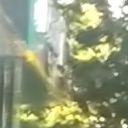}
	\end{subfigure}
 	\begin{subfigure}{0.135\linewidth}
		\includegraphics[width=0.99\linewidth]{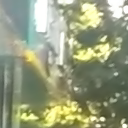}
	\end{subfigure}
	\begin{subfigure}{0.135\linewidth}
		\includegraphics[width=0.99\linewidth]{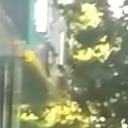}
	\end{subfigure}
     \begin{subfigure}{0.135\linewidth}
		\includegraphics[width=0.99\linewidth]{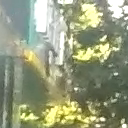}
	\end{subfigure}
	\quad
	\centering
	\begin{subfigure}{0.135\linewidth}
		\includegraphics[width=0.99\linewidth]{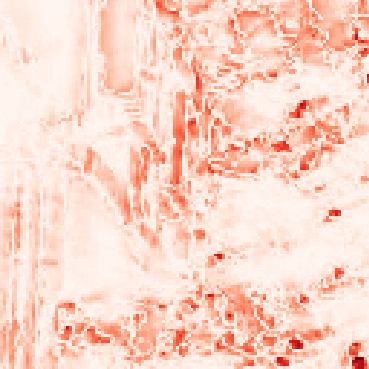}
	\end{subfigure}
  	\begin{subfigure}{0.135\linewidth}
		\includegraphics[width=0.99\linewidth]{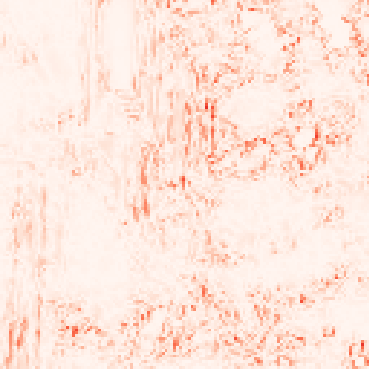}
	\end{subfigure}
 	\begin{subfigure}{0.135\linewidth}
		\includegraphics[width=0.99\linewidth]{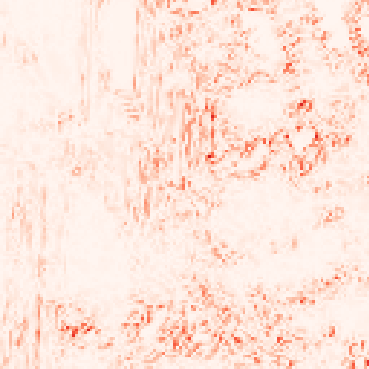}
	\end{subfigure}
	\begin{subfigure}{0.135\linewidth}
		\includegraphics[width=0.99\linewidth]{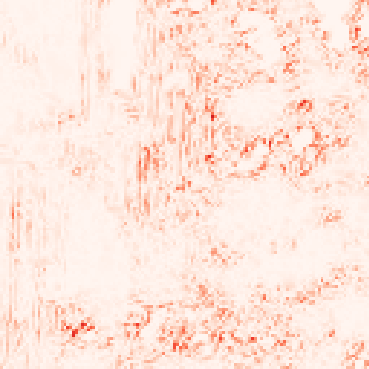}
	\end{subfigure}
    \begin{subfigure}{0.135\linewidth}
		\includegraphics[width=0.99\linewidth]{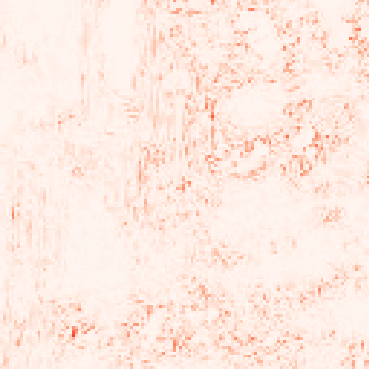}
	\end{subfigure}
	\begin{subfigure}{0.135\linewidth}
		\includegraphics[width=0.99\linewidth]{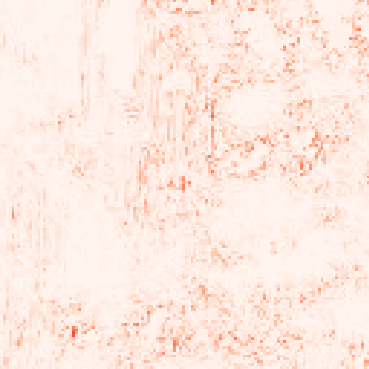}
	\end{subfigure}
     \begin{subfigure}{0.135\linewidth}
		\includegraphics[width=0.99\linewidth]{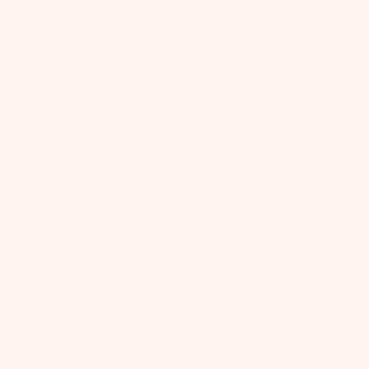}
	\end{subfigure}
    \quad
    \centering
	\begin{subfigure}{0.135\linewidth}
		\includegraphics[width=0.99\linewidth]{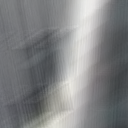}
	\end{subfigure}
  	\begin{subfigure}{0.135\linewidth}
		\includegraphics[width=0.99\linewidth]{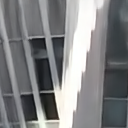}
	\end{subfigure}
 	\begin{subfigure}{0.135\linewidth}
		\includegraphics[width=0.99\linewidth]{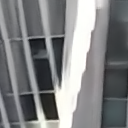}
	\end{subfigure}
	\begin{subfigure}{0.135\linewidth}
		\includegraphics[width=0.99\linewidth]{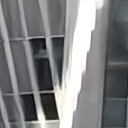}
	\end{subfigure}
 	\begin{subfigure}{0.135\linewidth}
		\includegraphics[width=0.99\linewidth]{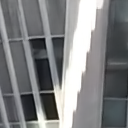}
	\end{subfigure}
	\begin{subfigure}{0.135\linewidth}
		\includegraphics[width=0.99\linewidth]{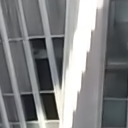}
	\end{subfigure}
     \begin{subfigure}{0.135\linewidth}
		\includegraphics[width=0.99\linewidth]{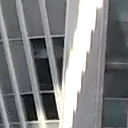}
	\end{subfigure}
	\quad
	\centering
	\begin{subfigure}{0.135\linewidth}
		\includegraphics[width=0.99\linewidth]{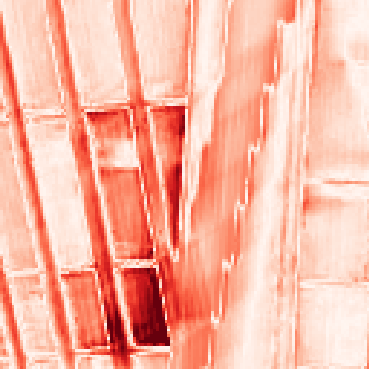}
	\end{subfigure}
  	\begin{subfigure}{0.135\linewidth}
		\includegraphics[width=0.99\linewidth]{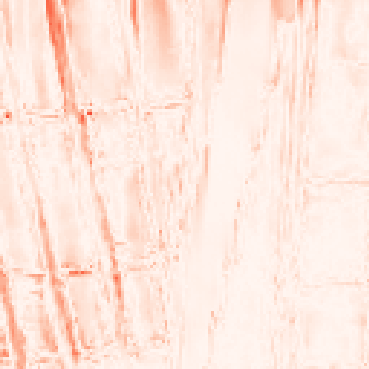}
	\end{subfigure}
 	\begin{subfigure}{0.135\linewidth}
		\includegraphics[width=0.99\linewidth]{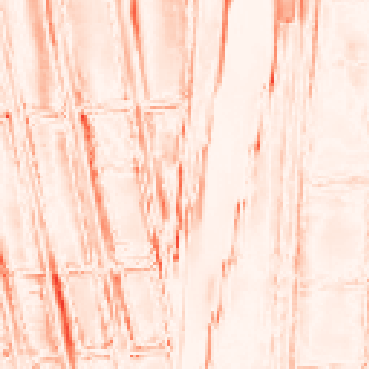}
	\end{subfigure}
	\begin{subfigure}{0.135\linewidth}
		\includegraphics[width=0.99\linewidth]{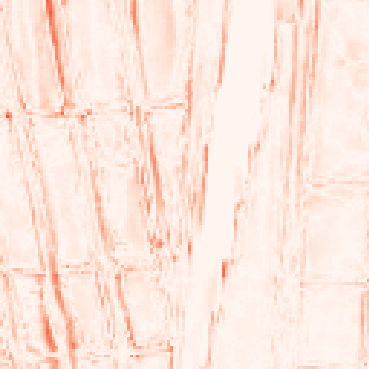}
	\end{subfigure}
    \begin{subfigure}{0.135\linewidth}
		\includegraphics[width=0.99\linewidth]{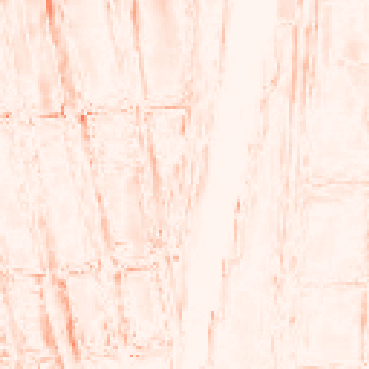}
	\end{subfigure}
	\begin{subfigure}{0.135\linewidth}
		\includegraphics[width=0.99\linewidth]{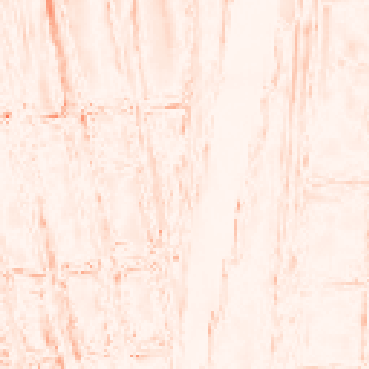}
	\end{subfigure}
     \begin{subfigure}{0.135\linewidth}
		\includegraphics[width=0.99\linewidth]{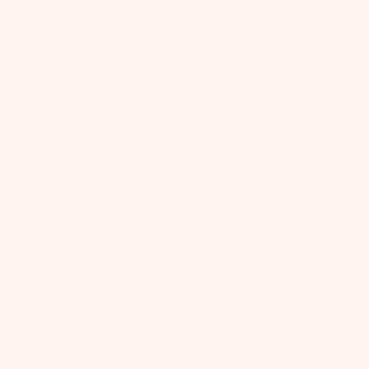}
	\end{subfigure}
    \quad
    \centering
	\begin{subfigure}{0.135\linewidth}
		\includegraphics[width=0.99\linewidth]{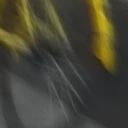}
	\end{subfigure}
  	\begin{subfigure}{0.135\linewidth}
		\includegraphics[width=0.99\linewidth]{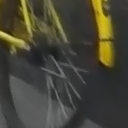}
	\end{subfigure}
 	\begin{subfigure}{0.135\linewidth}
		\includegraphics[width=0.99\linewidth]{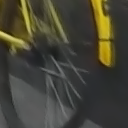}
	\end{subfigure}
	\begin{subfigure}{0.135\linewidth}
		\includegraphics[width=0.99\linewidth]{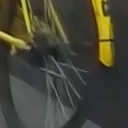}
	\end{subfigure}
 	\begin{subfigure}{0.135\linewidth}
		\includegraphics[width=0.99\linewidth]{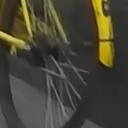}
	\end{subfigure}
	\begin{subfigure}{0.135\linewidth}
		\includegraphics[width=0.99\linewidth]{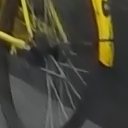}
	\end{subfigure}
    \begin{subfigure}{0.135\linewidth}
		\includegraphics[width=0.99\linewidth]{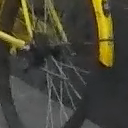}
	\end{subfigure}
	\quad
	\centering
	\begin{subfigure}{0.135\linewidth}
		\includegraphics[width=0.99\linewidth]{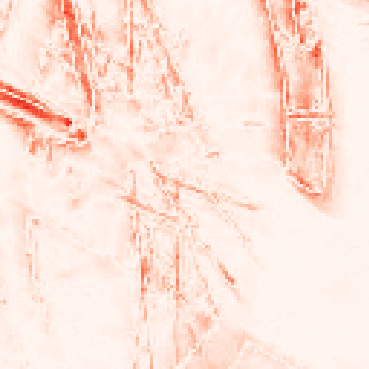}
	\end{subfigure}
  	\begin{subfigure}{0.135\linewidth}
		\includegraphics[width=0.99\linewidth]{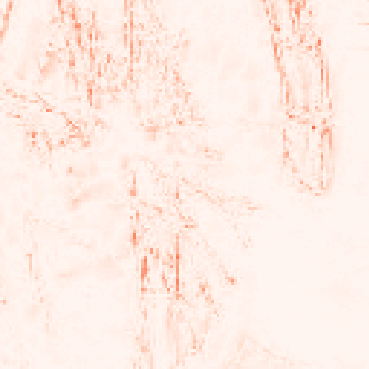}
	\end{subfigure}
 	\begin{subfigure}{0.135\linewidth}
		\includegraphics[width=0.99\linewidth]{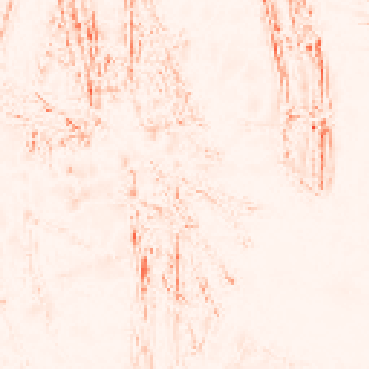}
	\end{subfigure}
	\begin{subfigure}{0.135\linewidth}
		\includegraphics[width=0.99\linewidth]{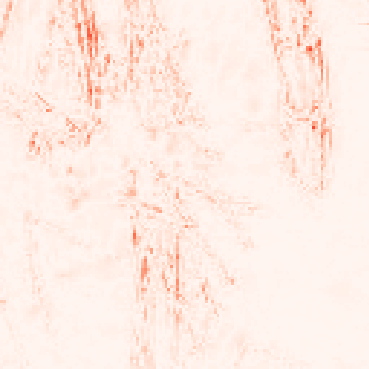}
	\end{subfigure}
    \begin{subfigure}{0.135\linewidth}
		\includegraphics[width=0.99\linewidth]{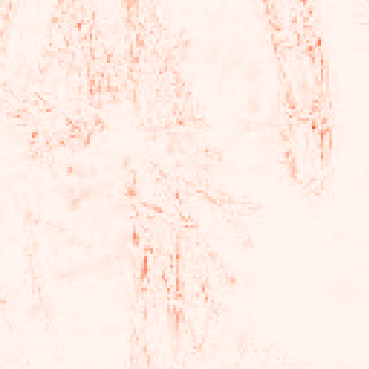}
	\end{subfigure}
	\begin{subfigure}{0.135\linewidth}
		\includegraphics[width=0.99\linewidth]{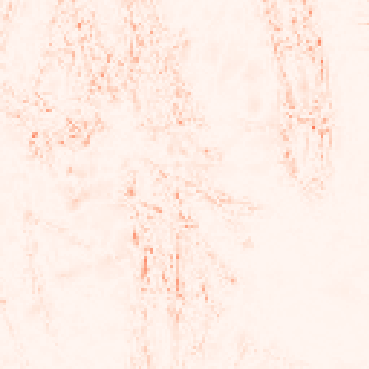}
	\end{subfigure}
    \begin{subfigure}{0.135\linewidth}
		\includegraphics[width=0.99\linewidth]{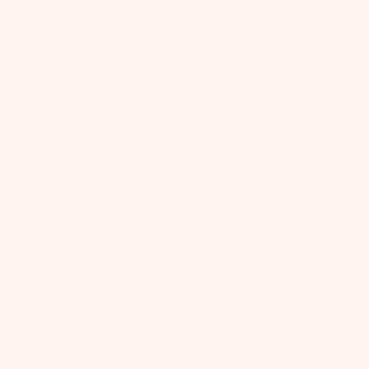}
	\end{subfigure}
    \quad
    \centering
	\begin{subfigure}{0.135\linewidth}
		\includegraphics[width=0.99\linewidth]{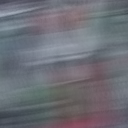}
	\end{subfigure}
  	\begin{subfigure}{0.135\linewidth}
		\includegraphics[width=0.99\linewidth]{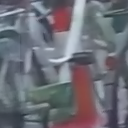}
	\end{subfigure}
 	\begin{subfigure}{0.135\linewidth}
		\includegraphics[width=0.99\linewidth]{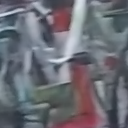}
	\end{subfigure}
	\begin{subfigure}{0.135\linewidth}
		\includegraphics[width=0.99\linewidth]{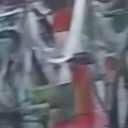}
	\end{subfigure}
 	\begin{subfigure}{0.135\linewidth}
		\includegraphics[width=0.99\linewidth]{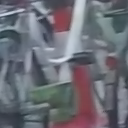}
	\end{subfigure}
	\begin{subfigure}{0.135\linewidth}
		\includegraphics[width=0.99\linewidth]{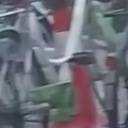}
	\end{subfigure}
     \begin{subfigure}{0.135\linewidth}
		\includegraphics[width=0.99\linewidth]{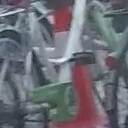}
	\end{subfigure}
	\quad
	\centering
	\begin{subfigure}{0.135\linewidth}
		\includegraphics[width=0.99\linewidth]{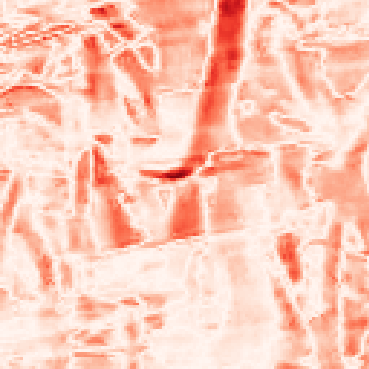}
        \caption{Blur}
	\end{subfigure}
  	\begin{subfigure}{0.135\linewidth}
		\includegraphics[width=0.99\linewidth]{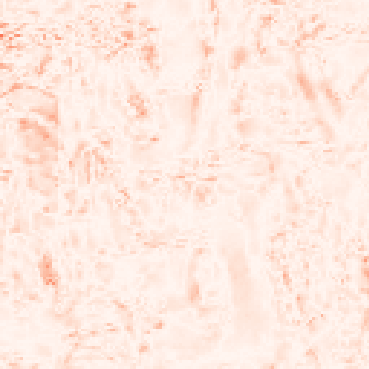}
        \caption{DeepRFT+\cite{XintianMao2023DeepRFT}} 
	\end{subfigure}
 	\begin{subfigure}{0.135\linewidth}
		\includegraphics[width=0.99\linewidth]{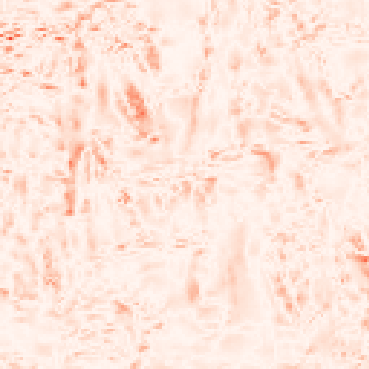}
  		\caption{NAFNet64~\cite{Chen2022simple}} 
	\end{subfigure}
	\begin{subfigure}{0.135\linewidth}
		\includegraphics[width=0.99\linewidth]{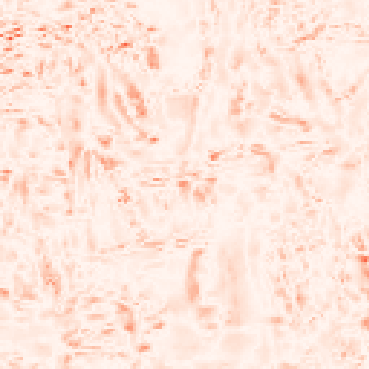}
 		\caption{UFPNet\cite{fang2023UFPNet}} 
	\end{subfigure}
    \begin{subfigure}{0.135\linewidth}
		\includegraphics[width=0.99\linewidth]{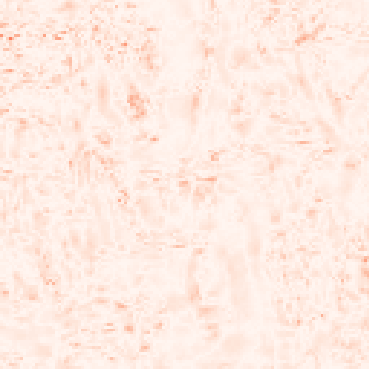}
     	\caption{AdaRevD-B} 
	\end{subfigure}
	\begin{subfigure}{0.135\linewidth}
		\includegraphics[width=0.99\linewidth]{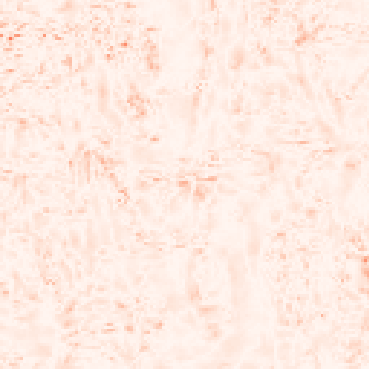}
     	\caption{AdaRevD-L} 
	\end{subfigure}
     \begin{subfigure}{0.135\linewidth}
		\includegraphics[width=0.99\linewidth]{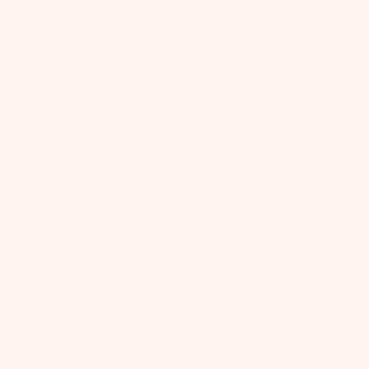}
     	\caption{Sharp}
	\end{subfigure}
\caption{Examples on the HIDE test dataset.}
\label{fig:supp_HIDE}
\vspace{-0.5em}
\end{figure*}

\begin{figure*}[htbp]
\captionsetup[subfigure]{justification=centering, labelformat=empty}
\centering
	\begin{subfigure}{0.135\linewidth}
		\includegraphics[width=0.99\linewidth]{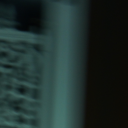}
	\end{subfigure}
	\begin{subfigure}{0.135\linewidth}
		\includegraphics[width=0.99\linewidth]{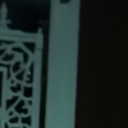}
	\end{subfigure}
	\begin{subfigure}{0.135\linewidth}
		\includegraphics[width=0.99\linewidth]{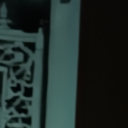}
	\end{subfigure}
	\begin{subfigure}{0.135\linewidth}
		\includegraphics[width=0.99\linewidth]{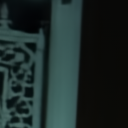}
	\end{subfigure}
	\begin{subfigure}{0.135\linewidth}
		\includegraphics[width=0.99\linewidth]{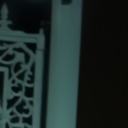}
	\end{subfigure}
    \begin{subfigure}{0.135\linewidth}
		\includegraphics[width=0.99\linewidth]{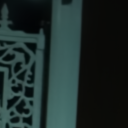}
	\end{subfigure}
     \begin{subfigure}{0.135\linewidth}
		\includegraphics[width=0.99\linewidth]{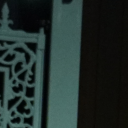}
	\end{subfigure}
	\quad
	\centering
		\begin{subfigure}{0.135\linewidth}
		\includegraphics[width=0.99\linewidth]{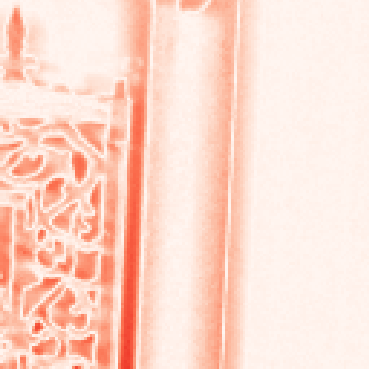}
	\end{subfigure}
	\begin{subfigure}{0.135\linewidth}
		\includegraphics[width=0.99\linewidth]{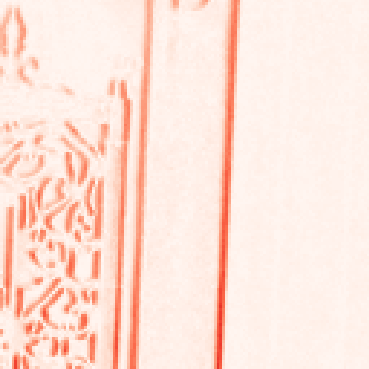}
	\end{subfigure}
	\begin{subfigure}{0.135\linewidth}
		\includegraphics[width=0.99\linewidth]{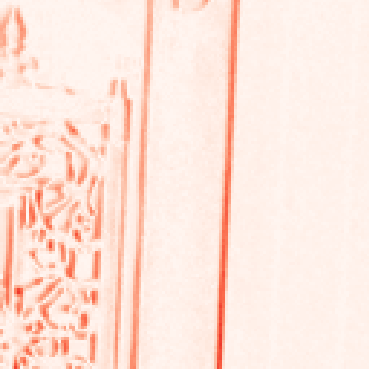}
	\end{subfigure}
	\begin{subfigure}{0.135\linewidth}
		\includegraphics[width=0.99\linewidth]{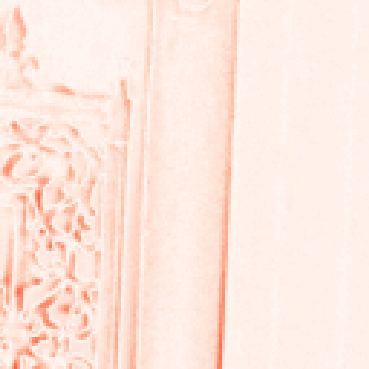}
	\end{subfigure}
	\begin{subfigure}{0.135\linewidth}
		\includegraphics[width=0.99\linewidth]{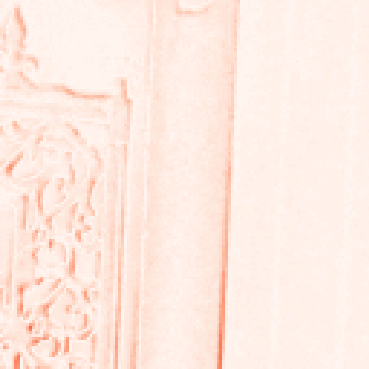}
	\end{subfigure}
    \begin{subfigure}{0.135\linewidth}
		\includegraphics[width=0.99\linewidth]{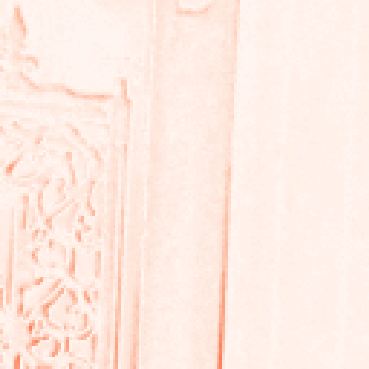}
	\end{subfigure}
     \begin{subfigure}{0.135\linewidth}

		\includegraphics[width=0.99\linewidth]{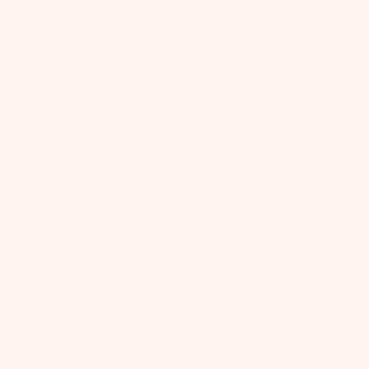}
	\end{subfigure}
 \quad
 	\begin{subfigure}{0.135\linewidth}
		\includegraphics[width=0.99\linewidth]{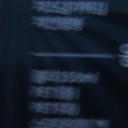}
	\end{subfigure}
	\begin{subfigure}{0.135\linewidth}
		\includegraphics[width=0.99\linewidth]{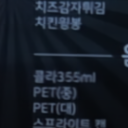}
	\end{subfigure}
	\begin{subfigure}{0.135\linewidth}
		\includegraphics[width=0.99\linewidth]{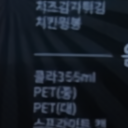}
	\end{subfigure}
	\begin{subfigure}{0.135\linewidth}
		\includegraphics[width=0.99\linewidth]{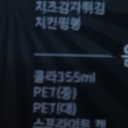}
	\end{subfigure}
	\begin{subfigure}{0.135\linewidth}
		\includegraphics[width=0.99\linewidth]{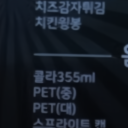}
	\end{subfigure}
    \begin{subfigure}{0.135\linewidth}
		\includegraphics[width=0.99\linewidth]{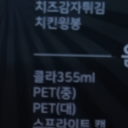}
	\end{subfigure}
     \begin{subfigure}{0.135\linewidth}
		\includegraphics[width=0.99\linewidth]{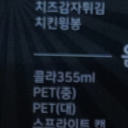}
	\end{subfigure}
	\quad
	\centering
		\begin{subfigure}{0.135\linewidth}
		\includegraphics[width=0.99\linewidth]{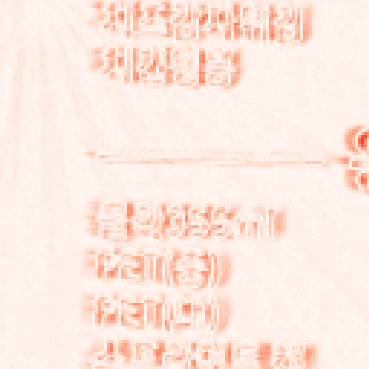}
	\end{subfigure}
	\begin{subfigure}{0.135\linewidth}
		\includegraphics[width=0.99\linewidth]{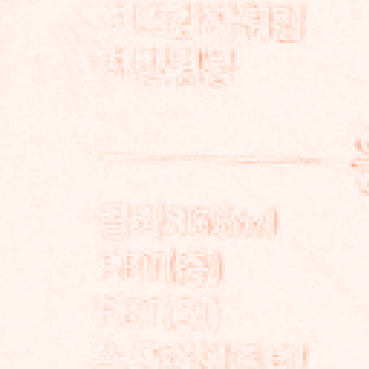}
	\end{subfigure}
	\begin{subfigure}{0.135\linewidth}
		\includegraphics[width=0.99\linewidth]{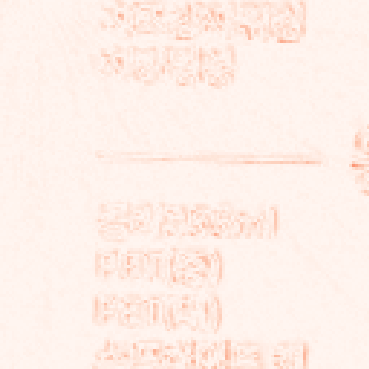}
	\end{subfigure}
	\begin{subfigure}{0.135\linewidth}
		\includegraphics[width=0.99\linewidth]{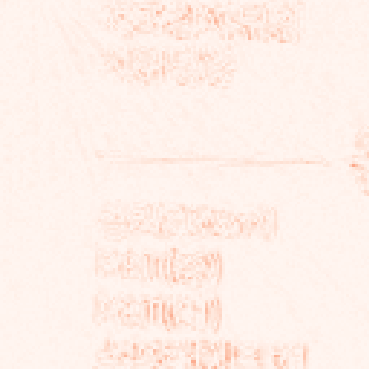}
	\end{subfigure}
	\begin{subfigure}{0.135\linewidth}
		\includegraphics[width=0.99\linewidth]{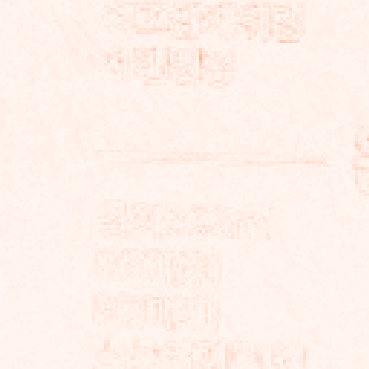}
	\end{subfigure}
    \begin{subfigure}{0.135\linewidth}
		\includegraphics[width=0.99\linewidth]{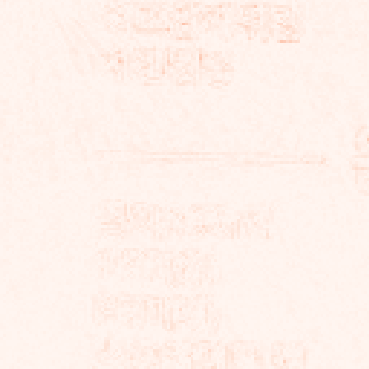}
	\end{subfigure}
     \begin{subfigure}{0.135\linewidth}

		\includegraphics[width=0.99\linewidth]{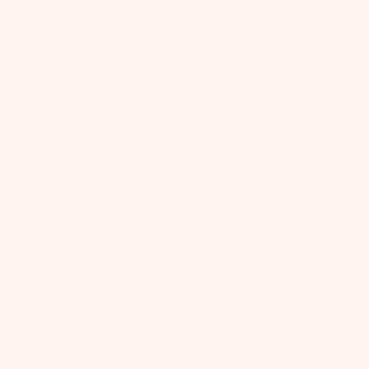}
	\end{subfigure}
 \quad
 \centering
	\begin{subfigure}{0.135\linewidth}
		\includegraphics[width=0.99\linewidth]{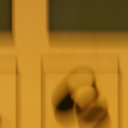}
	\end{subfigure}
	\begin{subfigure}{0.135\linewidth}
		\includegraphics[width=0.99\linewidth]{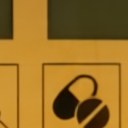}
	\end{subfigure}
	\begin{subfigure}{0.135\linewidth}
		\includegraphics[width=0.99\linewidth]{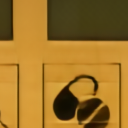}
	\end{subfigure}
	\begin{subfigure}{0.135\linewidth}
		\includegraphics[width=0.99\linewidth]{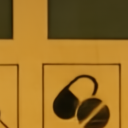}
	\end{subfigure}
	\begin{subfigure}{0.135\linewidth}
		\includegraphics[width=0.99\linewidth]{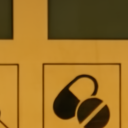}
	\end{subfigure}
    \begin{subfigure}{0.135\linewidth}
		\includegraphics[width=0.99\linewidth]{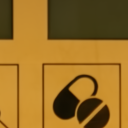}
	\end{subfigure}
     \begin{subfigure}{0.135\linewidth}
		\includegraphics[width=0.99\linewidth]{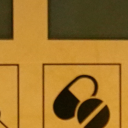}
	\end{subfigure}
	\quad
	\centering
		\begin{subfigure}{0.135\linewidth}
		\includegraphics[width=0.99\linewidth]{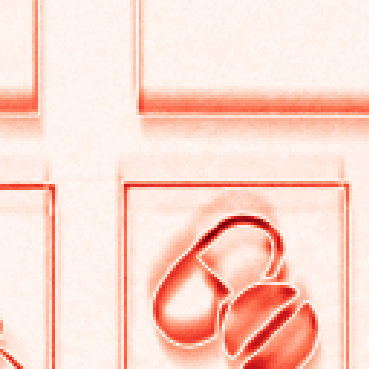}
	\end{subfigure}
	\begin{subfigure}{0.135\linewidth}
		\includegraphics[width=0.99\linewidth]{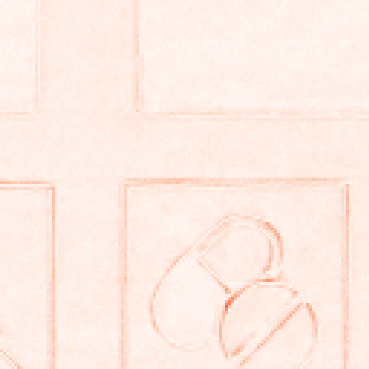}
	\end{subfigure}
	\begin{subfigure}{0.135\linewidth}
		\includegraphics[width=0.99\linewidth]{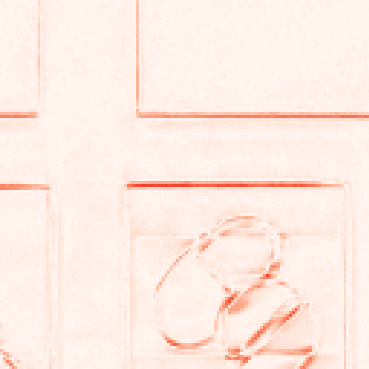}
	\end{subfigure}
	\begin{subfigure}{0.135\linewidth}
		\includegraphics[width=0.99\linewidth]{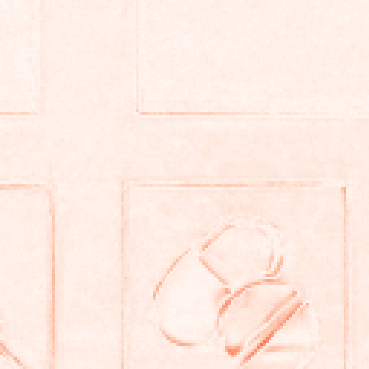}
	\end{subfigure}
	\begin{subfigure}{0.135\linewidth}
		\includegraphics[width=0.99\linewidth]{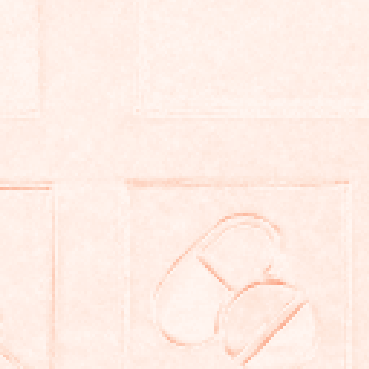}
	\end{subfigure}
    \begin{subfigure}{0.135\linewidth}
		\includegraphics[width=0.99\linewidth]{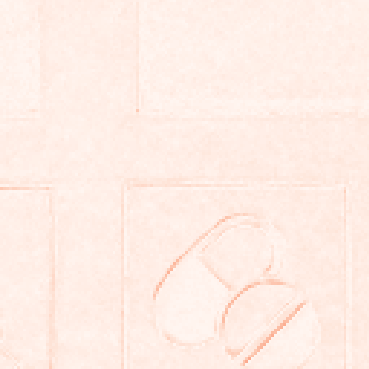}
	\end{subfigure}
     \begin{subfigure}{0.135\linewidth}

		\includegraphics[width=0.99\linewidth]{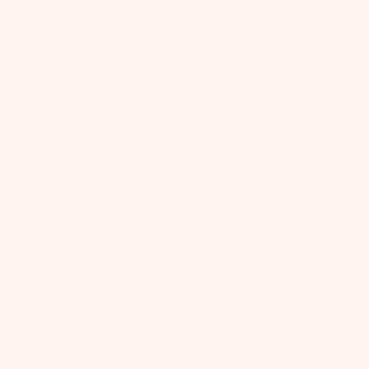}
	\end{subfigure}
  \quad
 \centering
	\begin{subfigure}{0.135\linewidth}
		\includegraphics[width=0.99\linewidth]{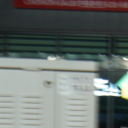}
	\end{subfigure}
	\begin{subfigure}{0.135\linewidth}
		\includegraphics[width=0.99\linewidth]{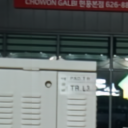}
	\end{subfigure}
	\begin{subfigure}{0.135\linewidth}
		\includegraphics[width=0.99\linewidth]{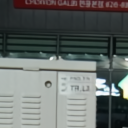}
	\end{subfigure}
	\begin{subfigure}{0.135\linewidth}
		\includegraphics[width=0.99\linewidth]{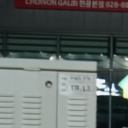}
	\end{subfigure}
	\begin{subfigure}{0.135\linewidth}
		\includegraphics[width=0.99\linewidth]{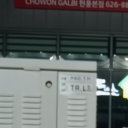}
	\end{subfigure}
    \begin{subfigure}{0.135\linewidth}
		\includegraphics[width=0.99\linewidth]{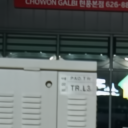}
	\end{subfigure}
     \begin{subfigure}{0.135\linewidth}
		\includegraphics[width=0.99\linewidth]{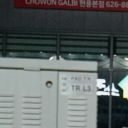}
	\end{subfigure}
	\quad
	\centering
		\begin{subfigure}{0.135\linewidth}
		\includegraphics[width=0.99\linewidth]{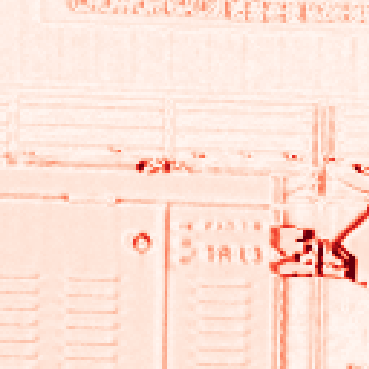}
        \caption{Blur}
	\end{subfigure}
	\begin{subfigure}{0.135\linewidth}
		\includegraphics[width=0.99\linewidth]{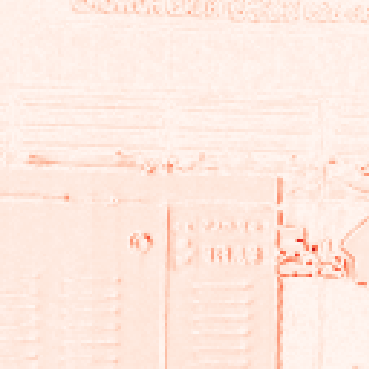}
        \caption{DeepRFT+~\cite{XintianMao2023DeepRFT}}
	\end{subfigure}
	\begin{subfigure}{0.135\linewidth}
		\includegraphics[width=0.99\linewidth]{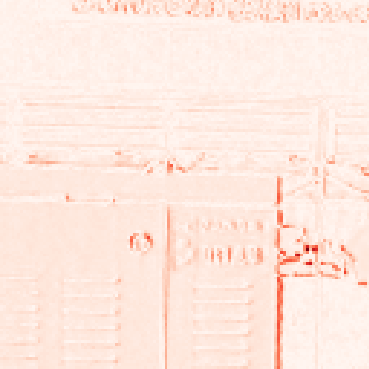}
        \caption{Stripformer~\cite{Tsai2022Stripformer}}
	\end{subfigure}
	\begin{subfigure}{0.135\linewidth}
		\includegraphics[width=0.99\linewidth]{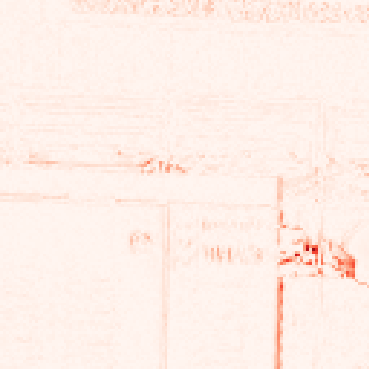}
          \caption{UFPNet~\cite{fang2023UFPNet}}
	\end{subfigure}
	\begin{subfigure}{0.135\linewidth}
		\includegraphics[width=0.99\linewidth]{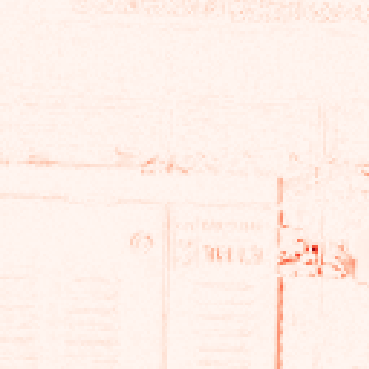}
        \caption{AdaRevD-B}
	\end{subfigure}
    \begin{subfigure}{0.135\linewidth}
		\includegraphics[width=0.99\linewidth]{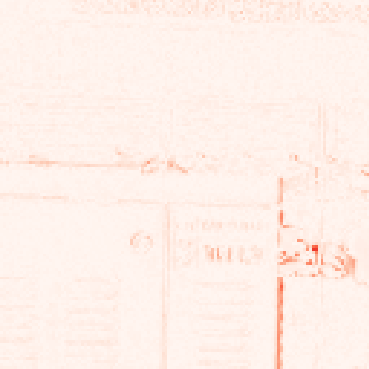}
        \caption{AdaRevD-L}
	\end{subfigure}
     \begin{subfigure}{0.135\linewidth}

		\includegraphics[width=0.99\linewidth]{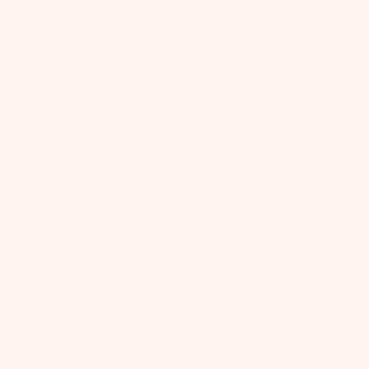}
        \caption{Sharp}
	\end{subfigure}
	
\caption{Examples on the RealBlur-J test dataset.}
\label{fig:supp-RealBlur-J}
\end{figure*}

\begin{figure*}[htbp]
\captionsetup[subfigure]{justification=centering, labelformat=empty}
\centering
\begin{subfigure}{0.135\linewidth}
		\includegraphics[width=0.99\linewidth]{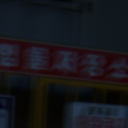}
	\end{subfigure}
	\begin{subfigure}{0.135\linewidth}
		\includegraphics[width=0.99\linewidth]{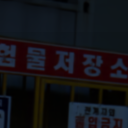}
	\end{subfigure}
	\begin{subfigure}{0.135\linewidth}
		\includegraphics[width=0.99\linewidth]{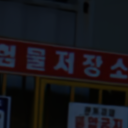}
	\end{subfigure}
	\begin{subfigure}{0.135\linewidth}
		\includegraphics[width=0.99\linewidth]{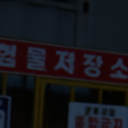}
	\end{subfigure}
	\begin{subfigure}{0.135\linewidth}
		\includegraphics[width=0.99\linewidth]{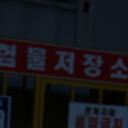}
	\end{subfigure}
    \begin{subfigure}{0.135\linewidth}
		\includegraphics[width=0.99\linewidth]{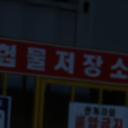}
	\end{subfigure}
     \begin{subfigure}{0.135\linewidth}
		\includegraphics[width=0.99\linewidth]{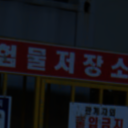}
	\end{subfigure}
	\quad
	\centering
		\begin{subfigure}{0.135\linewidth}
		\includegraphics[width=0.99\linewidth]{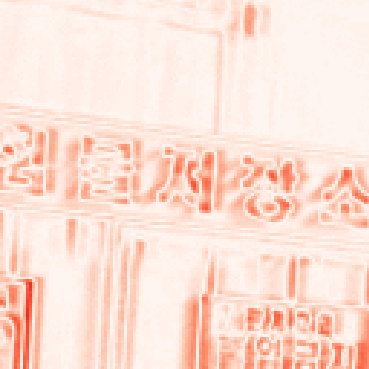}
	\end{subfigure}
	\begin{subfigure}{0.135\linewidth}
		\includegraphics[width=0.99\linewidth]{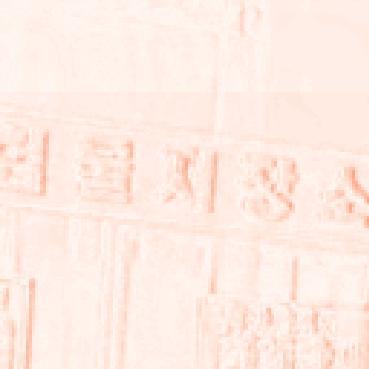}
	\end{subfigure}
	\begin{subfigure}{0.135\linewidth}
		\includegraphics[width=0.99\linewidth]{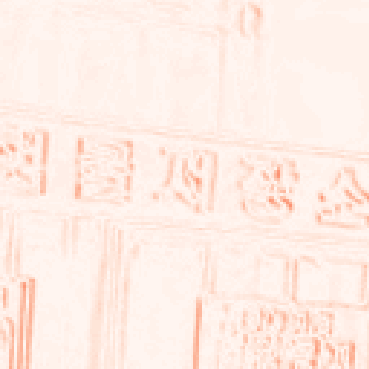}
	\end{subfigure}
	\begin{subfigure}{0.135\linewidth}
		\includegraphics[width=0.99\linewidth]{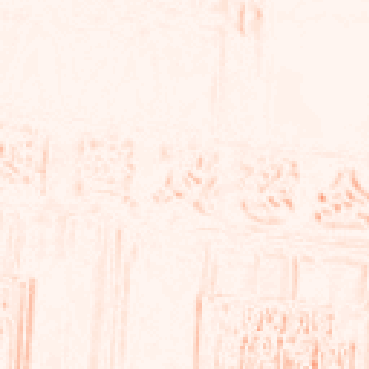}
	\end{subfigure}
	\begin{subfigure}{0.135\linewidth}
		\includegraphics[width=0.99\linewidth]{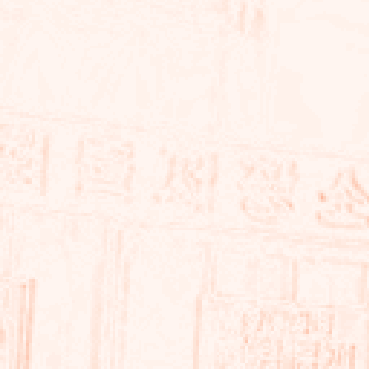}
	\end{subfigure}
    \begin{subfigure}{0.135\linewidth}
		\includegraphics[width=0.99\linewidth]{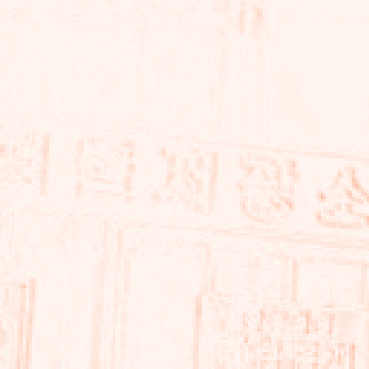}
	\end{subfigure}
     \begin{subfigure}{0.135\linewidth}

		\includegraphics[width=0.99\linewidth]{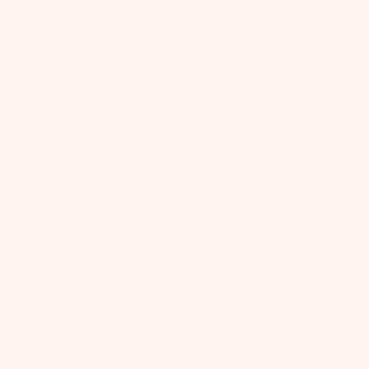}
	\end{subfigure}
 \quad
	\begin{subfigure}{0.135\linewidth}
		\includegraphics[width=0.99\linewidth]{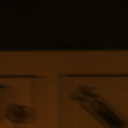}
	\end{subfigure}
	\begin{subfigure}{0.135\linewidth}
		\includegraphics[width=0.99\linewidth]{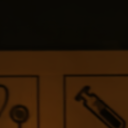}
	\end{subfigure}
	\begin{subfigure}{0.135\linewidth}
		\includegraphics[width=0.99\linewidth]{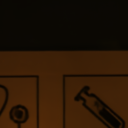}
	\end{subfigure}
	\begin{subfigure}{0.135\linewidth}
		\includegraphics[width=0.99\linewidth]{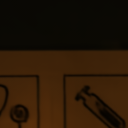}
	\end{subfigure}
	\begin{subfigure}{0.135\linewidth}
		\includegraphics[width=0.99\linewidth]{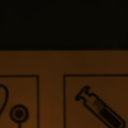}
	\end{subfigure}
    \begin{subfigure}{0.135\linewidth}
		\includegraphics[width=0.99\linewidth]{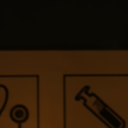}
	\end{subfigure}
     \begin{subfigure}{0.135\linewidth}
		\includegraphics[width=0.99\linewidth]{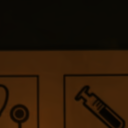}
	\end{subfigure}
	\quad
	\centering
		\begin{subfigure}{0.135\linewidth}
		\includegraphics[width=0.99\linewidth]{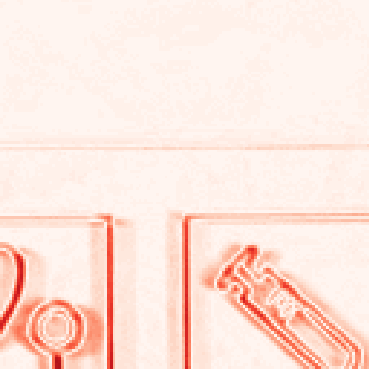}
	\end{subfigure}
	\begin{subfigure}{0.135\linewidth}
		\includegraphics[width=0.99\linewidth]{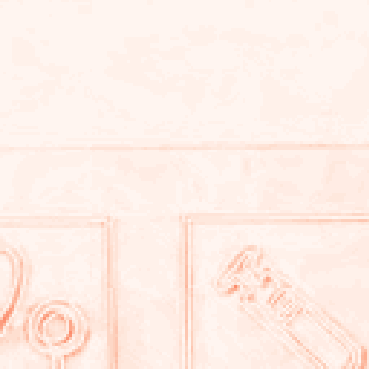}
	\end{subfigure}
	\begin{subfigure}{0.135\linewidth}
		\includegraphics[width=0.99\linewidth]{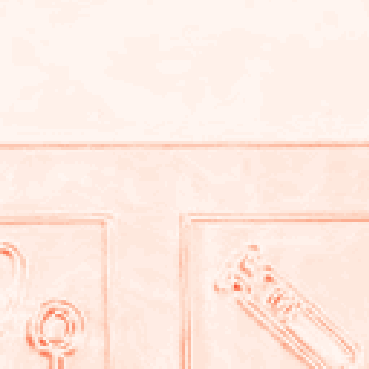}
	\end{subfigure}
	\begin{subfigure}{0.135\linewidth}
		\includegraphics[width=0.99\linewidth]{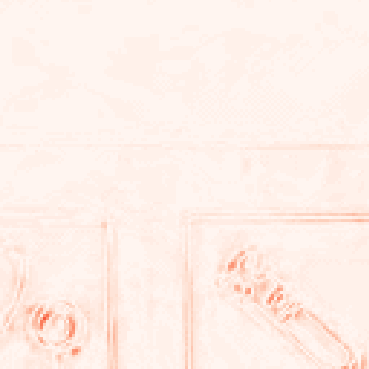}
	\end{subfigure}
	\begin{subfigure}{0.135\linewidth}
		\includegraphics[width=0.99\linewidth]{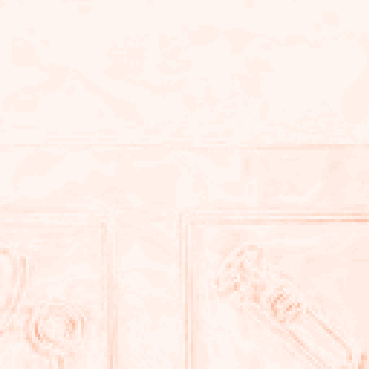}
	\end{subfigure}
    \begin{subfigure}{0.135\linewidth}
		\includegraphics[width=0.99\linewidth]{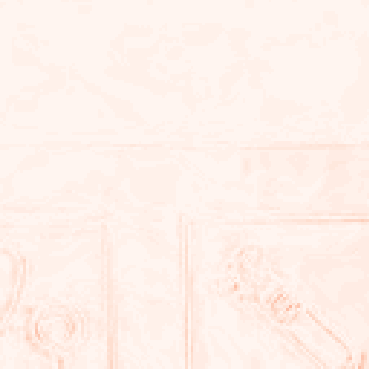}
	\end{subfigure}
     \begin{subfigure}{0.135\linewidth}

		\includegraphics[width=0.99\linewidth]{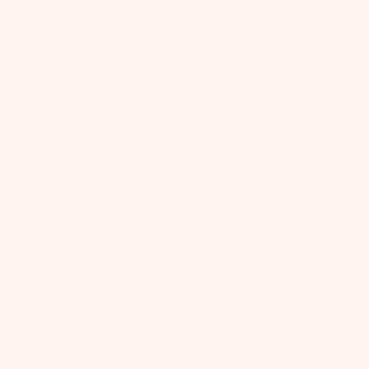}
	\end{subfigure}
 \quad
 \centering
	\begin{subfigure}{0.135\linewidth}
		\includegraphics[width=0.99\linewidth]{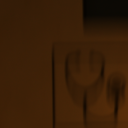}
	\end{subfigure}
	\begin{subfigure}{0.135\linewidth}
		\includegraphics[width=0.99\linewidth]{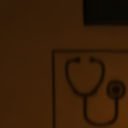}
	\end{subfigure}
	\begin{subfigure}{0.135\linewidth}
		\includegraphics[width=0.99\linewidth]{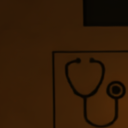}
	\end{subfigure}
	\begin{subfigure}{0.135\linewidth}
		\includegraphics[width=0.99\linewidth]{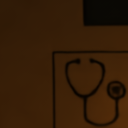}
	\end{subfigure}
	\begin{subfigure}{0.135\linewidth}
		\includegraphics[width=0.99\linewidth]{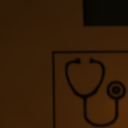}
	\end{subfigure}
    \begin{subfigure}{0.135\linewidth}
		\includegraphics[width=0.99\linewidth]{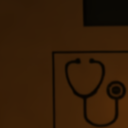}
	\end{subfigure}
     \begin{subfigure}{0.135\linewidth}
		\includegraphics[width=0.99\linewidth]{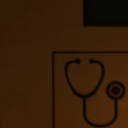}
	\end{subfigure}
	\quad
	\centering
		\begin{subfigure}{0.135\linewidth}
		\includegraphics[width=0.99\linewidth]{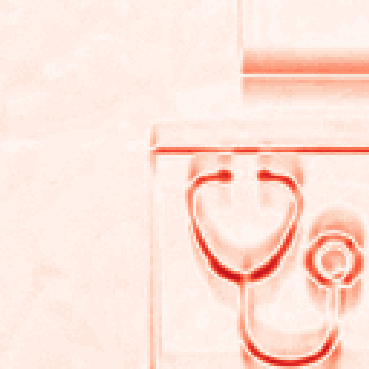}
	\end{subfigure}
	\begin{subfigure}{0.135\linewidth}
		\includegraphics[width=0.99\linewidth]{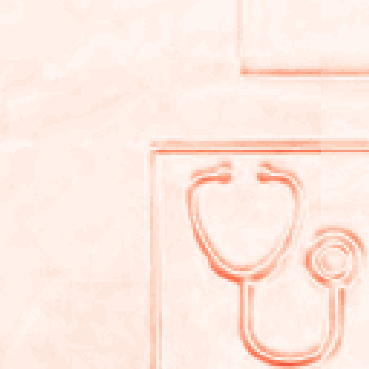}
	\end{subfigure}
	\begin{subfigure}{0.135\linewidth}
		\includegraphics[width=0.99\linewidth]{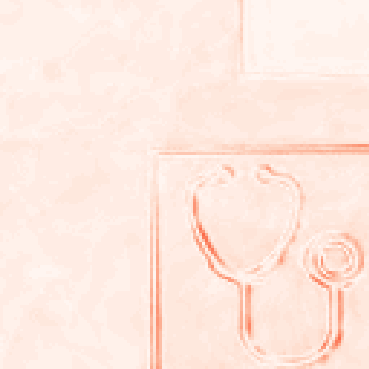}
	\end{subfigure}
	\begin{subfigure}{0.135\linewidth}
		\includegraphics[width=0.99\linewidth]{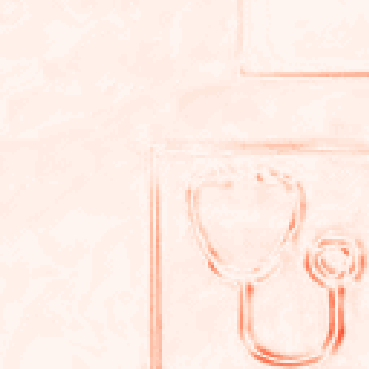}
	\end{subfigure}
	\begin{subfigure}{0.135\linewidth}
		\includegraphics[width=0.99\linewidth]{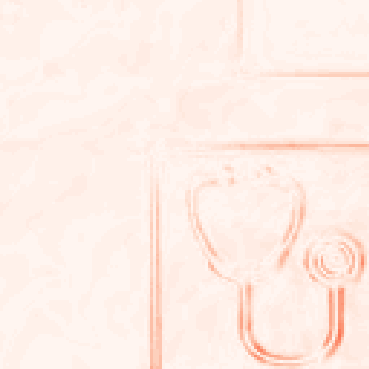}
	\end{subfigure}
    \begin{subfigure}{0.135\linewidth}
		\includegraphics[width=0.99\linewidth]{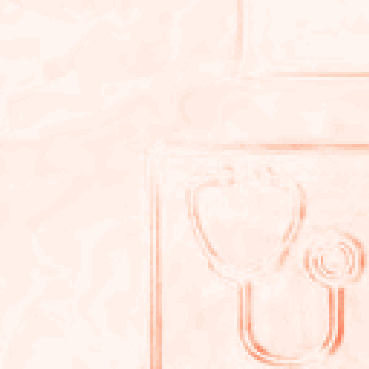}
	\end{subfigure}
     \begin{subfigure}{0.135\linewidth}

		\includegraphics[width=0.99\linewidth]{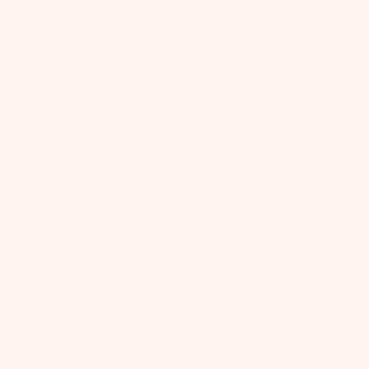}
	\end{subfigure}
  \quad
 \centering
	\begin{subfigure}{0.135\linewidth}
		\includegraphics[width=0.99\linewidth]{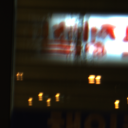}
	\end{subfigure}
	\begin{subfigure}{0.135\linewidth}
		\includegraphics[width=0.99\linewidth]{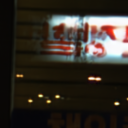}
	\end{subfigure}
	\begin{subfigure}{0.135\linewidth}
		\includegraphics[width=0.99\linewidth]{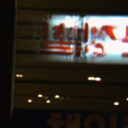}
	\end{subfigure}
	\begin{subfigure}{0.135\linewidth}
		\includegraphics[width=0.99\linewidth]{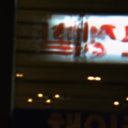}
	\end{subfigure}
	\begin{subfigure}{0.135\linewidth}
		\includegraphics[width=0.99\linewidth]{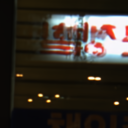}
	\end{subfigure}
    \begin{subfigure}{0.135\linewidth}
		\includegraphics[width=0.99\linewidth]{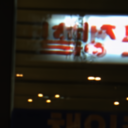}
	\end{subfigure}
     \begin{subfigure}{0.135\linewidth}
		\includegraphics[width=0.99\linewidth]{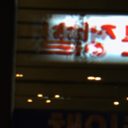}
	\end{subfigure}
	\quad
	\centering
		\begin{subfigure}{0.135\linewidth}
		\includegraphics[width=0.99\linewidth]{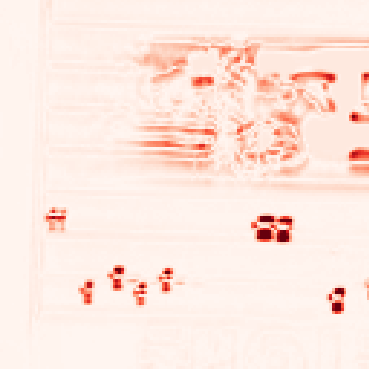}
        \caption{Blur}
	\end{subfigure}
	\begin{subfigure}{0.135\linewidth}
		\includegraphics[width=0.99\linewidth]{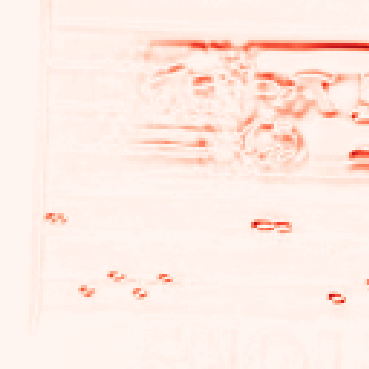}
        \caption{DeepRFT+~\cite{XintianMao2023DeepRFT}}
	\end{subfigure}
	\begin{subfigure}{0.135\linewidth}
		\includegraphics[width=0.99\linewidth]{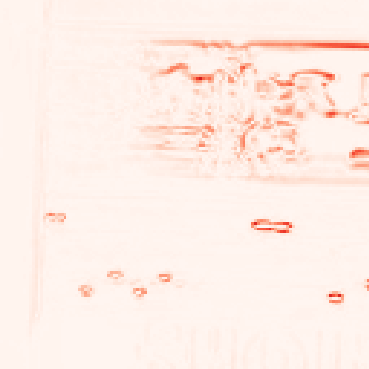}
        \caption{Stripformer~\cite{Tsai2022Stripformer}}
	\end{subfigure}
	\begin{subfigure}{0.135\linewidth}
		\includegraphics[width=0.99\linewidth]{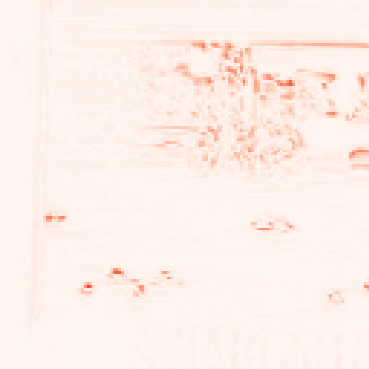}
          \caption{UFPNet~\cite{fang2023UFPNet}}
	\end{subfigure}
	\begin{subfigure}{0.135\linewidth}
		\includegraphics[width=0.99\linewidth]{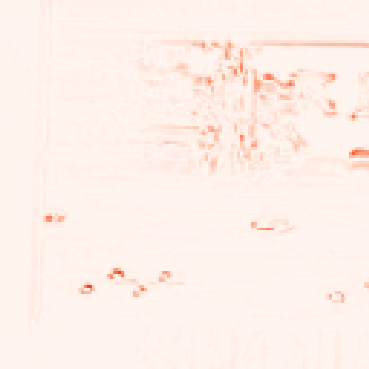}
        \caption{AdaRevD-B}
	\end{subfigure}
    \begin{subfigure}{0.135\linewidth}
		\includegraphics[width=0.99\linewidth]{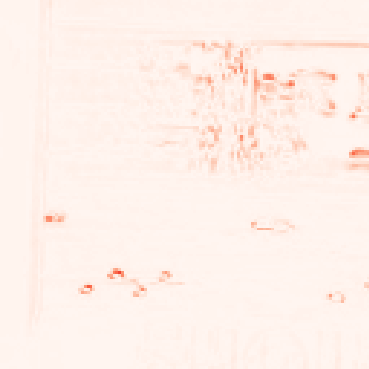}
        \caption{AdaRevD-L}
	\end{subfigure}
     \begin{subfigure}{0.135\linewidth}
		\includegraphics[width=0.99\linewidth]{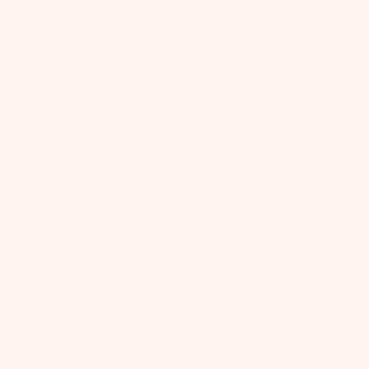}
        \caption{Sharp}
	\end{subfigure}
\caption{Examples on the RealBlur-R test dataset.}
\label{fig:supp-RealBlur-R}
\end{figure*}

The visual results for GoPro~\cite{Nah2017deep}, HIDE~\cite{Shen2019human}, RealBlur-R~\cite{Rim2020real} and RealBlur-J~\cite{Rim2020real} are presented in Figs.~\ref{fig:supp_GoPro},~\ref{fig:supp_HIDE},~\ref{fig:supp-RealBlur-J} and~\ref{fig:supp-RealBlur-R}, respectively. The visualizations depicted in Fig.~\ref{fig:supp_GoPro} and Fig.~\ref{fig:supp_HIDE} illustrate AdaRevD's capability to restore sharper images. We also show the visualization results on the RealBlur~\cite{Rim2020real} dataset in Fig.~\ref{fig:supp-RealBlur-J} and Fig.~\ref{fig:supp-RealBlur-R}. It can be observed that our model yields more visually pleasant outputs than other methods on both synthetic and real-world motion deblurring. This is evident when compared to other SOTA methods, such as DeepRFT~\cite{XintianMao2023DeepRFT} and UFPNet~\cite{fang2023UFPNet}. 




\begin{table*}[t]
\begin{center}
\caption{Improvement of diffent sub-decoders in RevD-L on GoPro dataset.}
\label{tab:RevD-L-GoPro}
\renewcommand\arraystretch{1}
\setlength{\tabcolsep}{1.9pt}
\resizebox{0.99\linewidth}{!}{
\begin{tabular}{c|c|c|c|c|c|c|c|c|c|c|c|c|c|c|c|c}
\toprule[0.15em]
& \multicolumn{8}{c}{\textbf{TrainSet}} &\multicolumn{8}{c}{\textbf{TestSet}}\\
\arrayrulecolor{black!30}\cmidrule(lr){2-9}\cmidrule(lr){10-17}
\textbf{Degree} & \textbf{dec1} & \textbf{dec2} & \textbf{dec3} & \textbf{dec4} & \textbf{dec5} & \textbf{dec6} & \textbf{dec7} & \textbf{dec8} & \textbf{dec1} & \textbf{dec2} & \textbf{dec3} & \textbf{dec4} & \textbf{dec5} & \textbf{dec6} & \textbf{dec7} & \textbf{dec8}\\
\midrule[0.15em] 
{$\le$20} & 11.068 &0.601 &0.257 &0.222 &0.198 &\cellcolor{color4}0.109 &0.047 &0.014 & 10.201 &0.610 &0.252 &0.269 &0.198 &\cellcolor{color4}0.107 &0.048 &0.015
\\
\arrayrulecolor{black!30}\midrule
{20-25} & 10.903 &0.400 &0.154 &0.135 &0.118 &\cellcolor{color4}0.064 &0.026 &0.007 & 9.569 &0.345 &0.129 &0.153 &0.122 &\cellcolor{color4}0.055 &0.021 &0.007
\\
\arrayrulecolor{black!30}\midrule
{25-30} & 9.141 &0.227 &0.076 &0.070 &\cellcolor{color4}0.060 &0.031 &0.012 &0.003 & 8.053 &0.206 &0.065 &0.073 &\cellcolor{color4}0.064 &0.030 &0.009 &0.003
\\
\arrayrulecolor{black!30}\midrule
{30-35} & 6.176 &\cellcolor{color4}0.145 &0.037 &0.035 &0.029 &0.015 &0.007 &0.002 & 5.362 &\cellcolor{color4}0.125 &0.030 &0.041 &0.031 &0.015 &0.005 &0.002
\\
\arrayrulecolor{black!30}\midrule
{35-40} & 3.437 &\cellcolor{color4}0.101 &0.019 &0.018 &0.014 &0.007 &0.004 &0.001 & 2.798 &\cellcolor{color4}0.116 &0.012 &0.023 &0.016 &0.010 &0.003 &0.001
\\
\arrayrulecolor{black!30}\midrule
{$>$40} & 2.365 &\cellcolor{color4}0.057 &0.016 &0.010 &0.007 &0.007 &0.008 &0.002 & 1.470 &\cellcolor{color4}0.067 &0.013 &0.018 &0.014 &0.006 &0.003 &0.001
\\
\arrayrulecolor{black}\bottomrule[0.15em]
\end{tabular} }
\end{center} 
\end{table*}

\begin{table*}[t]
\begin{center}
\caption{Improvement of diffent sub-decoders in RevD-L on RealBlur-J dataset.}
\label{tab:RevD-L-RealBlur-J}
\renewcommand\arraystretch{1}
\setlength{\tabcolsep}{1.9pt}
\resizebox{0.99\linewidth}{!}{
\begin{tabular}{c|c|c|c|c|c|c|c|c|c|c|c|c|c|c|c|c}
\toprule[0.15em]
& \multicolumn{8}{c}{\textbf{TrainSet}} &\multicolumn{8}{c}{\textbf{TestSet}}\\
\arrayrulecolor{black!30}\cmidrule(lr){2-9}\cmidrule(lr){10-17}
\textbf{Degree} & \textbf{dec1} & \textbf{dec2} & \textbf{dec3} & \textbf{dec4} & \textbf{dec5} & \textbf{dec6} & \textbf{dec7} & \textbf{dec8} & \textbf{dec1} & \textbf{dec2} & \textbf{dec3} & \textbf{dec4} & \textbf{dec5} & \textbf{dec6} & \textbf{dec7} & \textbf{dec8}\\
\midrule[0.15em] 
{$\le$20} & 11.718 &0.788 &0.429 &0.227 &0.221 &0.131 &\cellcolor{color4}0.101 &0.006 &5.131 &0.074 &0.065 &0.052 &\cellcolor{color4}0.056 &0.032 &0.022 &0.002
\\
\arrayrulecolor{black!30}\midrule
{20-25} & 11.113 &0.606 &0.330 &0.190 &0.161 &0.082 &\cellcolor{color4}0.066 &0.004 &4.813 &0.125 &0.092 &0.050 &\cellcolor{color4}0.056 &0.033 &0.024 &0.002
\\
\arrayrulecolor{black!30}\midrule
{25-30} & 9.956 &0.419 &0.218 &0.133 &\cellcolor{color4}0.096 &0.045 &0.039 &0.003 &4.119 &0.089 &\cellcolor{color4}0.067 &0.045 &0.037 &0.019 &0.014 &0.001
\\
\arrayrulecolor{black!30}\midrule
{30-35} & 8.348 &0.263 &0.129 &0.082 &\cellcolor{color4}0.050 &0.022 &0.025 &0.002 &3.486 &\cellcolor{color4}0.066 &0.038 &0.028 &0.021 &0.009 &0.010 &0.001
\\
\arrayrulecolor{black!30}\midrule
{35-40} & 7.040 &0.186 &\cellcolor{color4}0.091 &0.048 &0.029 &0.0159 &0.019 &0.002 &2.444 &0.052 &\cellcolor{color4}0.050 &0.032 &0.013 &0.004 &0.006 &0.000
\\
\arrayrulecolor{black!30}\midrule
{$>$40} & 5.444 &0.158 &\cellcolor{color4}0.081 &0.039 &0.022 &0.014 &0.018 &0.001 &\cellcolor{color4}1.462 &-0.021 &-0.018 &0.026 &-0.021 &0.007 &0.003 &-0.003

\\
\arrayrulecolor{black}\bottomrule[0.15em]
\end{tabular} }
\end{center} 
\end{table*}

\begin{table*}[t]
\begin{center}
\caption{Improvement of diffent sub-decoders in RevD-L on RealBlur-R dataset.}
\label{tab:RevD-L-RealBlur-R}
\renewcommand\arraystretch{1}
\setlength{\tabcolsep}{1.9pt}
\resizebox{0.99\linewidth}{!}{
\begin{tabular}{c|c|c|c|c|c|c|c|c|c|c|c|c|c|c|c|c}
\toprule[0.15em]
& \multicolumn{8}{c}{\textbf{TrainSet}} &\multicolumn{8}{c}{\textbf{TestSet}}\\
\arrayrulecolor{black!30}\cmidrule(lr){2-9}\cmidrule(lr){10-17}
\textbf{Degree} & \textbf{dec1} & \textbf{dec2} & \textbf{dec3} & \textbf{dec4} & \textbf{dec5} & \textbf{dec6} & \textbf{dec7} & \textbf{dec8} & \textbf{dec1} & \textbf{dec2} & \textbf{dec3} & \textbf{dec4} & \textbf{dec5} & \textbf{dec6} & \textbf{dec7} & \textbf{dec8}\\
\midrule[0.15em] 
{$\le$20} & 12.037 &1.174 &0.357 &0.105 &0.188 &\cellcolor{color4}0.117 &0.029 &0.000 &5.524 &0.178 &\cellcolor{color4}0.062 &0.033 &0.062 &0.032 &0.010 &0.000
\\
\arrayrulecolor{black!30}\midrule
{20-25} & 11.988 &1.102 &0.314 &0.098 &0.174 &\cellcolor{color4}0.098 &0.024 &0.000 &5.252 &0.227 &\cellcolor{color4}0.062 &0.037 &0.064 &0.023 &0.007 &0.000
\\
\arrayrulecolor{black!30}\midrule
{25-30} & 11.895 &1.030 &0.259 &0.079 &0.132 &\cellcolor{color4}0.070 &0.019 &0.000 &5.611 &0.214 &\cellcolor{color4}0.062 &0.036 &0.046 &0.021 &0.006 &0.000
\\
\arrayrulecolor{black!30}\midrule
{30-35} & 11.260 &0.891 &0.215 &0.060 &0.094 &\cellcolor{color4}0.055 &0.017 &0.000 &4.190 &\cellcolor{color4}0.190 &0.041 &0.023 &0.032 &0.016 &0.005 &0.000
\\
\arrayrulecolor{black!30}\midrule
{35-40} & 9.419 &0.640 &\cellcolor{color4}0.136 &0.035 &0.055 &0.038 &0.013 &0.000 &3.372 &\cellcolor{color4}0.189 &0.037 &0.019 &0.026 &0.017 &0.006 &0.000

\\
\arrayrulecolor{black!30}\midrule
{40-45} & 7.243 &0.424 &\cellcolor{color4}0.085 &0.022 &0.030 &0.027 &0.009 &0.000 &3.016 &\cellcolor{color4}0.141 &0.031 &0.016 &0.018 &0.016 &0.004 &0.000
\\
\arrayrulecolor{black!30}\midrule
{45-50} & 5.093 &\cellcolor{color4}0.259 &0.044 &0.013 &0.014 &0.015 &0.005 &0.000 &1.991 &\cellcolor{color4}0.058 &0.020 &0.007 &0.006 &0.008 &0.003 &0.000
\\
\arrayrulecolor{black!30}\midrule
{$>$50} & 3.139 &\cellcolor{color4}0.183 &0.020 &0.008 &0.006 &0.007 &0.003 &0.000 &\cellcolor{color4}1.128 &-0.014 &0.010 &0.007 &0.003 &0.008 &0.002 &0.000
\\
\arrayrulecolor{black}\bottomrule[0.15em]
\end{tabular} }
\end{center} 
\end{table*}



\end{document}